\documentclass[11pt]{article}

\usepackage[preprint]{acl}

\usepackage{times}
\usepackage{latexsym}
\usepackage{placeins}

\usepackage[T1]{fontenc}

\usepackage[utf8]{inputenc}
\usepackage[dvipsnames]{xcolor}

\usepackage{microtype}

\usepackage{inconsolata}

\usepackage{graphicx}
\usepackage[utf8]{inputenc} %
\usepackage[T1]{fontenc}    %
\usepackage{hyperref}       %
\usepackage{url}            %
\usepackage{booktabs}       %
\usepackage{amsfonts}       %
\usepackage{nicefrac}       %
\usepackage{microtype}      %
\usepackage{xcolor}         %
\usepackage{graphicx}
\usepackage{tcolorbox}
\usepackage{array}
\usepackage{tabularx}
\usepackage{multirow}
\usepackage{amsmath}
\usepackage{booktabs}
\usepackage{subcaption}
\usepackage{cuted}
\usepackage{caption}
\usepackage{adjustbox}
\usepackage[capitalize,noabbrev,nameinlink]{cleveref}
\usepackage{algorithm}
\usepackage{algorithmic}
\usepackage{wrapfig}
\usepackage{enumitem}
\usepackage{tcolorbox}

\tcbuselibrary{most}

\newtcolorbox{promptbox}[2][black]{
    colframe=#1,                %
    colback=white,              %
    coltitle=white,             %
    title={\textbf{#2}},        %
    fonttitle=\bfseries,        %
    boxrule=1pt,                %
    arc=1mm,                    %
    left=3pt, right=3pt, top=3pt, bottom=3pt, %
    bottomrule=1pt, toprule=1pt %
}

\title{\textsc{GFlowPO}: Generative Flow Network \\ as a Language Model Prompt Optimizer}

\author{
  Junmo Cho$^{1}$\thanks{Equal contribution.} \quad
  Suhan Kim$^{2}$\footnotemark[1] \quad
  Sangjune An$^{2}$\footnotemark[1] \quad
  Minsu Kim$^{1,3}$ \\
  \bfseries Dong Bok Lee$^{1}$ \quad
  Heejun Lee$^{1}$ \quad
  Sung Ju Hwang$^{1,4}$ \quad
  Hae Beom Lee$^{2}$\thanks{Corresponding author.} \\[2pt]
  {\normalfont $^{1}$KAIST \quad $^{2}$Korea University} \\
  {\normalfont $^{3}$Mila -- Qu\'ebec AI Institute \quad $^{4}$DeepAuto.ai} \\
  {\normalfont\ttfamily junmokane@kaist.ac.kr \quad \{suhan,ics8088,haebeomlee\}@korea.ac.kr}
}

\begin{document}
\maketitle
\begin{abstract}
Finding effective prompts for language models (LMs) is critical yet notoriously difficult: the prompt space is combinatorially large, and reward feedback is costly because each candidate prompt must be evaluated with a target LM. Existing RL-based prompt optimizers often rely on on-policy updates and meta-prompts sampled from a fixed distribution, leading to poor sample efficiency. We propose \textsc{GFlowPO}, a probabilistic prompt optimization framework that casts prompt search as a \textit{posterior inference} problem over latent prompts regularized by a meta-prompted reference-LM prior. In the first step, we fine-tune a lightweight prompt-LM with an \textit{off-policy} Generative Flow Network (GFlowNet) objective, using replay-based training to reuse past prompt evaluations for sample-efficient exploration. In the second step, we introduce \textit{Dynamic Memory Update} (DMU), a training-free mechanism that updates the meta-prompt by injecting both (i) diverse prompts from a replay buffer and (ii) top-performing prompts from a small priority queue, thereby progressively concentrating the search process on high-reward regions. Across 9 benchmarks spanning few-shot text classification, instruction induction, question answering, and reasoning, \textsc{GFlowPO} consistently outperforms recent discrete prompt optimization baselines achieving superior sample efficiency.
\end{abstract}

\section{Introduction}
Injecting appropriate prompts into language models (LMs) is critical to their performance, as small changes in prompts can lead to significant differences in model outputs \citep{salinas-morstatter-2024-butterfly, chatterjee-etal-2024-posix}.
As a result, exploration over the prompt space has become as important as optimization over the parameter space of LMs, giving rise to new gradient-free learning paradigms such as in-context learning, training-free reinforcement learning, and related approaches \citep{deng2022rlprompt, yang2024opro, fernando2023promptbreederselfreferentialselfimprovementprompt}. However, prompt engineering in practice remains a highly manual and labor-intensive process, relying heavily on empirical tuning.

To address this limitation, automatic prompt optimization has recently gained popularity \citep{ramnath-etal-2025-systematic}. In this setting, searching over the prompt space can be viewed as a black-box combinatorial optimization problem, where the goal is to maximize the performance of an LM on a target task. Reinforcement learning (RL) has emerged as a promising approach for automating this process, particularly by fine-tuning prompt-generating LMs (prompt-LMs) that already encode rich prior world knowledge \citep{kwon-etal-2024-stableprompt, batorski2025prlpromptsreinforcementlearning, gao2026p1betterpromptoptimization}. 
By guiding such prompt-LMs with a meta-prompt and optimizing them through RL with rewards from target tasks, these models can evolve to generate effective prompts for improving target LM performance. 
\
\begin{figure}[t]
  \centering
  \includegraphics[width=\columnwidth]{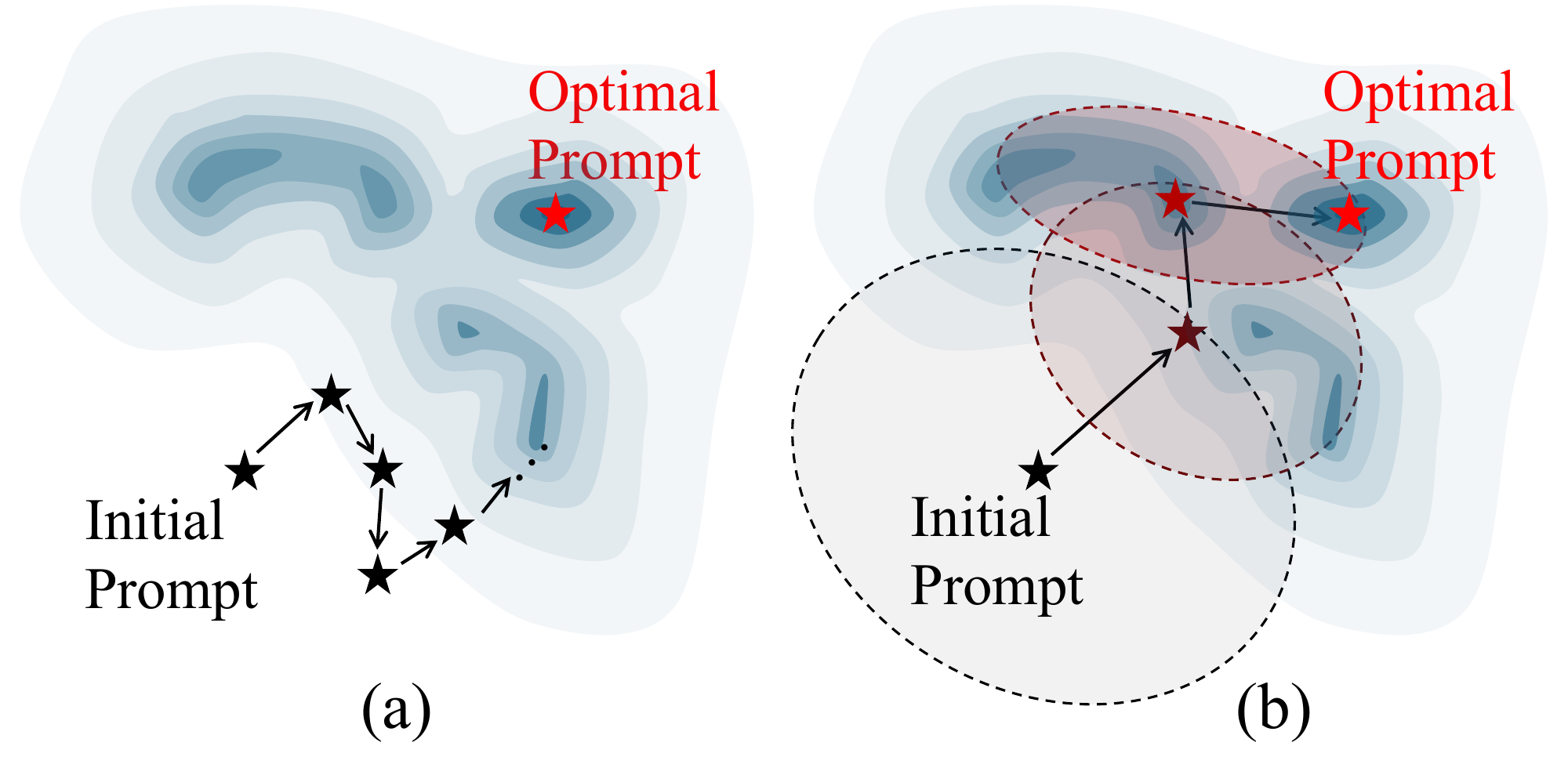}
  \vspace{-0.2in}
  \caption{\textbf{Concepts.} Blue contour indicates high performing prompt regions. \textbf{(a)} Existing on-policy RL frameworks fail to explore the huge combinatorial search space with poor sample efficiency. \textbf{(b)} Our \textsc{GFlowPO} can efficiently explore the search space by gradually annealing the posterior density (dotted ellipses) toward more promising areas with off-policy GFlowNet training in \textsc{step-A} and DMU in \textsc{step-B}.}
  \label{fig:main}
\end{figure}

Despite their promise, existing RL-based approaches face exploration challenges in prompt optimization over large combinatorial spaces.
First, many methods rely on on-policy learning, where training signals are obtained from Monte Carlo estimates based on samples from the current policy. In prompt optimization settings with sparse rewards (due to expensive cost of evaluating target LMs) and massive action space, this often requires a large number of samples to obtain reliable gradients, leading to poor sample efficiency.
Second, prompt search performance critically depends on contextual conditioning, yet the meta-prompt of the prompt-LM is sampled from a fixed distribution. This conditioning limits the incorporation of accumulated high-reward experience, resulting in weakly guided exploration and further reducing sample efficiency, as shown in Figure~\ref{fig:main}(a).

In this paper, we propose a novel probabilistic framework for prompt optimization, termed \textsc{GFlowPO}, which addresses the aforementioned key limitations of existing RL-based approaches. 
\textsc{GFlowPO} consists of two alternating steps.
\textbf{\textsc{step-A}}: sample-efficient off-policy posterior inference over latent prompts using Generative Flow Networks~\citep[GFlowNets;][]{bengio2021flownetworkbasedgenerative}, and \textbf{\textsc{step-B}}: training-free updates of the conditioning meta-prompt with the posterior samples.
This design allows the search process to progressively concentrate on high-reward regions of the prompt space, thereby significantly improving the sample efficiency of exploration. An overview of the framework is illustrated in Figure~\ref{fig:main}(b).

Specifically, \textsc{step-A} fine-tunes a prompt-LM with GFlowNets~\citep{bengio2021flownetworkbasedgenerative, JMLR:v24:22-0364}, an off-policy soft RL framework for amortized inference that samples prompts proportionally to reward. Its off-policy nature allows us to reuse past prompt evaluations through a replay buffer, which substantially improves sample efficiency under sparse reward signals. \textsc{step-B} then updates the meta-prompt through a Dynamic Memory Update (DMU) mechanism: it injects reference prompts drawn from both diverse replayed prompts and top-performing prompts into the meta-prompt, steering subsequent search toward high-reward regions without any parameter updates while preserving exploration (Figure~\ref{fig:main}(b)).

To validate the efficacy of our method, we conduct extensive experiments across a diverse set of tasks and LLMs. 
The datasets include text classification \citep{wang-etal-2018-glue, wang2020supergluestickierbenchmarkgeneralpurpose}, instruction induction \citep{honovich2022induction, Ghazal2013BigBench}, question answering \citep{hendrycks2021measuringmassivemultitasklanguage, mihaylov2018openbookqa}, and reasoning tasks \citep{cobbe2021GSM8Kdataset, suzgun2022challenging, balunovic_srimatharena_2025}, covering 9 benchmarks.
We consider both prompt-LMs and target LMs with model sizes ranging from 2B to 11B, including Llama ~\citep{grattafiori2024llama3}, Mistral \citep{jiang2023mistral7b}, Gemma \citep{gemmateam2024gemmaopenmodelsbased}, Falcon \citep{almazrouei2023falconseriesopenlanguage}, and Qwen \citep{yang2025qwen3technicalreport}. Across these settings, our method consistently achieves strong performance across diverse tasks and model scales.

We summarize our main contributions as follows: \textbf{(1)} we cast prompt optimization as posterior inference and propose \textsc{GFlowPO}, a sample-efficient prompt optimizer based on off-policy GFlowNet learning; 
\textbf{(2)} we introduce a training-free Dynamic Memory Update that adapts prompt search toward high-reward regions; and 
\textbf{(3)} we demonstrate strong performance of \textsc{GFlowPO} across 9 benchmarks spanning text classification and generation, induction, question answering, and reasoning.

\section{Related Work}
\begin{figure*}[!t]
    \centering

    \vspace{-0.1in}
    \includegraphics[width=0.95\textwidth, keepaspectratio]{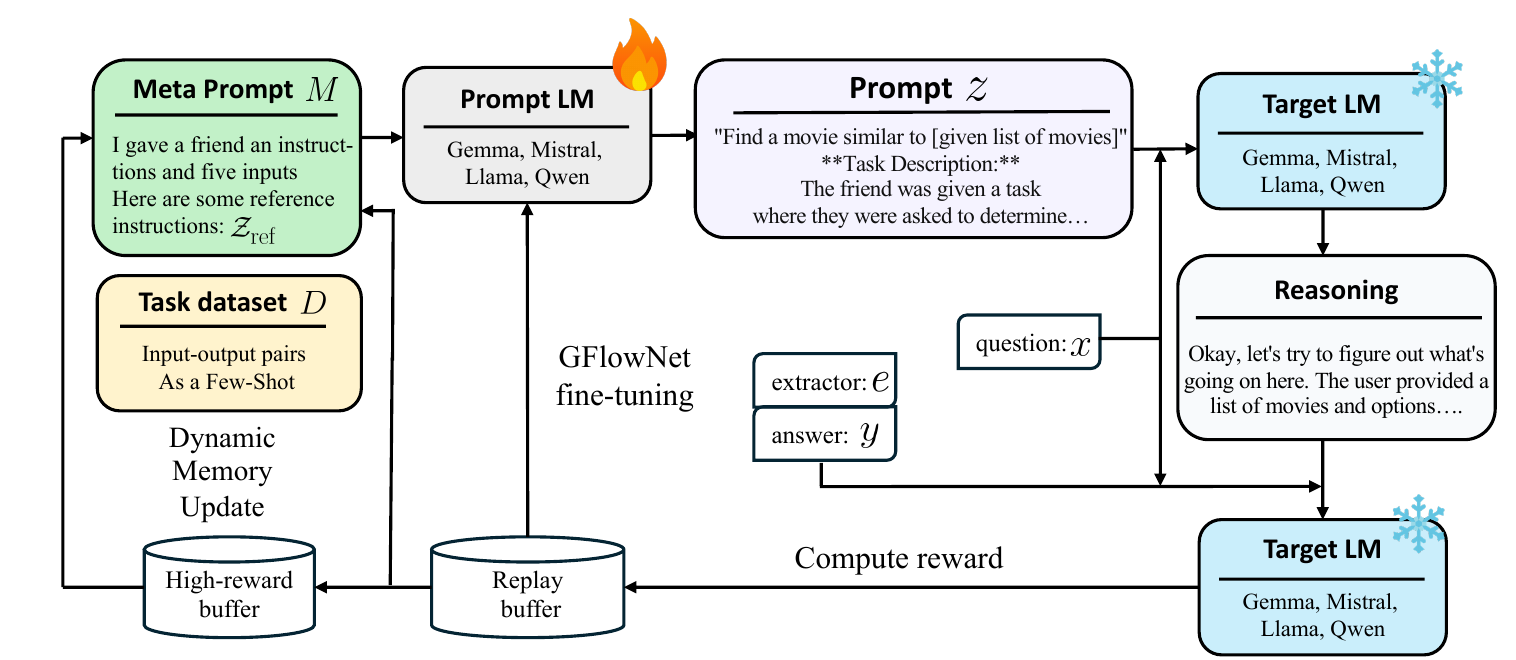}
    \caption{
    \textbf{GFlowPO pipeline.} The optimizer prompt-LM samples prompts conditioned on a meta-prompt $M$, and the target LLM provides task rewards. \textsc{GFlowPO} alternates between off-policy GFlowNet training for replay-based posterior inference and Dynamic Memory Update (DMU), which injects diverse and high-reward prompts into $M$ to improve exploration and prompt quality.
    }
    \label{fig:figure1}
\end{figure*}

\paragraph{Prompt optimization.}
Early works tune task-specific continuous prompts with gradients~\citep{li2021prefixtuningoptimizingcontinuousprompts, lester2021powerscaleparameterefficientprompt, Qin_2021} or search discrete tokens through gradients~\citep{shin2020autopromptelicitingknowledgelanguage, shi2022humanreadableprompttuning, das2025greatergradientsreasoningmakes}.
Since APE~\citep{zhou2023ape} showed that LLMs can serve as prompt optimizers, subsequent methods have explored textual gradients~\citep{pryzant2023protegi}, optimization in text space~\citep{yang2024opro, yuksekgonul2025optimizing}, evolutionary search~\citep{guo2025evopromptconnectingllmsevolutionary}, meta-prompt engineering~\citep{ye-etal-2024-prompt}, agent-based reasoning~\citep{wang2023promptagentstrategicplanninglanguage, shinn2023reflexion}, Bayesian optimization~\citep{agarwal2025searchingoptimalsolutionsllmsbayesianoptimization, liu2024llambo}, and LLM-based feedback~\citep{agarwal-etal-2025-promptwizard, shi2025lossgaingatedrefinement, agrawal2026gepareflectivepromptevolution}. Most of these rely on powerful closed-source LLMs, as lightweight optimizers often degrade performance~\citep{zhang2024revisitingoprolimitationssmallscale}.
A complementary line casts prompt optimization as RL: RLPrompt~\citep{deng2022rlprompt} and TEMPERA~\citep{zhang2023tempera} use small policy networks or edit-based action spaces, while StablePrompt~\citep{kwon-etal-2024-stableprompt} and PRL~\citep{batorski2025prlpromptsreinforcementlearning} fine-tune LLM-based policies with PPO- and GRPO-style updates, respectively. 
\textsc{GFlowPO} follows these RL-based methods in that it directly fine-tunes a lightweight open-source prompt-LM rather than relying on heavy external LLMs, but differs from existing RL methods by replacing on-policy updates with off-policy GFlowNet training and by introducing a training-free Dynamic Memory Update (DMU) that injects diverse and high-reward prompts into the meta-prompt, enabling more sample-efficient exploration in the combinatorial prompt space.

\paragraph{LM post-training with GFlowNets.}
GFlowNets \citep{bengio2021flownetworkbasedgenerative, JMLR:v24:22-0364} have been applied to LM post-training for their amortized inference, diversity, and off-policy sampling: infilling and chain-of-thought reasoning~\citep{hu2024amortizingintractableinferencelarge}, mathematical and puzzle reasoning with off-policy local search~\citep{yu2025flow}, decentralized training for reasoning-LMs~\citep{bartoldson2025trajectorybalanceasynchronydecoupling}, diverse red-teaming~\citep{lee2025learningdiverseattackslarge, yun2025active}, and diverse prompt generation for text-to-image models~\citep{yun2025learning}.
Our work also post-trains an LM with GFlowNets, but to the best of our knowledge, it is the first to use in-context memory updates in both the prior and posterior distributions. We defer a more comprehensive discussion of related work to Appendix~\ref{append:related_acl}.

\section{Approach}
\label{sec:approach}

We now introduce \textsc{GFlowPO}; the pipeline and pseudocode are provided in Figure~\ref{fig:figure1} and Appendix~\ref{appen:full_algorithm}, respectively.

\subsection{Overview of \textsc{GFlowPO}}

Given a task-specific distribution $p(x,y)$ over context $x$ and answer $y$, we aim to optimize a prompt $z$ that maximizes the performance of a target LM over $p(x,y)$:

{\begingroup
\begin{equation}
\max_{z}\ \mathbb{E}_{p(x,y)} \big[f(\hat{y}(x,z), y)\big]. \label{eq:objective}
\end{equation}
\endgroup}

where $\hat{y}(x,z)$ denotes the prediction of the target LM conditioned on context $x$ and prompt $z$ (e.g., greedy decoding). $f$ is a task-specific evaluation metric (e.g., accuracy).

In usual prompt optimization settings, $p(x,y)$ is unknown and only a tiny set of training dataset $\mathcal{D}=\{(x_i,y_i)\}_{i=1}^n$ is given for each task. We thus consider the following Bayesian posterior inference problem to prevent overfitting and adhere to linguistic plausibility:

{\begingroup
\begin{equation}
p(z|\mathcal{D},M) \propto p(\mathcal{D}|z) \cdot p_\text{ref}(z|M), \label{eq:posterior2}
\end{equation}
\endgroup}

where $p_\text{ref}(z|M)$ is the conditional prior over prompts induced by a \emph{reference LM}, conditioned on the meta-prompt $M$ which is an instruction given to the reference LM as an input (see Figure~\ref{fig:metaprompt_template} for a meta-prompt template). $p(\mathcal{D}|z)$ is the likelihood, and $p(z|\mathcal{D},M)$ is the corresponding posterior. In this way, we can effectively regularize the prompt solution under scarce data and huge combinatorial search space, balancing task performance $p(\mathcal{D}|z)$ and linguistic plausibility via the reference LM $p_\text{ref}(z|M)$.

Note that the RHS in Equation~\ref{eq:posterior2} is unnormalized and highly multi-modal in the huge combinatorial search space, making the inference problem very challenging. We thus use GFlowNets~\citep{bengio2021flownetworkbasedgenerative} to fine-tune another LM $p_{\theta}(z|M)$, which we call \emph{prompt-LM}, to approximately amortize the posterior $p(z|\mathcal{D},M)$ in Equation~\ref{eq:posterior2}. GFlowNets are a natural option to model a highly multi-modal unnormalized distribution, especially when the search space is discrete. GFlowNets also enable more sample-efficient off-policy training compared to the existing on-policy RL which resorts to MC estimates based on samples from on-policy, resulting in poor exploration under huge search spaces. This GFlowNet stage is what we call \textbf{\textsc{step-A}}.

Ideally, we want our approximate posterior $p_\theta$ to find all the modes in the search space, but GFlowNet (and other RL-based methods) can focus only on a relatively narrow region at a time~\citep{pan2023pretrainingfinetuninggenerativeflow}. Therefore, the role of meta-prompt $M$ is critical to properly guiding the search over the huge combinatorial search space. To this end, 
we let $M$ include a few reference prompts that guide the search, and progressively update those reference prompts into newer ones so that \emph{the search area can be reshaped gradually} (Figure~\ref{fig:main}(b)). While simple, we empirically observe that such training-free heuristic update of $M$ works well in practice. This 
second stage is what we call \textbf{\textsc{step-B}}.

We alternate between \textsc{step-A} and \textsc{step-B} for each iteration. The optimal prompt can be chosen by storing one or a few top-performing prompts found throughout the training. We next detail \textsc{step-A} in Section~\ref{sec:e_step_off_policy} and \textsc{step-B} in Section~\ref{sec:m_step_dmu}, respectively; together, they constitute the proposed \textsc{GFlowPO}.

\subsection{\textsc{step-A}: Off-policy Training with GFlowNets}
\label{sec:e_step_off_policy}
Generative Flow Networks~\citep[GFlowNets;][]{bengio2021flownetworkbasedgenerative} are off-policy RL algorithms designed for amortized inference of distributions of the form $p(z)\propto R(z)$. This framework naturally applies to approximating the posterior in Equation~\ref{eq:posterior2} by defining an unnormalized target reward as $R(z;M)= p(\mathcal{D}|z) \cdot p_{\text{ref}}(z|M)$. However, we empirically observed that the training likelihood $p(\mathcal{D}|z)$ is weakly correlated with the actual test accuracy for most of the prompt optimization tasks we considered (Appendix~\ref{appen:correlation}). We thus replace the likelihood with the training correct count $A_\mathcal{D}(z)$:

{\begingroup
\vspace{-4mm}
\begin{align}
R(z;M) = A_\mathcal{D}(z) \cdot p_\text{ref}(z|M) \label{eq:reward}
\vspace{-4mm}
\end{align}
\endgroup}

where $A_\mathcal{D}(z) := \epsilon + \sum_{(x,y) \in \mathcal{D}} f(\hat{y}(x,z),y)$ and $\epsilon \in \mathbb{R}_{>0}$ is a small constant to prevent $A_\mathcal{D}(z)$ from reducing to $0$.

\paragraph{Objective function.}
GFlowNets are trained via consistency-matching objectives that enforce forward--backward flow consistency with the target reward. In autoregressive settings such as language models, the backward transitions are fixed by the tokenization order, causing the objective to reduce to Path Consistency Learning~\citep[PCL;][]{nachum2017bridging}. We adopt a global path consistency-matching objective of the form:
\begin{equation}
\label{eqn:loss}
\begin{aligned}
\mathbb{E}_{z \sim \pi(z)}
\Bigl[
\bigl(
\log Z + \log p_{\theta}(z|M) - \log R(z;M)
\bigr)^2
\Bigr].
\end{aligned}
\end{equation}

where $\log Z$ denotes the log-partition function and $\pi(z)$ the training policy, e.g., $z$ is sampled from a tempered version of $p_\theta$ or a replay buffer. Under a sufficiently broad support of $\pi$, we can encourage $p_{\theta}(z|M)$ to asymptotically converge to the true target posterior. The log-partition $\log Z$ plays the role of a global value for each global path $z$. In Trajectory Balance~\citep[TB;][]{malkin2023trajectorybalanceimprovedcredit} and PCL, $\log Z$ is explicitly parameterized and learned. In contrast, VarGrad~\citep{richter2020vargradlowvariancegradientestimator, zhang2023robust} estimates $\log Z$ from minibatch samples $\{z_i\}_{i=1}^B \overset{\text{i.i.d.}}{\sim} \pi(z)$ as:
\begin{equation}
\begin{aligned}
&\log Z \\
&\approx \frac{1}{B}\sum_{i=1}^B \Big(\log R(z_i;M) - \log p_{\theta}(z_i | M)\Big).
\end{aligned}
\end{equation}

This relationship mirrors the distinction between value-based and value-free policy optimization methods (e.g., PPO vs.\ GRPO), where explicit value learning is replaced by minibatch-based estimation. In this work, we adopt the VarGrad objective, as explicitly learning $\log Z$ was found to be unstable in prompt optimization tasks. Further, we smooth the $\log Z$ estimation with exponential moving average (EMA) for learning stability under small $B$.

\paragraph{Replay buffer and off-policy learning.}
A key advantage of the GFlowNet objective is its compatibility with off-policy learning. Unlike on-policy methods such as PPO commonly used in prior prompt optimization work, GFlowNets naturally support replay-based training. Specifically, we let the training policy $\pi(z)$ be a mixture of $\tilde{p}_\theta$, a tempered version of $p_\theta$, and a uniform distribution $u_\mathcal{B}$ over the replay buffer $\mathcal{B}$, which stores $(z, A_\mathcal{D}(z))$ pairs collected from previous iterations:

{\begingroup
\vspace{-5mm}
\begin{align}
\pi(z) = \rho \tilde{p}_\theta(z) + (1-\rho) u_\mathcal{B}(z) 
\end{align}
\endgroup}

where $\rho \in [0,1]$ (e.g., $\rho=0.5$). Therefore, our off-policy training scheme substantially improves sample efficiency and stabilizes learning in the large and sparse prompt space. Note that, in the reward $R(z;M) = A_\mathcal{D}(z) \cdot p_\text{ref}(z|M)$, the prior $p_\text{ref}$ keeps evolving as $M$ is updated in our M-step. We thus need to re-evaluate $p_\text{ref}(z|M)$ every iteration for correct reward calculation, even with our off-policy training scheme. However, the cost of evaluating $p_\text{ref}(z|M)$ is significantly less than that of $A_\mathcal{D}(z)$, which makes it sufficient to store $(z, A_\mathcal{D}(z))$ pairs alone, preserving the sample efficiency of our off-policy training scheme. In addition, we find that initializing the replay buffer $\mathcal{B}$ with the prompts sampled from the initial $p_\text{ref}(z|M)$ facilitates exploration in the early stage of training. More details are provided in Appendix~\ref{appen:pre_step}.

\subsection{\textsc{step-B}: Dynamic Memory Update}
\label{sec:m_step_dmu}
While \textsc{step-A} performs amortized inference of the posterior over prompt space under a fixed meta-prompt $M$, \textsc{step-B} updates the meta-prompt itself ($M \rightarrow M'$) and correspondingly adapts the focus of search area with the updated approximate posterior $p_\theta(z|M')$ and reward $R(z;M')$. We refer to this process as Dynamic Memory Update (DMU).

\begin{figure}[t]
  \centering
  \begin{tcolorbox}[
    colback=gray!5,
    colframe=gray!40,
    width=\linewidth,
    boxrule=1pt,
    arc=3mm,
    left=2mm,
    right=2mm,
    top=1mm,
    bottom=1mm,
    title=Meta-prompt template,
    fonttitle=\bfseries,
    center title
  ]
  \small
  I gave a friend an instruction and five inputs. The friend read the instruction and wrote an output for every one of the inputs.\par\vspace{0.1em}

  Here are the input-output pairs:\par\vspace{0.1em}

  \begingroup
  \renewcommand{\arraystretch}{0.85}
  \noindent\makebox[\linewidth][c]{%
  \begin{tabular}{l @{\hspace{20pt}} l} 
  \textbf{Input:} $x_1$ & \textbf{Output:} $y_1$ \\
  \multicolumn{1}{c}{$\vdots$} & \multicolumn{1}{c}{$\vdots$} \\
  \textbf{Input:} $x_5$ & \textbf{Output:} $y_5$ \\
  \end{tabular}
  }
  \endgroup

  \par\vspace{0.1em}
  Here are some reference instructions: $\mathcal{Z}_{\mathrm{ref}}$\par
  \par\vspace{0.1em}
  \textbf{The instruction is:}
  \end{tcolorbox}
  \vspace{-0.18in}
  \caption{\textbf{Meta-prompt template} for \textsc{GFlowPO}.}
  \label{fig:metaprompt_template}
\end{figure}

\paragraph{Reference prompt update.}
We use a meta-prompt similar to \citet{zhou2023ape} and \citet{kwon-etal-2024-stableprompt}. As illustrated in Figure~\ref{fig:metaprompt_template}, the meta-prompt $M$ consists of (1) task-agnostic instructions, (2) $k$-shot input-output pairs $\left\{\left(x_i,y_i\right)\right\}_{i=1}^{k}$ randomly sampled from $\mathcal{D}$, and (3) reference prompts $\mathcal{Z}_\text{ref}$. Here, we focus on updating $\mathcal{Z}_{\text{ref}}$ in the meta-prompt $M$. 
The goal of DMU is to progressively update the focus of search towards more promising areas -- both the approximate posterior $p_\theta(z|M)$ and the reward $R(z;M)$ evolve as $M$ is updated.
To this end, DMU constructs $\mathcal{Z}_\text{ref}$ at each iteration and includes $\mathcal{Z}_\text{ref}$ in $M$,
by carefully sampling a few prompts from the two types of buffers: (1) a replay buffer $\mathcal{B}$ that stores the previous prompts sampled from $p_\theta$ throughout the training, and (2) a small high-reward buffer $\mathcal{Q}$ that keeps a few prompts with the highest accuracy so far:
\begin{equation}
\begin{aligned}
\mathcal{Z}_\text{ref}
&= \{z_b^{(i)}\}_{i=1}^{k_b}
\cup \{z_q^{(j)}\}_{j=1}^{k_q},
\quad \text{where} \notag\\[-0.2em]
&\hspace{-1.5em}
\{z_b^{(i)}\}_{i=1}^{k_b}
\overset{\text{i.i.d.}}{\sim} u_\mathcal{B}(z),
\;
\{z_q^{(j)}\}_{j=1}^{k_q}
\overset{\text{i.i.d.}}{\sim} u_\mathcal{Q}(z).
\label{eqn:ref_prompt}
\end{aligned}
\end{equation}
where $u_\mathcal{B}$ and $u_\mathcal{Q}$ are uniform distributions over $\mathcal{B}$ and $\mathcal{Q}$, respectively. Such a simple heuristic is computationally efficient and does not incur any additional parameter updates. The rationale behind $\mathcal{Q}$ is simply to select the prompts with the highest accuracy for exploitation, whereas random sampling from $\mathcal{B}$ is to promote exploration.
We found that $k_b=2$ and $k_q=1$ balance well between performance and computational cost. %
See Appendix~\ref{appen:ref_sampling} for more details.

\section{Experiment}
\begin{table*}[t]

\centering
\begin{sc}
\resizebox{\linewidth}{!}{
\begin{tabular}{@{}clccccccc@{}}
\toprule
& Method & SST-2 & MRPC & RTE & QNLI & MNLI & SNLI & Average \\
\midrule
\multirow{3}{*}{Fixed Prompt}
  & Few-Shot Prompt
  & $66.2{\scriptstyle \pm13.9}$ & $68.5{\scriptstyle \pm0.8}$ & $79.8{\scriptstyle \pm2.8}$ & $78.7{\scriptstyle \pm2.7}$ & $63.2{\scriptstyle \pm1.1}$ & $67.9{\scriptstyle \pm5.9}$ & $72.8{\scriptstyle \pm1.0}$ \\
  & Manual Prompt
  & $89.1$ & $51.0$ & $64.0$ & $73.0$ & $67.0$ & $47.0$ & $65.2$ \\
\midrule
\multirow{11}{*}{\shortstack[l]{Discrete\\Prompt\\Tuning}}
  & APE
  & $83.2$ & $55.3$ & $78.6$
  & $75.0$ & $54.6$ & $72.3$ & $70.1$ \\
  & ProTeGi
  & $69.2$ & $48.8$ & $73.2$
  & $74.2$ & $56.6$ & $61.3$ & $64.0$ \\
  & EvoPrompt
  & $89.8{\scriptstyle \pm0.0}$ & $\underline{71.7}{\scriptstyle \pm1.2}$ & $78.6{\scriptstyle \pm1.1}$
  & $78.8{\scriptstyle \pm0.3}$ & $62.2{\scriptstyle \pm1.9}$ & $58.8{\scriptstyle \pm3.8}$ & $73.3{\scriptstyle \pm0.5}$ \\
  & PromptWizard
  & $89.8{\scriptstyle \pm0.0}$ & $58.5{\scriptstyle \pm16.2}$ & $71.0{\scriptstyle \pm8.2}$
  & $64.9{\scriptstyle \pm1.9}$ & $58.4{\scriptstyle \pm2.1}$ & $62.5{\scriptstyle \pm5.5}$ & $67.5{\scriptstyle \pm3.7}$ \\
  & GRACE
  & $90.3{\scriptstyle \pm0.8}$ & $70.2{\scriptstyle \pm1.5}$ & $68.2{\scriptstyle \pm1.8}$
  & $65.0{\scriptstyle \pm3.4}$ & $63.3{\scriptstyle \pm2.3}$ & $62.4{\scriptstyle \pm1.0}$ & $69.9{\scriptstyle \pm0.2}$ \\
  & StablePrompt
  & $86.5{\scriptstyle \pm5.1}$ & $\textbf{73.2}{\scriptstyle \pm1.6}$ & $\underline{81.8}{\scriptstyle \pm1.5}$
  & $78.3{\scriptstyle \pm3.0}$ & $65.3{\scriptstyle \pm2.9}$ & $\underline{75.2}{\scriptstyle \pm1.2}$ & $75.8{\scriptstyle \pm1.2}$ \\
  & PRL
  & $\underline{90.9}{\scriptstyle \pm0.1}$ & $71.3{\scriptstyle \pm1.6}$ & $77.1{\scriptstyle \pm1.5}$
  & $\underline{78.9}{\scriptstyle \pm0.0}$ & $\underline{67.6}{\scriptstyle \pm2.5}$ & $70.1{\scriptstyle \pm4.6}$ & $\underline{76.0}{\scriptstyle \pm1.2}$ \\
  & GEPA
  & $87.6{\scriptstyle \pm7.6}$ & $70.2{\scriptstyle \pm0.5}$ & $79.2{\scriptstyle \pm0.8}$
  & $77.7{\scriptstyle \pm1.9}$ & $66.5{\scriptstyle \pm3.1}$ & $70.1{\scriptstyle \pm1.5}$ & $75.2{\scriptstyle \pm2.0}$ \\
\cmidrule(lr){2-9}
  & \textbf{GFlowPO}
  & $\textbf{93.0}{\scriptstyle \pm0.6}$ & $69.6{\scriptstyle \pm4.2}$ & $\textbf{82.0}{\scriptstyle \pm2.5}$
  & $\textbf{80.2}{\scriptstyle \pm3.4}$ & $\textbf{68.7}{\scriptstyle \pm3.2}$ & $\textbf{78.6}{\scriptstyle \pm2.7}$ & $\textbf{78.7}{\scriptstyle \pm0.9}$ \\
\bottomrule
\end{tabular}
}
\end{sc}
\caption{Results on few-shot text classification datasets. 
We report average accuracy with standard deviation over three runs using different random seeds. The best results are \textbf{bolded}, and the second-best results are \underline{underlined}.}
\label{tab:few_shot_tc_6}

\end{table*}

We next show the efficacy of our method over diverse benchmark datasets. The hyperparameters used in \textsc{GFlowPO} are provided in Appendix~\ref{appen:hparam}. For all our experiments, the prompt-LM is fine-tuned with LoRA~\citep{hu2021loralowrankadaptationlarge}, and each run took approximately two hours on a single B200 GPU. We present the full generated prompts in Appendix~\ref{appen:generated_prompts}, and analyze the prompt results in Appendix~\ref{appen:prompt_result_analysis}. We consider the following baselines.

\paragraph{Baselines.} We first consider fixed prompting strategies, such as hand-crafted \textbf{Manual Prompt}, and \textbf{Few-shot Prompt}~\citep{wei2022cotprompting}. For a direct comparison with our approach, we include several discrete prompt-tuning methods such as \textbf{APE} \citep{zhou2023ape}, \textbf{ProTeGi} \citep{pryzant2023protegi}, and \textbf{EvoPrompt}~\citep{guo2025evopromptconnectingllmsevolutionary}.
We also include \textbf{PromptWizard}~\citep{agarwal-etal-2025-promptwizard}, \textbf{GRACE}~\citep{shi2025lossgaingatedrefinement}, and \textbf{GEPA}~\citep{agrawal2026gepareflectivepromptevolution} as recent strong reflective prompt optimization baselines. Lastly, we include RL-based methods: \textbf{StablePrompt}~\citep{kwon-etal-2024-stableprompt} and \textbf{PRL}~\citep{batorski2025prlpromptsreinforcementlearning} which directly fine-tune the prompt-LM with PPO and GRPO, respectively, using accuracy-based rewards. StablePrompt uses the same meta-prompt as our method except for the reference prompts (see Appendix~\ref{appen:meta_prompt_design}), making it the most direct baseline for isolating the effect of our off-policy GFlowNet objective and DMU mechanism.

\subsection{Few-Shot Text Classification}
\label{sec:few_shot_tc}

\begin{figure}[t]
  \centering
  \includegraphics[width=0.85\linewidth]{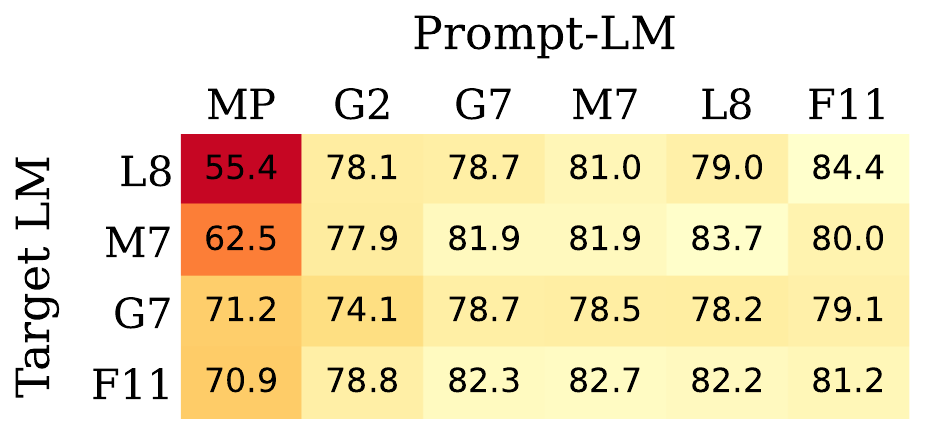}
  \vspace{-0.05in}
  \caption{Heatmap of performances on few-shot text classification tasks (6 datasets) with various prompt-LM, target LM pairs. Each number is averaged over six tasks. \textit{MP}: Manual Prompt, \textit{G2}: Gemma-2B, \textit{G7}: Gemma-7B, \textit{M7}: Mistral-7B, \textit{L8}: Llama-3-8B, \textit{F11}: Falcon-11B. 
  }
  \label{fig:heatmap}
\end{figure}

\paragraph{Datasets.} Consistent with common practice in prompt optimization research, we consider subsets of GLUE \citep{wang-etal-2018-glue} and SuperGLUE \citep{wang2020supergluestickierbenchmarkgeneralpurpose}, including sentiment analysis (SST-2) and natural language inference datasets (MRPC, MNLI, QNLI, SNLI, and RTE). During inference, we use a verbalizer that maps each answer class to a predefined label token. When determining the prediction of the target LM, we compute the probability of predefined label tokens from verbalizer and select the token that has the highest probability among them. Dataset statistics and verbalizer settings are summarized in Appendix~\ref{appen:few_shot_data_stat}.

\paragraph{Implementation details.} We consider two settings for few-shot text classification tasks. In the first setting, we experiment with both prompt-LM and target LM fixed to gemma-1.1-7B-it \citep[Gemma-7B;][]{gemmateam2024gemmaopenmodelsbased} to see how \textsc{GFlowPO} performs compared to baselines. In the second setting, we run \textsc{GFlowPO} on four target LMs: Gemma-7B, Mistral-7B-it-v2.0~\citep[Mistral-7B;][]{jiang2023mistral7b}, llama3-8B-it~\citep[Llama3-8B;][]{grattafiori2024llama3}, and falcon-11B~\citep[Falcon-11B;][]{almazrouei2023falconseriesopenlanguage}, and four prompt-LMs: gemma-1.1-2B-it (Gemma-2B), Gemma-7B, Llama3-8B, and Mistral-7B to assess on various LM pairs.

\paragraph{Results.} Table~\ref{tab:few_shot_tc_6} shows the performance of various baselines and \textsc{GFlowPO}. \textsc{GFlowPO} achieves the highest performance on SST-2, RTE, and SNLI. Overall, \textsc{GFlowPO} surpasses StablePrompt and achieves the highest average accuracy, demonstrating the effectiveness of our GFlowNet-based off-policy learning objective and the DMU mechanism. Figure~\ref{fig:heatmap} shows the performance of \textsc{GFlowPO} across various prompt-LM and target LM pairs. \textsc{GFlowPO} consistently outperforms manual prompts across all pairs, showing robustness to the choice of various (prompt-LM, target LM) pairs.

\begin{table*}[t]

\centering
\begin{sc}
\setlength{\tabcolsep}{1.5pt} 
\resizebox{\textwidth}{!}{
\begin{tabular}{@{}lcccccccccc@{}}
\toprule
 & \multicolumn{7}{c}{Instruction Induction (II)} & & \multicolumn{2}{c}{BBII} \\
\cmidrule(lr){2-8} \cmidrule(lr){10-11}
 & Spelling & Syntax & Semantics & Translation & GLUE & Others & All & & Text class. & Text gen. \\
Method & (4) & (4) & (5) & (4) & (3) & (4) & (24) & & (12) & (6) \\
\midrule
Fewshot
 & 3.8 & 0.0 & 0.6 & 0.0 & 35.0 & 0.5 & 5.2 &
 & 51.5 & 6.2 \\
Manual
 & 29.6 & 36.8 & 14.6 & 23.3 & 49.7 & 57.1 & 33.7 &
 & 51.6 & 37.6 \\
APE
 & 43.8 & 67.2 & \underline{44.5} & 35.9 & 45.9 & 74.7 & 51.9 &
 & 56.5 & 49.6 \\
ProTeGi
 & 43.8 & 68.8 & 27.5 & \textbf{56.3} & 58.3 & 62.3 & 51.6 &
 & 56.6 & 55.6 \\
EvoPrompt
 & $43.2{\scriptstyle\pm1.3}$ & $48.9{\scriptstyle\pm2.6}$ & $24.1{\scriptstyle\pm3.2}$ & $38.2{\scriptstyle\pm3.4}$ & $47.0{\scriptstyle\pm1.5}$ & $57.7{\scriptstyle\pm2.2}$ & $42.2{\scriptstyle\pm0.7}$ &
 & $\underline{57.8}{\scriptstyle\pm1.5}$ & $53.4{\scriptstyle\pm5.7}$ \\
PromptWizard
 & $46.6{\scriptstyle\pm6.4}$ & $58.1{\scriptstyle\pm3.3}$ & $25.1{\scriptstyle\pm5.4}$ & $34.3{\scriptstyle\pm6.5}$ & $48.2{\scriptstyle\pm1.6}$ & $68.0{\scriptstyle\pm6.6}$ & $45.8{\scriptstyle\pm2.0}$ &
 & $53.4{\scriptstyle\pm0.8}$ & $56.5{\scriptstyle\pm3.5}$ \\
GRACE
 & $\underline{52.2}{\scriptstyle\pm15.3}$ & $56.8{\scriptstyle\pm12.9}$ & $27.5{\scriptstyle\pm9.8}$ & $39.5{\scriptstyle\pm4.5}$ & $38.2{\scriptstyle\pm2.3}$ & $67.6{\scriptstyle\pm1.3}$ & $46.5{\scriptstyle\pm4.4}$ &
 & $53.0{\scriptstyle\pm0.4}$ & $54.4{\scriptstyle\pm4.0}$ \\
StablePrompt
 & $50.6{\scriptstyle\pm2.9}$ & $\underline{70.8}{\scriptstyle\pm1.5}$ & $40.5{\scriptstyle\pm3.8}$ & $47.0{\scriptstyle\pm1.9}$ & $\underline{59.7}{\scriptstyle\pm2.0}$ & $\underline{82.6}{\scriptstyle\pm1.2}$ & $\underline{57.7}{\scriptstyle\pm0.4}$ &
 & $57.1{\scriptstyle\pm0.6}$ & $57.8{\scriptstyle\pm1.1}$ \\
PRL
 & $35.3{\scriptstyle\pm0.3}$ & $65.0{\scriptstyle\pm6.7}$ & $36.0{\scriptstyle\pm13.4}$ & $41.3{\scriptstyle\pm3.3}$ & $48.7{\scriptstyle\pm1.7}$ & $76.9{\scriptstyle\pm5.6}$ & $53.2{\scriptstyle\pm2.4}$ &
 & $56.2{\scriptstyle\pm1.2}$ & $58.2{\scriptstyle\pm3.4}$ \\
GEPA
 & $43.2{\scriptstyle\pm2.8}$ & $69.3{\scriptstyle\pm1.1}$ & $27.6{\scriptstyle\pm5.4}$ & $46.9{\scriptstyle\pm3.4}$ & $53.4{\scriptstyle\pm4.7}$ & $65.0{\scriptstyle\pm7.6}$ & $49.8{\scriptstyle\pm2.0}$ &
 & $56.1{\scriptstyle\pm1.4}$ & $\underline{60.2}{\scriptstyle\pm0.9}$ \\
\midrule
\textbf{GFlowPO}
 & $\textbf{59.2}{\scriptstyle\pm0.6}$ & $\textbf{73.3}{\scriptstyle\pm0.3}$ & $\textbf{57.5}{\scriptstyle\pm2.4}$ & $\underline{52.6}{\scriptstyle\pm2.0}$ & $\textbf{63.8}{\scriptstyle\pm2.8}$ & $\textbf{85.0}{\scriptstyle\pm0.7}$ & $\textbf{65.0}{\scriptstyle\pm0.5}$ &
 & $\textbf{60.1}{\scriptstyle\pm0.9}$ & $\textbf{62.3}{\scriptstyle\pm1.6}$ \\
\bottomrule
\end{tabular}
}
\end{sc}
\vspace{-0.05in}
\caption{Results on instruction induction datasets.
}
\label{tab:bbh_comparison_updated}
\vspace{-3mm}
\end{table*}

\begin{table}[!htb]  \centering \footnotesize \setlength{\tabcolsep}{8pt} \begin{tabular}{lcc} \toprule Method & MMLU & OpenBookQA \\ \midrule APE & 52.1 & 70.7 \\ ProTeGi & 53.5 & 71.5 \\ EvoPrompt & 53.7 & 69.4 \\ PromptWizard & 47.3 & 61.6 \\ GRACE & 47.2 & 70.8 \\ StablePrompt & \underline{55.8} & \underline{72.2} \\ PRL & 51.8 & 70.2 \\ GEPA & \textbf{56.3} & 70.4 \\ \textbf{GFlowPO} & 55.6 & \textbf{76.2} \\ \bottomrule \end{tabular} 
\caption{Results on question answering datasets.} \label{tab:qa}\vspace{-3mm}
\end{table}
\subsection{Instruction Induction and QA}

\label{sec:induction_qa}

\paragraph{Datasets and Implementation Details.}
Beyond few-shot text classification, we further evaluate \textsc{GFlowPO} on instruction induction and question answering to validate its effectiveness across more diverse task settings.
For instruction induction, we use Instruction Induction~\citep[II;][]{honovich2022induction} and BigBench Instruction Induction~\citep[BBII;][]{zhou2023ape}, a subset of BIG-bench~\citep{Ghazal2013BigBench}, where the prompt-LM generates an instruction that describes the rule behind given input-output pairs. These benchmarks include both text classification and text generation tasks, covering spelling, syntax, semantics, translation, GLUE-style tasks, and simple rule-based reasoning. Following \citet{honovich2022induction}, we group II into six categories: \textit{Spelling}, \textit{Syntax}, \textit{Semantics}, \textit{Translation}, \textit{GLUE}, and \textit{Others}. In total, we evaluate on 24 II tasks and 18 BBII tasks. For question answering, we use MMLU~\citep{hendrycks2021measuringmassivemultitasklanguage} and OpenBookQA~\citep{mihaylov2018openbookqa}. For MMLU, we report results on 57 topics spanning STEM, Humanities, Social Science, and Others. In OpenBookQA, each question is provided with a supporting fact, which we prepend to the question. For classification and multiple-choice tasks, we use the verbalizer-based evaluation protocol described in Section~\ref{sec:few_shot_tc}; for text generation tasks, a prediction is considered correct only when the target LM output exactly matches the ground-truth tokens. We use Mistral-7B as the prompt-LM and Gemma-7B as the target LM for induction tasks, and Gemma-7B for both prompt-LM and target LM for question answering. Additional dataset details are provided in Appendix~\ref{appen:dataset_details_bbii_ii} and \ref{appen:dataset_details_mmlu}, respectively.

\paragraph{Results.}
As shown in Tables~\ref{tab:bbh_comparison_updated} and~\ref{tab:qa}, \textsc{GFlowPO} achieves strong performance across both instruction induction and question answering tasks. On BBII and II, \textsc{GFlowPO} outperforms all baselines in average accuracy, achieving the best performance in five of six II categories while remaining comparable to ProTeGi in Translation. These results indicate that \textsc{GFlowPO} is effective even for text generation tasks requiring exact token-level matching. On OpenBookQA, \textsc{GFlowPO} achieves the best performance among all baselines, and on MMLU, it performs comparably to GEPA and StablePrompt while consistently outperforming other baselines. These results demonstrate the effectiveness of \textsc{GFlowPO} on question answering tasks, with robust performance across diverse topics. Full results on BBII, II, and MMLU are provided in Appendix \ref{appen:full_results}.

\subsection{Reasoning-Centric Tasks}
\label{sec:reasoning}

\begin{table*}[t]

\centering
\begin{sc}
\setlength{\tabcolsep}{4.5pt}
\resizebox{\linewidth}{!}{
\begin{tabular}{@{}lccccccccc@{}}
\toprule
\multirow{2}{*}{Method} 
& \multirow{2}{*}{GSM8K} 
& \multirow{2}{*}{AIME-2025} 
& \multicolumn{7}{c}{BBH (5 randomly chosen tasks)} \\
\cmidrule(lr){4-10}
& & & movie rec. & object count. & tracking five & hyperbaton & causal & Avg. (5) & All. (19) \\
\midrule
EvoPrompt 
& $93.2{\scriptstyle\pm0.2}$ 
& $46.2{\scriptstyle\pm3.1}$
& $68.7{\scriptstyle\pm0.6}$ 
& $\underline{97.3}{\scriptstyle\pm0.6}$ 
& $81.7{\scriptstyle\pm1.5}$ 
& $\textbf{99.3}{\scriptstyle\pm0.6}$ 
& $\underline{59.7}{\scriptstyle\pm0.6}$ 
& $81.3$
& $\underline{80.9}{\scriptstyle\pm0.5}$ \\

PromptWizard 
& $91.4{\scriptstyle\pm1.9}$ 
& $45.1{\scriptstyle\pm2.1}$
& $67.7{\scriptstyle\pm0.6}$ 
& $95.7{\scriptstyle\pm0.6}$ 
& $79.3{\scriptstyle\pm1.5}$ 
& $97.3{\scriptstyle\pm1.2}$ 
& $58.3{\scriptstyle\pm2.3}$ 
& $79.7$
& $78.0{\scriptstyle\pm0.8}$ \\

GRACE 
& $93.3{\scriptstyle\pm0.0}$ 
& $45.1{\scriptstyle\pm2.8}$
& $66.7{\scriptstyle\pm1.5}$ 
& $96.0{\scriptstyle\pm0.0}$ 
& $83.3{\scriptstyle\pm12.1}$ 
& $97.3{\scriptstyle\pm1.5}$ 
& $56.0{\scriptstyle\pm2.7}$ 
& $79.9$
& $79.4{\scriptstyle\pm0.7}$ \\

StablePrompt 
& $87.3{\scriptstyle\pm6.0}$ 
& $40.2{\scriptstyle\pm3.7}$
& $66.3{\scriptstyle\pm4.2}$ 
& $92.3{\scriptstyle\pm5.5}$ 
& $71.3{\scriptstyle\pm9.8}$ 
& $98.3{\scriptstyle\pm1.2}$ 
& $59.0{\scriptstyle\pm1.7}$ 
& $77.4$
& $76.8{\scriptstyle\pm0.3}$ \\

PRL 
& $92.2{\scriptstyle\pm1.2}$ 
& $\underline{46.9}{\scriptstyle\pm1.9}$
& $67.3{\scriptstyle\pm0.6}$ 
& $95.7{\scriptstyle\pm3.5}$ 
& $70.3{\scriptstyle\pm3.1}$ 
& $96.3{\scriptstyle\pm2.5}$ 
& $58.7{\scriptstyle\pm4.2}$ 
& $77.7$
& $76.4{\scriptstyle\pm1.4}$ \\

GEPA 
& $\underline{93.5}{\scriptstyle\pm0.9}$ 
& $42.9{\scriptstyle\pm1.9}$
& $\underline{70.3}{\scriptstyle\pm1.2}$ 
& $97.0{\scriptstyle\pm0.0}$ 
& $\textbf{90.0}{\scriptstyle\pm1.0}$ 
& $98.3{\scriptstyle\pm0.6}$ 
& $57.3{\scriptstyle\pm0.6}$ 
& $\underline{82.6}$
& $80.1{\scriptstyle\pm1.0}$ \\

\textbf{GFlowPO} 
& $\textbf{94.2}{\scriptstyle\pm0.2}$ 
& $\textbf{50.5}{\scriptstyle\pm3.8}$
& $\textbf{79.0}{\scriptstyle\pm5.2}$ 
& $\textbf{97.7}{\scriptstyle\pm2.1}$ 
& $\underline{83.7}{\scriptstyle\pm2.1}$ 
& $\underline{99.0}{\scriptstyle\pm0.0}$ 
& $\textbf{61.3}{\scriptstyle\pm0.6}$ 
& $\textbf{84.1}$
& $\textbf{82.8}{\scriptstyle\pm0.7}$ \\
\bottomrule
\end{tabular}
}
\end{sc}
\vspace{-0.05in}
\caption{Results on GSM8K, AIME, and BBH.}
\label{tab:reasoning}
\vspace{-2mm}
\end{table*}

\begin{figure*}[t]
    \centering
    \includegraphics[width=\textwidth]{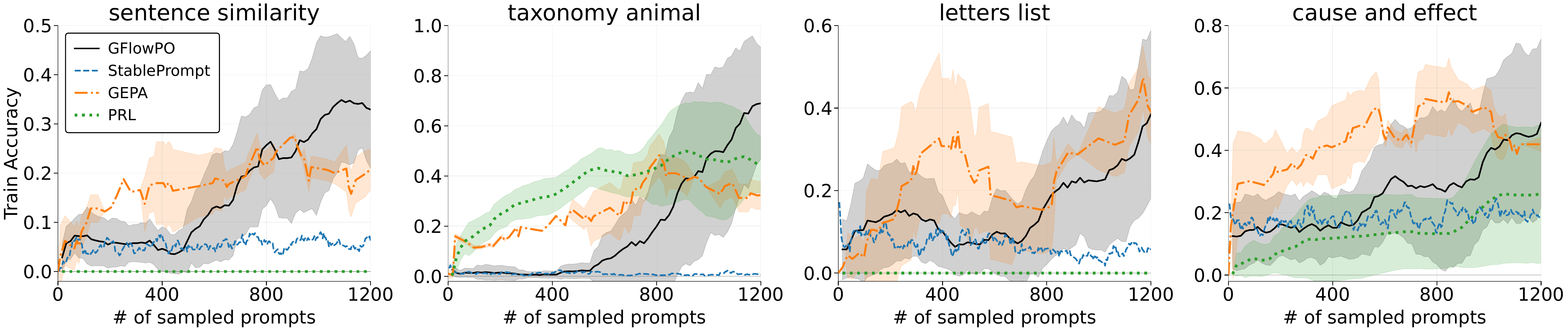}
    \vspace{-0.25in}
    \caption{
        Training accuracy as a function of the number of sampled prompts on four II tasks. The first 400 evaluations of \textsc{GFlowPO} correspond to the Pre-Step, where the prompt-LM is not updated (see Appendix \ref{appen:early_stage_comparison_gepa}).
    }
    \label{fig:gfnpo_exploration}
    \vspace{-3mm}
\end{figure*}

\paragraph{Datasets and Implementation Details.}
To further evaluate \textsc{GFlowPO} in challenging reasoning-centric benchmarks, we conduct experiments on Big-Bench Hard (BBH)~\citep{suzgun2022challenging}, GSM8K~\citep{cobbe2021GSM8Kdataset}, and AIME~\citep{balunovic_srimatharena_2025}. These benchmarks cover diverse reasoning abilities, ranging from symbolic and linguistic reasoning in BBH to grade-school and competition-level mathematical problem solving in GSM8K and AIME. For prompt evaluation, we first generate reasoning texts from the target LM, and then use an extractor phrase (e.g., ``The final answer is ...'') to obtain the final answer, following a standard evaluation protocol for reasoning tasks. We use Qwen3-8B~\citep{yang2025qwen3technicalreport} as both the prompt-LM and target LM for BBH, and Qwen3-4B-Instruct-2507 for GSM8K and AIME. Dataset details are provided in Appendix~\ref{appen:dataset_details_reasoning}.
\paragraph{Results.}
As shown in Table~\ref{tab:reasoning}, \textsc{GFlowPO} achieves the best performance across all reasoning-centric benchmarks. On BBH, \textsc{GFlowPO} outperforms all baselines, including strong recent prompt optimization methods such as GEPA and EvoPrompt, suggesting that GFlowNet-based posterior inference is effective for discovering high-reward prompts in challenging reasoning settings. 
On GSM8K and AIME, \textsc{GFlowPO} again achieves the highest average accuracy, demonstrating its effectiveness on both grade-school and competition-level mathematical reasoning tasks.
We also observe that PromptWizard and GRACE often generate substantially longer prompts, which appear to be less effective with relatively small target LMs in our setting. Overall, these results show that \textsc{GFlowPO} provides robust gains across diverse reasoning-centric tasks. The full experimental results and learning curves on BBH tasks are provided in Appendix \ref{appen:full_results} and \ref{appen:learning_curves_bbh}, respectively.

\subsection{Other Analysis}
\label{sec:ablation_study_exploration}

\paragraph{Analysis on exploration ability.}

We next analyze whether \textsc{GFlowPO} can discover high-reward prompts more sample-efficiently than strong prompt optimization baselines. We conduct experiments on four II tasks and evaluate the average training accuracy of prompts sampled throughout optimization under the same prompt-sampling budget. We compare \textsc{GFlowPO} with RL-based methods, StablePrompt and PRL, as well as GEPA, which shows strong sample efficiency. All other experimental settings are identical to those described in Section~\ref{sec:induction_qa}. 

As shown in Figure~\ref{fig:gfnpo_exploration}, \textsc{GFlowPO} generally discovers higher-reward prompts under the same number of sampled prompts. Compared to RL-based baselines such as StablePrompt and PRL, \textsc{GFlowPO} achieves better sample efficiency, suggesting that off-policy GFlowNet replay training is effective for exploring the combinatorial prompt space. GEPA often finds reasonably strong prompts early in optimization and can be competitive on some tasks, but tends to plateau with further sampling. In contrast, \textsc{GFlowPO} continues to discover prompts with higher reward as sampling progresses. Overall, these results demonstrate that \textsc{GFlowPO} provides more sample-efficient exploration than existing prompt optimization baselines.

\begin{table}[t]

\centering
\begin{footnotesize}
\begin{sc}
\setlength{\tabcolsep}{5pt}
\begin{tabular}{@{}lcccc@{}}
\toprule
\textsc{GFlowPO} variants 
& \textit{-on-X} & \textit{-off-X} & \textit{-on-O} & \textit{-off-O} \\
\midrule
Avg. Acc. 
& 74.4 & 75.2 & 78.0 & \textbf{80.2} \\
\bottomrule
\end{tabular}
\end{sc}
\caption{Ablation on \textsc{GFlowPO} variants averaged over 10 II text generation tasks. \textit{on/off}: on-/off-policy training; \textit{O/X}: with/without DMU.}
\label{tab:gflowpo_variants}
\end{footnotesize}
\vspace{-3mm}
\end{table}

\paragraph{Ablation study.}
We conduct an ablation study to examine the contribution of individual components in \textsc{GFlowPO}. Specifically, we consider four variants: (1) \textsc{GFlowPO}\textit{-on-X}, (2) \textsc{GFlowPO}\textit{-off-X}, (3) \textsc{GFlowPO}\textit{-on-O}, and (4) \textsc{GFlowPO}\textit{-off-O} (i.e., \textsc{GFlowPO}). The \textit{on} and \textit{off} variants correspond to on-policy and off-policy training, respectively. The \textit{O} and \textit{X} suffixes indicate whether the DMU mechanism is enabled or disabled. This design allows us to isolate the effects of off-policy training and the DMU mechanism in \textsc{GFlowPO}. For evaluation, we randomly select 10 text generation tasks from Instruction Induction (II) and use the same prompt-LM, and target LM from the experimental setup in Section~\ref{sec:induction_qa}.

Table~\ref{tab:gflowpo_variants} shows that incorporating each component of \textsc{GFlowPO} leads to consistent performance improvements across variants. Introducing off-policy training to \textsc{GFlowPO}\textit{-on-X} yields a modest gain, which can be attributed to improved sample efficiency from replay-based training. Enabling the DMU mechanism (\textsc{GFlowPO}\textit{-on-O}) results in a more substantial improvement, reflecting the benefit of adaptively reshaping the meta-prompt to focus the search on high-reward regions of the prompt space. When both components are combined (\textsc{GFlowPO}\textit{-off-O}), performance further improves and achieves the best overall results. These results indicate that off-policy training and DMU contribute complementarily, rather than simply providing redundant gains. We additionally provide the complementary property of GFlowNet-specific learning and DMU in Appendix~\ref{appen:gfn_contribution}.

\section{Conclusion}
\label{sec:conclusion}
In this work, we proposed \textsc{GFlowPO}, a novel prompt optimization framework that formulates prompt search as posterior inference and solves it using off-policy GFlowNet training combined with Dynamic Memory Update (DMU). With these two complementary components, \textsc{GFlowPO} enables efficient exploration of the discrete prompt space while progressively reshaping the search focus on high-reward regions. Empirically, \textsc{GFlowPO} achieves strong and consistent improvements over existing baselines across few-shot text classification, instruction induction, question answering, and reasoning-centric benchmarks, demonstrating robustness across diverse tasks. 

\section*{Limitations}
While \textsc{GFlowPO} demonstrates strong prompt optimization performance across diverse tasks, it currently optimizes prompts under a fixed inference budget. Incorporating test-time compute strategies such as self-consistency or search may further improve performance. In addition, the current approach remains task-specific, and extending \textsc{GFlowPO} to a meta-learning setting that amortizes prompt optimization across tasks and generalizes to unseen tasks is an important direction for future work.

\section*{Ethical Considerations}
\label{appen:broader_impact}
This paper presents a method for improving prompt optimization by combining off-policy GFlowNet training with dynamic meta-prompt updates. The goal of this work is to advance the efficiency and robustness of prompt-based adaptation for large language models under limited supervision. While improved prompt optimization may indirectly affect downstream applications of language models, such as decision support or content generation, these impacts are consistent with those commonly associated with advances in machine learning research. We do not foresee any unique ethical concerns arising from this work beyond those already well studied in the deployment of large language models.

\bibliography{custom}

\clearpage
\appendix
\label{sec:appendices}

\section{Extended Related Work}
\label{append:related_acl}
\paragraph{Prompt optimization.}
Prompt optimization (or prompt tuning) has been explored in previous works as task-specific continuous vectors tuned by gradient-based methods to improve task performance~\citep{li2021prefixtuningoptimizingcontinuousprompts, lester2021powerscaleparameterefficientprompt, Qin_2021}. Discrete prompts, on the other hand, involve searching for discrete vocabulary tokens through gradients~\citep{shin2020autopromptelicitingknowledgelanguage, shi2022humanreadableprompttuning, das2025greatergradientsreasoningmakes}. 
A complementary line of work formulates prompt optimization as an RL problem. Early methods such as RLPrompt~\citep{deng2022rlprompt} and TEMPERA~\citep{zhang2023tempera} optimize discrete prompts or prompt-editing policies using task rewards, but rely on small policy networks or edit-based action spaces. More recent methods scale this direction to LLM-based policies: StablePrompt trains a prompt-LM with PPO-style updates for prompt tuning~\citep{kwon-etal-2024-stableprompt}, and PRL~\citep{batorski2025prlpromptsreinforcementlearning} explores GRPO-style optimization for automatic generation of both prompt and few-shot examples. Compared to these methods, \textsc{GFlowPO} uses off-policy GFlowNet and a training-free Dynamic Memory Update (DMU) that injects diverse and high-reward prompts into the meta-prompt, enabling more sample-efficient exploration in combinatorial prompt space. 

\paragraph{LLMs as prompt optimizers.}
Early work such as APE \citep{zhou2023ape} demonstrated that LLMs can be directly used as prompt optimizers by inferring instructions from input--output pairs, while subsequent methods explored textual gradients \citep{pryzant2023protegi}, optimization in text space \citep{yang2024opro, yuksekgonul2025optimizing}, evolutionary search~\citep{guo2025evopromptconnectingllmsevolutionary}, meta-prompt engineering \citep{ye-etal-2024-prompt}, agent-based reasoning \citep{wang2023promptagentstrategicplanninglanguage, shinn2023reflexion}, Bayesian optimization \citep{agarwal2025searchingoptimalsolutionsllmsbayesianoptimization, liu2024llambo}, RL~\citep{dong2024pace}, and LLM-based feedback~\citep{agarwal-etal-2025-promptwizard, shi2025lossgaingatedrefinement, agrawal2026gepareflectivepromptevolution}.
Most existing methods rely on powerful, closed-source LLMs, since prompt optimization with lightweight models often leads to degraded performance \citep{zhang2024revisitingoprolimitationssmallscale}. In contrast, our method directly fine-tunes a lightweight, open-source prompt-LM as the prompt optimizer without relying on heavy external LLMs. 

\paragraph{Generative Flow Networks.}
Generative Flow Networks~\citep[GFlowNets;][]{bengio2021flownetworkbasedgenerative, JMLR:v24:22-0364} are a family of reinforcement learning methods for performing amortized inference, sampling proportionally to an unnormalized density or reward. GFlowNets model trajectories from an initial state to a terminal state in a step-by-step manner, where the terminal state represents a compositional objective such as a graph or a string. Such sequential decision-making policies and value functions (i.e., flows) are parameterized by deep networks and learned via constraint-based objectives. Based on the level of constraint coverage, representative objectives include detailed balance~\citep[DB;][]{JMLR:v24:22-0364} and flow matching~\citep{bengio2021flownetworkbasedgenerative}, which enforce local one-step transition constraints; sub-trajectory balance~\citep[SubTB;][]{madan2023learning}, which matches multi-step transition constraints; and trajectory balance~\citep[TB;][]{malkin2023trajectorybalanceimprovedcredit}, which enforces global constraints over full trajectories. \citet{zhang2023robust} proposed variants of TB in the form of log-partition gradients~\citep[VarGrad;][]{richter2020vargradlowvariancegradientestimator}, which estimate the partition flow from minibatch data rather than learning it directly.

\paragraph{Language model post-training with GFlowNets.} The amortized inference capability of GFlowNets has been applied to language model post-training, as it is beneficial for promoting diversity and enabling efficient off-policy sampling. \citet{hu2024amortizingintractableinferencelarge} applies GFlowNets with expectation–maximization~\citep[GFlowEM;][]{hu2023gflownetem} to infilling tasks and chain-of-thought reasoning. \citet{yu2025flow} applied GFlowNets to mathematical and puzzle reasoning using an off-policy local search method~\citep{zhang2022ebgfn, kim2024local}, while \citet{bartoldson2025trajectorybalanceasynchronydecoupling} applied decentralized GFlowNets for faster reasoning-LM training. \citet{lee2025learningdiverseattackslarge} applied GFlowNets to LM red-teaming with a focus on diversity, which \citet{yun2025active} further extends to an active learning-based red-teaming framework. \citet{yun2025learning} applied GFlowNets with local credit assignment~\citep{pan2023better} for diverse prompt generation into text-to-image model. Our work is also an LM post-training application of GFlowNets, but to the best of our knowledge, it is the first approach to use in-context memory updates in both the prior and posterior distributions.

\section{\textsc{GFlowPO} Full Algorithm}
\label{appen:full_algorithm}
We provide the full learning algorithm of \textsc{GFlowPO} in \Cref{alg:gfnpo_full}.

\begin{algorithm*}[!htb]
  \caption{\textsc{GFlowPO} Full Algorithm}
  \label{alg:gfnpo_full}
  \begin{algorithmic}[1]
    \STATE {\bfseries Input:} Prompt-LM $p_\theta$, small training dataset $\mathcal{D}$, learning rate $\alpha$, pre-step size $n$, batch size $m$, training steps $T$, meta-prompt $M$, set of reference prompts $\mathcal{Z}_\text{ref}$, high-reward buffer $\mathcal{Q}$, replay buffer $\mathcal{B}$, update frequency $K$, number of reference prompts sampled from the replay buffer $k_b$, number of reference prompts sampled from the high-reward buffer $k_q$. 
    \STATE $p_{\text{ref}} \leftarrow \text{deepcopy}(p_\theta)$, $\mathcal{Q} \leftarrow \emptyset$, $\mathcal{B} \leftarrow \emptyset$, $\mathcal{Z}_\text{ref} \leftarrow \emptyset$, $\ell \leftarrow 0$
    \FOR[\textcolor{blue}{Pre-Step: Replay buffer initialization}]{$i = 1, \ldots, n$}
        \STATE $\tau \leftarrow \text{Uniform}(0.5, 2.0)$
        \STATE Sample $z$ from $p_\theta(z|M)$  with temperature $\tau$
        \STATE $\mathcal{B}.\text{Add}\left(z, A_\mathcal{D}(z)\right)$
    \ENDFOR
    
    \FOR{$t = 1, \ldots, T$}  
        \FOR[\textcolor{blue}{GFlowNet fine-tuning}]{$i = 1, \ldots, m$}
            \STATE Sample $s$ from $\{0,1\}$
            \STATE $\tau \leftarrow \text{Uniform}(0.5, 2.0)$
            \IF{$s=0$}
                \STATE Sample $z$ from $p_\theta(z|M)$ with temperature $\tau$ 
                \STATE $\mathcal{B}.\text{Add}\left(z,A_\mathcal{D}(z)\right)$
                \STATE $\mathcal{Q}.\text{Add}\left(z,A_\mathcal{D}(z)\right)$
            \ELSE
                \STATE $\left(z,A_\mathcal{D}(z)\right) \sim \mathcal{B}$
            \ENDIF
            \STATE Compute reward $R(z;M)=A_\mathcal{D}(z) \cdot p_\text{ref}(z\mid M)$ in \Cref{eq:reward}
            \STATE Compute the loss $\ell \leftarrow \ell + \mathcal{L}(\theta;M)/m$ using \Cref{eqn:loss}
        \ENDFOR
        \STATE Update $\theta \leftarrow \theta - \alpha \nabla_\theta \ell$ and reset $\ell \leftarrow 0$
        \IF[\textcolor{blue}{Dynamic Memory Update~(DMU)}]{$t \bmod K = 0$}
            \STATE Sample $\{z_b^{(i)}\}_{i=1}^{k_b}$ from $\mathcal{B}$
            \STATE Sample $\{z_q^{(j)}\}_{j=1}^{k_q}$ from $\mathcal{Q}$ 
            \STATE Update $\mathcal{Z}_\text{ref} \leftarrow \{z_b^{(i)}\}_{i=1}^{k_b} \cup \{z_q^{(j)}\}_{j=1}^{k_q}$
            \STATE Update $M$ with $\mathcal{Z}_\text{ref}$ as in \Cref{sec:m_step_dmu}
        \ENDIF
    \ENDFOR
    \STATE {\bfseries Return:} Prompt-LM $p_\theta$
  \end{algorithmic}
\end{algorithm*}

\section{Correlation between Likelihood and Accuracy}
\label{appen:correlation}

\begin{figure*}[ht] 
    \centering
    \includegraphics[width=0.81\textwidth, keepaspectratio]{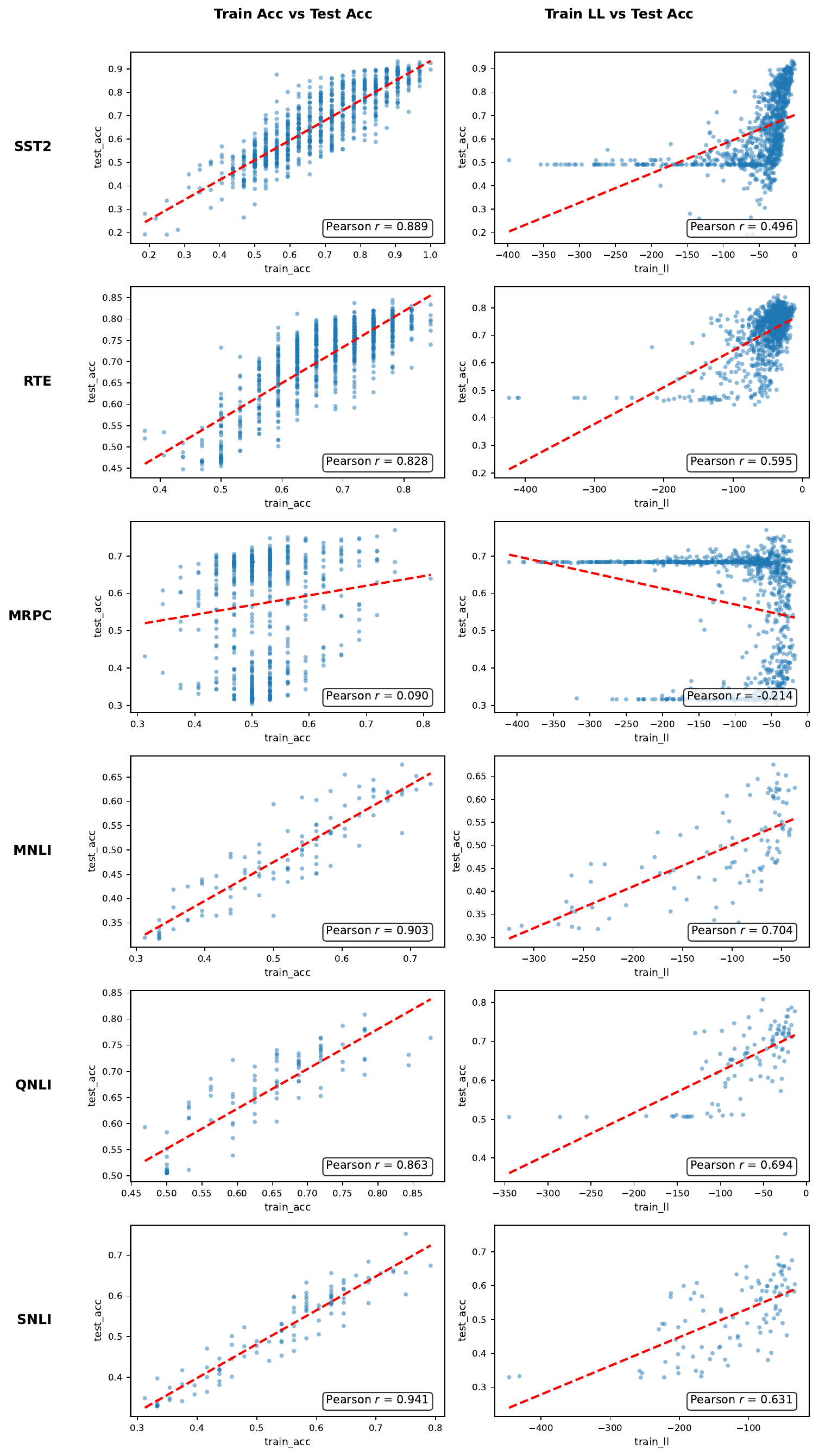}
    \vspace{-0.1in}
    \caption{Correlation plots on train accuracy vs test accuracy (\textbf{left}), and train log-likelihood vs test accuracy (\textbf{right}) on six text classification tasks (SST2, RTE, MRPC, MNLI, QNLI, and SNLI).}
    \label{fig:appen_corr}
\end{figure*}

In this work, we define an unnormalized target reward as
$R(z; M) = p(\mathcal{D} \mid z) \cdot p_{\text{ref}}(z \mid M)$.
Here, the training log-likelihood $\log p(\mathcal{D} \mid z)$ can be computed by summing the log-probabilities of the ground-truth output tokens $y$ from target LM given prompt $z$ and context $x$ over all $(x, y)$ pairs in $\mathcal{D}_{\text{task}}$. However, we empirically observe that the training likelihood $p(\mathcal{D} \mid z)$ is weakly correlated with test accuracy for most prompt optimization tasks we consider. To verify this, we sample 1,000 prompts using the meta-prompt from StablePrompt (see \Cref{fig:appen_metaprompt}\textcolor{Red}{(a)}) across six few-shot text classification tasks, and compute the training log-likelihood $\log p(\mathcal{D} \mid z)$, training accuracy $A_{\mathcal{D}}(z)$ (used as the learning objective in this work), and test accuracy for each prompt. We then measure the correlation between training accuracy and test accuracy, as well as between training log-likelihood and test accuracy. As shown in \Cref{fig:appen_corr}, the correlation between training log-likelihood and test accuracy is consistently lower than that between training accuracy and test accuracy across all six tasks. Therefore, we directly use training accuracy as the optimization objective in this work.

\makeatletter
\newpage
\if@firstcolumn\else
  \null\newpage
\fi
\makeatother

\section{Effect of Collecting Prompts in Replay Buffer before GFlowNet Training}
\label{appen:pre_step}
One strength of \textsc{GFlowPO} is its off-policy learning strategy, which allows sampled prompts from the prompt-LM to be stored in a replay buffer and reused for training. This off-policy setting also enables collecting diverse prompts with the prompt-LM and storing them in the replay buffer before the actual GFlowNet training starts. As a result, the replay buffer may already contain high-reward prompts at initialization, which can facilitate exploration in the early stage of training. We refer to this procedure as \textit{Pre-Step}. Details of Pre-Step and the overall training algorithm are provided in \Cref{alg:gfnpo_full}.

To examine whether Pre-Step improves performance, we compare \textsc{GFlowPO} with \textsc{GFlowPO w/o Pre-Step}. For a fair comparison, \textsc{GFlowPO w/o Pre-Step} is trained for the same total number of sampled prompts as \textsc{GFlowPO}. All remaining experimental settings are kept the same as in \Cref{sec:induction_qa}. As shown in \Cref{tab:prestep_reference_prompt_sampling_ablation}, \textsc{GFlowPO} achieves higher average accuracy than \textsc{GFlowPO w/o Pre-Step}, indicating that Pre-Step is beneficial.

\begin{table}[!htb]

\centering
\normalsize
\begin{sc}
\setlength{\tabcolsep}{3pt}
\vspace{0.5em}
\resizebox{\columnwidth}{!}{%
\begin{tabular}{@{}lcccc@{}}
\toprule
Task & \shortstack{GFlowPO\\w/o pre-step} & \shortstack{GFlowPO\\-$\mathcal{B}$} & \shortstack{GFlowPO\\-$\mathcal{Q}$} & GFlowPO \\
\midrule
Cause and effect    & \underline{80.0} & 76.0             & 52.0             & \textbf{88.0} \\
Negation            & 83.0             & \textbf{89.0}    & 86.0             & \underline{87.0} \\
Letters list        & \underline{94.0} & \underline{94.0} & \textbf{96.0}    & \textbf{96.0} \\
Larger animal       & 56.0             & \underline{91.0} & 90.0             & \textbf{93.0} \\
Sentence similarity & \textbf{39.0}    & \underline{36.0} & \underline{36.0} & \textbf{39.0} \\
Num to verbal       & 95.0             & 94.0             & \textbf{100.0}   & \underline{98.0} \\
Word in context     & 62.0             & \underline{63.0} & 60.0             & \textbf{64.0} \\
Sentiment           & \underline{90.0} & 89.0             & 88.0             & \textbf{91.0} \\
Informal to formal  & 45.7             & \textbf{52.7}    & 38.8             & \underline{49.2} \\
Singular to plural  & \underline{98.0} & \textbf{99.0}    & \underline{98.0} & 97.0 \\
\midrule
Average (10 tasks)  & 74.3             & \underline{78.4} & 74.5             & \textbf{80.2} \\
\bottomrule
\end{tabular}%
}
\end{sc}
\caption{Comparison of \textsc{GFlowPO} variants. \textsc{GFlowPO w/o Pre-Step} removes the Pre-Step procedure. \textsc{GFlowPO-$\mathcal{B}$} samples reference prompts only from the replay buffer $\mathcal{B}$, while \textsc{GFlowPO-$\mathcal{Q}$} samples reference prompts only from the high-reward buffer $\mathcal{Q}$. \textsc{GFlowPO} achieves the best performance across all variants.}
\label{tab:prestep_reference_prompt_sampling_ablation}
\end{table}

\section{Effect of Sampling Reference Prompts on both Replay Buffer and High-Reward Buffer}
\label{appen:ref_sampling}

In \textsc{GFlowPO}, reference prompts $\mathcal{Z}_{\text{ref}}$ in the meta-prompt $M$ are sampled from both the replay buffer $\mathcal{B}$ and the high-reward buffer $\mathcal{Q}$ to balance exploration and exploitation. In this section, we study the effect of sampling reference prompts exclusively from either $\mathcal{B}$ or $\mathcal{Q}$. An example of a meta-prompt that samples reference prompts only from the high-reward buffer $\mathcal{Q}$ is shown in Figure \ref{fig:appen_metaprompt}\textcolor{Red}{(b)}. 

Specifically, we compare \textsc{GFlowPO} with two variants: \textsc{GFlowPO-$\mathcal{B}$}, which samples reference prompts only from the replay buffer $\mathcal{B}$, and \textsc{GFlowPO-$\mathcal{Q}$}, which samples reference prompts only from the high-reward buffer $\mathcal{Q}$. We use the same experimental settings as in \Cref{sec:induction_qa}. As reported in \Cref{tab:prestep_reference_prompt_sampling_ablation}, \textsc{GFlowPO} outperforms both variants, suggesting that balancing exploration and exploitation is important for performance. Sensitivity to k is further analyzed in Appendix \ref{appen:dmu_sensitivity}.

\section{GFlowPO Hyperparameter Setting}
\label{appen:hparam}
Hyperparameter used in \textsc{GFlowPO} are presented in \Cref{tab:hyperparameters}.
\begin{table}[!t]
\centering
\normalsize
\begin{sc}
\begin{tabular}{@{}lc@{}}
\toprule
Hyperparameter & Value \\
\midrule
Learning rate           & $10^{-4}$ \\
Train buffer size       & 1000 \\
High-reward buffer size & 5 \\
Total train steps       & 200 \\
Pre-steps               & 100 \\
Max prompt length       & 150 \\
Num example             & 5 \\
LoRA $r$                & 16 \\
LoRA $\alpha$           & 32 \\
LoRA dropout            & 0.05 \\
EMA decay               & 0.99 \\
Sampling temperature    & $[0.5, 2.0]$ \\
Online/offline ratio    & 0.5 \\
M-step frequency        & 1 \\
\bottomrule
\end{tabular}
\end{sc}
\caption{Hyperparameters of \textsc{GFlowPO}.}
\label{tab:hyperparameters}
\end{table}

\section{Generated Prompts}
\label{appen:generated_prompts}
In this section, we provide the prompts generated by \textsc{GFlowPO}: few-shot text classification in \ref{appen:few_shot_tc_gen_prompts}, BigBench-Hard Instruction Induction in \ref{appen:bbh-ii-cls_gen_prompts}--\ref{appen:bbh-ii-gen_gen_prompts}, Instruction Induction in \ref{appen:ii_gen_prompts}, BBH in \ref{appen:bbh_gen_prompts}, and AIME in \ref{appen:aime_gen_prompts}. For each task, we report the prompt selected from the best-performing prompts.
\vspace*{\fill}

\subsection{Few-Shot Text Classification}
\label{appen:few_shot_tc_gen_prompts}
\begin{promptbox}[lightgray]{SST2}
    \small
    Write yes or no for each input, indicating whether the statement is likely to result in a good movie. **Input:** too bad maggio couldn't come up with a better script. **Output:** no **Input:** mostly leaves him shooting blanks **Output:** no **Input:** is sandler running on empty , repeating what he 's already done way too often **Output:** no **Input:** sit through than this hastily dubbed disaster **Output:** no **Input:** , it manages to maintain both a level of sophisticated intrigue and human-scale characters that suck the audience in. **Output:** yes
    
    \vspace{0.5em}
    \noindent\makebox[\linewidth]{\color{blue}\dotfill}
    \par\vspace{0.1em}
    \centering
    Accuracy: 92.8\%
\end{promptbox}

\begin{promptbox}[lightgray]{MRPC}
    \small
    **For each input, determine whether the two sentences are semantically equivalent or not.** **Criteria for equivalence:** - The sentences convey the same meaning and information. - The grammatical structure and word choice are similar. **Results:** **1. Sentence 1 \& 2:** - Both sentences express that Sarah O'Hare thanked the matriarch for her generosity. - Grammatical structure and word choice are similar. **Output: Yes** **2. Sentence 1 \& 2:** - Both sentences describe Lee commenting on Brianna's behavior. - Similar grammatical structure and word choice. **Output: Yes** **3. Sentence 1 \& 2:** - Sentence
    
    \vspace{0.5em}
    \noindent\makebox[\linewidth]{\color{blue}\dotfill}
    \par\vspace{0.1em}
    \centering
    Accuracy: 67.2\%
\end{promptbox}

\begin{promptbox}[lightgray]{RTE}
    \small
	1. For each premise, assess whether the hypothesis is true or false. 2. For each of the given premises, generate a logical output related to the given hypothesis. **Results:** **1. Premise:** In 1969, he drew up the report proposing the expulsion from the party of the Manifesto group. **Hypothesis:** Natta supported the Manifesto group. **Output:** No **2. Premise:** The increased amounts of carbon dioxide (CO2) and other greenhouse gases (GHGs) are the primary causes of the human-induced component of global warming. **Hypothesis:** Greenhouse effect changes global climate. **Output:** Yes **3. Premise:** Sida does
    
    \vspace{0.5em}
    \noindent\makebox[\linewidth]{\color{blue}\dotfill}
    \par\vspace{0.1em}
    \centering
    Accuracy: 79.1\%
\end{promptbox}

\begin{promptbox}[lightgray]{QNLI}
    \small
    **1. For each input, write an output indicating whether the sentence mentions the given question.** **Results:** - Question: How many territories is the Premier League broadcast to? Sentence: The Premier League rank second in the UEFA coefficients of leagues based on performances in European competitions over the past five seasons. Output: No - Question: What regime were Hussein loyalists part of? Sentence: Other elements of the insurgency were led by fugitive members of President Hussein's Ba'ath regime, which included Iraqi nationalists and pan-Arabists. Output: Yes - Question: If Neptune formed closer to the sun, what caused it to migrate to it's current orbit? Sentence: The provided text does not contain information regarding Neptune's
    
    \vspace{0.5em}
    \noindent\makebox[\linewidth]{\color{blue}\dotfill}
    \par\vspace{0.1em}
    \centering
    Accuracy: 84.1\%
\end{promptbox}

\begin{promptbox}[lightgray]{MNLI}
    \small
    1. For each of the given premise-hypothesis pairs, write an output. 2. For each premise, write an output based on the hypothesis provided. 3. Based on the given premise, write an output for the given hypothesis. **Results:** **Premise-Hypothesis Pairs Output:** - Premise: And other stars probably have planets. Hypothesis: Other stars are made of ice. Output: No - Premise: No doubt he will do better in his next book. Hypothesis: He will probably do even worse in his future works. Output: No - Premise: The man gives good movies. Hypothesis: The man gives a good movie but still can't find success. Output: Maybe **For
    
    \vspace{0.5em}
    \noindent\makebox[\linewidth]{\color{blue}\dotfill}
    \par\vspace{0.1em}
    \centering
    Accuracy: 70.5\%
\end{promptbox}

\begin{promptbox}[lightgray]{SNLI}
    \small
    **For each premise, write an output based on the hypothesis.** **The outputs are as follows:** **1.**Premise: People attending a concert dressed in yellow. Hypothesis: People are wearing yellow. Output: Yes **2.**Premise: Newly married couple with money attached to their clothing. Hypothesis: Newlyweds pin money to their clothing to represent good fortune. Output: Maybe **3.**Premise: A man in a green coat crossing the street with graffiti in the background. Hypothesis: The man has a green coat. Output: Yes **4.**Premise: Children play soccer on a narrow city street. Hypothesis: Children are playing soccer. Output: Yes **5.**Premise:
    
    \vspace{0.5em}
    \noindent\makebox[\linewidth]{\color{blue}\dotfill}
    \par\vspace{0.1em}
    \centering
    Accuracy: 81.5\%
\end{promptbox}

\subsection{BigBench Instruction Induction (BBII) - Text Classification}
\label{appen:bbh-ii-cls_gen_prompts}
We randomly choose five tasks from the text classification tasks of BBII dataset.

\begin{promptbox}[brown]{Implicatures}
    \small
    Based on the conversation between Speaker 1 and Speaker 2, choose the answer that best fits the situation. The Output for each Input is: 1. Input: Speaker 1: 'I feel horrible. Debbie was furious that I lost her notes. Do you think I should apologize her again?' Speaker 2: 'If I were you I would cool off for some days before I talk to her again.' Choices: A : yes, B : no Output: B Explanation: Speaker 2 advised Speaker 1 to wait before talking to Debbie again, indicating that it might not be the best time for an apology. 2	
    
    \vspace{0.5em}
    \noindent\makebox[\linewidth]{\color{blue}\dotfill}
    \par\vspace{0.1em}
    \centering
    Accuracy: 86.3\%
\end{promptbox}

\begin{promptbox}[brown]{Movie Recommendation}
    \small
    2. Based on the given list of movies, choose an alternative movie from the provided choices that is not in the given input but shares some similarity with the given movies in terms of genre or theme. Your friend seemed to have misunderstood the instruction and instead chose a movie from the input for each set of choices. The output matches the movies in the choices, although not necessarily the movies in the input that they were asked to find an alternative for. To provide the correct answer for this instruction, we should consider the genre or theme of the given movies and find an alternative movie from the choices that shares some similarity with them. For example, if the given movies are: The Godfather, The Dark Knight,
    
    \vspace{0.5em}
    \noindent\makebox[\linewidth]{\color{blue}\dotfill}
    \par\vspace{0.1em}
    \centering
    Accuracy: 70.6\%
\end{promptbox}

\begin{promptbox}[brown]{Snarks}
    \small
	"Choose the answer that is the opposite of the given input." Based on this instruction, the incorrect outputs by your friend for the third and fourth pairs: Input: (a) I'm sure you wouldn't have remained completely calm if a teacher attacked you. Choices: A: (a) (input without the given phrase) B: (b) (input with the given phrase) Output: B (correction: The output should be 'A' as the opposite of (b) is 'You're sure you would have remained completely calm if a teacher attacked you' or more literally 'I'm sure you would have Panicked.') Input: (a) College
    \vspace{0.5em}
    \noindent\makebox[\linewidth]{\color{blue}\dotfill}
    \par\vspace{0.1em}
    \centering
    Accuracy: 65.6\%
\end{promptbox}

\begin{promptbox}[brown]{Sports Understanding}
    \small
	"Given the following inputs, choose whether each is plausible or implausible based on your knowledge of baseball and soccer." The friend's outputs for each input are: Input: Alex Pietrangelo was called for slashing in the Stanley Cup [Choices]: A: plausible B: implausible Output: A (Plausible as slashing is a penalty called in ice hockey) Input: Wayne Rooney shot from outside the eighteen [Choices]: A: plausible B: implausible Output: A (Plausible as soccer players, including Wayne Rooney, do shoot from outside the eighteen-yard box)
    \vspace{0.5em}
    \noindent\makebox[\linewidth]{\color{blue}\dotfill}
    \par\vspace{0.1em}
    \centering
    Accuracy: 67.2\%
\end{promptbox}

\begin{promptbox}[brown]{presuppositions as nli}
    \small
1. "Based on the given context, determine if the relationship between the two sentences is entailment, neutral, or contradiction." The Outputs are correct for all the given Inputs. Input 1, 2, and 3 are neutral because the second sentence does not follow or contradict the first sentence but also doesn't entail it. Input 4 is an example of entailment because the second sentence (He was standing earlier) entails the first sentence (He didn't fall on his knees in the street). The first sentence implies that the man was not standing earlier since he was running and his knees were not on the ground. The second sentence explicitly states that he was standing earlier
    \vspace{0.5em}
    \noindent\makebox[\linewidth]{\color{blue}\dotfill}
    \par\vspace{0.1em}
    \centering
    Accuracy: 43.6\%
\end{promptbox}

\subsection{BigBench Instruction Induction (BBII) - Text Generation}
\label{appen:bbh-ii-gen_gen_prompts}
We randomly choose five tasks from the text generation tasks of BBII dataset.

\begin{promptbox}[purple]{gender inclusive sentences german}
    \small
    1. Add an 'innen' to every noun in the sentence that indicates people. Your friend followed the instruction and accurately implemented it in all the given examples, making sure all human nouns in the output now carry the suffix '-innen' indicating people. Instruction 2: Given a German text, make sure that all gendered nouns and pronouns in the output are inclusive. Your friend followed this instruction as well, by marking all the gendered nouns and pronouns with an asterisk (*) in the outputs to indicate inclusivity. The Instruction 3 that you mentioned is related to agreeing the maskulin singular pronouns and plural pronoun
    \vspace{0.5em}
    \noindent\makebox[\linewidth]{\color{blue}\dotfill}
    \par\vspace{0.1em}
    \centering
    Accuracy: 89.5\%
\end{promptbox}

\begin{promptbox}[purple]{tense}
    \small
    "Change the tense of each sentence to the one indicated. The given Inputs and their corresponding Outputs in the required tenses are: input: "Nikita has cooked dinner for us." (Change tense to FUTURE PERFECT) output: "Nikita will have cooked dinner for us." (Future Perfect) input: "The girl bought the telescope her teacher will have recommended to her." (Change tense to PAST PERFECT) output: "The girl bought the telescope her teacher had recommended to her." (Past Perfect) input: "The child had stopped crying." (Change tense to PRESENT PERFECT) output: "The child has stopped crying." (Present Perfect) input
    \vspace{0.5em}
    \noindent\makebox[\linewidth]{\color{blue}\dotfill}
    \par\vspace{0.1em}
    \centering
    Accuracy: 96.7\%
\end{promptbox}

\begin{promptbox}[purple]{word sorting}
    \small
	"Change the order of the first two words in each input." Based on the provided instruction, the friend has correctly followed the instruction for the given input-output pairs, except for the case where the instruction was misinterpreted as "Swap the position of the first and second words in the output" ( Input : duma cordage , Output: cordage duma ). For the correct application of the instruction, the output for the second input should have been "cordage duma" instead of "duma cordage" or "cordage duma" instead of "duma cordage" respectively, as in the reference instructions you have provided. Therefore, based on the consistency of your friend's application of the
    \vspace{0.5em}
    \noindent\makebox[\linewidth]{\color{blue}\dotfill}
    \par\vspace{0.1em}
    \centering
    Accuracy: 91.5\%
\end{promptbox}

\begin{promptbox}[purple]{object counting}
    \small
    "[Input] I have $<$something$>$, [Input] I have $<$different things$>$, [Output] = $<$sum of the number of items in the different things$>$." Based on the provided examples, the friend understood that the instruction is to find the sum of the number of objects in the different lists given the inputs, and the outputs are the respective sums. Therefore, their answers are correct, as the sum of number of items in the input with 'chair, fridge, lamp, table' is indeed 4. The same goes for the second and third input which evaluates to 18 and 11 (which should be 11 but I assume a typo in your final output
    \vspace{0.5em}
    \noindent\makebox[\linewidth]{\color{blue}\dotfill}
    \par\vspace{0.1em}
    \centering
    Accuracy: 37.5\%
\end{promptbox}

\begin{promptbox}[purple]{operators}
    \small
    "op i is the ith power of 2." For this instruction, the friend correctly computed the 5th power of 2 (2\^5), which is 16, for both inputs. The Instruction is: "op i is the absolute value of i." For this instruction, the friend correctly found the absolute value of -58, which is 58, for the input. The Instruction is: "op i is the number j such that the addition of i and j is 0." For this instruction, the friend correctly identified that for the input of 0, the number j to make the sum of i and j equal to zero	
    \vspace{0.5em}
    \noindent\makebox[\linewidth]{\color{blue}\dotfill}
    \par\vspace{0.1em}
    \centering
    Accuracy: 54.7\%
\end{promptbox}

\subsection{Instruction Induction (II)}
\label{appen:ii_gen_prompts}
We randomly choose five tasks from the II dataset.

\begin{promptbox}[black]{Num to Verbal}
    \small
    "Convert the given number to its written form, e.g. 123 as One Hundred and Twenty-Three." Your friend correctly followed this instruction for all the given inputs and outputs were written as "number, number hundred and number" form as per the instruction. Therefore, all the output pairs in your case are valid according to the instruction given.
    \vspace{0.5em}
    \noindent\makebox[\linewidth]{\color{blue}\dotfill}
    \par\vspace{0.1em}
    \centering
    Accuracy: 98.0\%
\end{promptbox}

\begin{promptbox}[black]{Cause and Effect}
    \small
	"For each given pair of sentences, determine which sentence is the cause of the situation described in the other sentence and write only that sentence as the output." Your friend has correctly identified and provided the sentence that represents the cause as output for each pair. It seems that your friend understood the instruction properly. Good job by your friend! Output: Input : Sentence 1: The man ate hot wings. Sentence 2: The man reached for a glass of water. Output : The man ate hot wings. Input : Sentence 1: The child dropped the toy. Sentence 2: The toy was broken. Output : The child dropped the toy.
    \vspace{0.5em}
    \noindent\makebox[\linewidth]{\color{blue}\dotfill}
    \par\vspace{0.1em}
    \centering
    Accuracy: 88.0\%
\end{promptbox}

\begin{promptbox}[black]{Taxonomy Animal}
    \small
	2. "Filter out the non-animal inputs from each list." Based on this instruction, your friend has produced correct outputs. The friend identified and removed non-animal words from each input list to produce output lists containing only animal words. This is in compliance with the instruction that requires the extraction of animal words only from the input strings and leaving out non-animal elements. Here's how this applies to some of the provided examples: Input: swan, plane, frog, pajamas, shark, butterfly Output: shark, butterfly, frog, swan Non-animal words - plane, pajamas - were removed, leaving behind shark, butter
    \vspace{0.5em}
    \noindent\makebox[\linewidth]{\color{blue}\dotfill}
    \par\vspace{0.1em}
    \centering
    Accuracy: 95.0\%
\end{promptbox}

\begin{promptbox}[black]{Larger Animal}
    \small
    "Choose the larger animal from the given pair and output only its name." Based on the provided information, the friend has followed the instruction correctly for each input-output pair. Therefore, the friend's outputs are valid according to the instruction. For instance, in the first pair, the instruction is to choose the larger animal between "bat" and "jaguar". The friend has correctly identified "jaguar" as the larger animal and outputted it as the result. Similarly, in the second pair, the friend has identified "puppy" as the larger animal and outputted it as the result. Therefore, the instruction and the friend's outputs are consistent, and the friend has
    \vspace{0.5em}
    \noindent\makebox[\linewidth]{\color{blue}\dotfill}
    \par\vspace{0.1em}
    \centering
    Accuracy: 93.0\%
\end{promptbox}

\begin{promptbox}[black]{Diff}
    \small
	-- Subtract the second number from the first number and find the difference's absolute value if the result is negative-- Take the output from the previous calculation and subtract 8 from it. Let's verify each of the input-output pairs based on this instruction: Input : 84 16 Output : 68 Calculation: $|$84-16$|$ = 68 and $|$68-8$|$ = 60, but we did not get this from input and the instruction didn't ask for that, so it seems to be a correct output (the first part of the instruction was satisfied: $|$84-16$|$ = 68 )
    \vspace{0.5em}
    \noindent\makebox[\linewidth]{\color{blue}\dotfill}
    \par\vspace{0.1em}
    \centering
    Accuracy: 100.0\%
\end{promptbox}

\subsection{Big-Bench Hard (BBH)}
\label{appen:bbh_gen_prompts}
We randomly choose five tasks from the BBH dataset.

\begin{promptbox}[black]{navigate}
    \small
    Always face forward unless stated otherwise. To determine if you return to the starting point, calculate the net displacement by summing steps along the x-axis (left/right) and y-axis (forward/backward) as vectors. Opposite directions on the same axis cancel each other. For the x-axis, left steps are negative and right steps are positive. For the y-axis, forward steps are positive and backward steps are negative. If the net displacement for both axes is zero, return Yes; otherwise, return No. Provide detailed step-by-step calculations, accounting for direction and magnitude of each step. Use this generalizable strategy for future problems.
    \vspace{0.5em}
    \noindent\makebox[\linewidth]{\color{blue}\dotfill}
    \par\vspace{0.1em}
    \centering
    Accuracy: 91.0\%
\end{promptbox}

\begin{promptbox}[black]{hyperbaton}
    \small
    The task is to determine the correct adjective order in a noun phrase by arranging adjectives from the most general to the most specific, following the sequence: opinion, size, age, shape, color, origin, material, purpose. The noun follows the purpose adjective. Given options, select the one that adheres to this order. The correct answer will have adjectives ordered as: opinion $>$ size $>$ age $>$ shape $>$ color $>$ origin $>$ material $>$ purpose, with the noun following the purpose adjective. Adjectives like `rubber' are considered material adjectives. The strategy involves analyzing the sequence of adjectives in noun phrases to ensure they follow the established order. The assistant should apply this generalizable strategy to identify the correct sentence structure from
    \vspace{0.5em}
    \noindent\makebox[\linewidth]{\color{blue}\dotfill}
    \par\vspace{0.1em}
    \centering
    Accuracy: 99.0\%
\end{promptbox}

\begin{promptbox}[black]{causal judgement}
    \small
    The task is to determine whether a specific event was a necessary and direct cause of an outcome, meaning the outcome would not have occurred without it. If the event is indispensable for the outcome, answer ``Yes''; otherwise, answer ``No''. Focus solely on the specified cause and consider that the outcome requires the simultaneous presence of multiple conditions. The cause is only valid if it is indispensable, and accidental or coincidental factors, such as errors or unrelated events, do not establish causation. For example, in a machine that short-circuits only when both wires touch the battery at the same time, the short circuit is caused by both wires touching, not by either wire alone. When evaluating causation, the outcome must depend on the presence
    \vspace{0.5em}
    \noindent\makebox[\linewidth]{\color{blue}\dotfill}
    \par\vspace{0.1em}
    \centering
    Accuracy: 61.0\%
\end{promptbox}

\begin{promptbox}[black]{geometric shapes}
    \small
    Analyze the given SVG path element and determine the geometric shape it represents by counting the number of line segments and vertices, considering that each `M' command starts a new path and each `L' command connects two points. The shape is formed by connecting the points in sequence, and the number of sides corresponds to the number of line segments, excluding overlapping or repeated segments. The options include circle, heptagon, hexagon, kite, line, octagon, pentagon, rectangle, sector, and triangle. Provide the correct letter corresponding to the shape.
    \vspace{0.5em}
    \noindent\makebox[\linewidth]{\color{blue}\dotfill}
    \par\vspace{0.1em}
    \centering
    Accuracy: 55.0\%
\end{promptbox}

\begin{promptbox}[black]{object counting}
    \small
    Count the total number of items by summing all the quantities mentioned for each type of object, ensuring
    that each item is counted according to its specified number. Convert numerical words like ‘three’ to their
    numeric form (e.g., ‘three’ becomes 3) before summing. Single instances of objects should be counted
    as 1. The task applies to any category of objects, such as fruits, musical instruments, or other items.
    Identify numerical quantities in the input, convert words to numbers when necessary, and sum all values
    to obtain the total count. The solution involves parsing the input text to extract quantities, handling both
    numeric and word-based representations, and aggregating them to provide the final total. Ensure that
    only items explicitly mentioned in the input are counted, and ignore any implicit or assumed quantities.
    \vspace{0.5em}
    \noindent\makebox[\linewidth]{\color{blue}\dotfill}
    \par\vspace{0.1em}
    \centering
    Accuracy: 100.0\%
\end{promptbox}

\subsection{AIME}
\label{appen:aime_gen_prompts}

\begin{promptbox}[black]{AIME}
    \small
When solving AIME problems, begin by carefully analyzing the given conditions and identifying key mathematical structures—such as symmetry, number theory properties, geometry configurations, or algebraic identities. Look for patterns or known formulas, and consider simplifying the problem through substitution, symmetry, or known theorems. Pay close attention to constraints like integer values, divisibility, or geometric relationships. For problems involving probability, use complementary counting or expected values; for number theory, factor numbers and use modular arithmetic. For geometry, look for right triangles, similar figures, or coordinate geometry applications. For algebraic expressions, simplify step-by-step, and watch for telescoping series or known identities. If the answer is a fraction, reduce it to lowest terms and compute the sum of numerator and denominator. Always verify that intermediate steps are consistent with the problem constraints, and check for integer solutions or specific conditions like coprimality or divisibility. Finally, double-check arithmetic and avoid careless errors in computation.
    \vspace{0.5em}
    \noindent\makebox[\linewidth]{\color{blue}\dotfill}
    \par\vspace{0.1em}
    \centering
    Accuracy: 56.0\%
\end{promptbox}

\section{Prompt Result Analysis}
\label{appen:prompt_result_analysis}

\Cref{tab:appen_prompt_comparison} presents the full prompts of Manual Prompt, GEPA, and \textsc{GFlowPO} for one II task and one BBII task from \Cref{sec:induction_qa}: Sentence Similarity and Hyperbaton. The Manual Prompts are concise fixed task descriptions, whereas GEPA and \textsc{GFlowPO} generate their prompts through optimization over task-specific training examples.

As shown in \Cref{tab:appen_prompt_comparison}, \textsc{GFlowPO} generates prompts that better capture the underlying task than either the Manual Prompt or GEPA. While the Manual Prompts provide only brief task descriptions, \textsc{GFlowPO} extracts more task-specific information from the training examples. For Sentence Similarity, \textsc{GFlowPO} correctly formulates the problem as graded similarity scoring and specifies the output scale, whereas GEPA drifts to an unrelated transformation objective. For Hyperbaton, \textsc{GFlowPO} frames the task as reordering words and provides a concrete worked example, whereas GEPA mainly adds a generic multiple-choice output template. These examples illustrate that \textsc{GFlowPO} discovers task-specific rules rather than merely rephrasing instructions or producing generic prompting templates, which is consistent with its performance improvements over both baselines.

\begin{table*}[!htb]
\centering
\footnotesize
\renewcommand{\arraystretch}{1.2}
\begin{tabular}{@{}l>{\raggedright\arraybackslash}p{0.41\textwidth}>{\raggedright\arraybackslash}p{0.41\textwidth}@{}}
\toprule
& \textbf{Sentence Similarity} & \textbf{Hyperbaton} \\
\midrule
\textbf{Manual}
& Rate from 0 (definitely not) to 5 (perfectly) the degree in which both sentences describe the same event. \newline (\textbf{Acc: 20.0})
& Which sentence has the correct adjective order? \newline (\textbf{Acc: 56.5}) \\
\midrule
\textbf{GEPA}
& You are an AI assistant that applies a transformation to given inputs. Your task is to identify and extract specific details from two sentences and then concatenate them into one sentence, maintaining the same tense and voice. The sentences will always be provided in a pair. The transformation should result in a clear, concise and grammatically correct answer that is logically equivalent to the original sentences \ldots \newline (\textbf{Acc: 18.8})
& The assistant is an AI designed to answer multiple-choice questions by selecting the single best answer. The assistant will be provided with an input and a list of choices, and is expected to output a JSON object with the following format: \texttt{\{"final\_answer": <the letter of the correct choice>, "solution\_pad": <your reasoning>\}}. Strictly follow the provided format for outputting your final answer and solution padding \ldots \newline (\textbf{Acc: 63.3}) \\
\midrule
\textbf{\textsc{GFlowPO}}
& I will provide you with two sentences. Based on the given sentences, determine the level of similarity between them. The output should be one of the following numbers: 0 - definitely not 1 - probably not 2 - possibly 3 - probably 4 - almost perfectly Based on the provided input-output pairs, it seems that your friend correctly identified the level of similarity between the two sentences for each pair. \newline (\textbf{Acc: 50.0})
& ``Rearrange the adjective and the noun in each input.'' based on this instruction, the friend's output for each input is correct. They have successfully rearranged the order of the adjective and the noun in each input to match the expected output. For example, with the first input: Input: a ``gold awful knife'' b ``awful gold knife'' Output: A ``awful gold knife'' Your friend has correctly rearranged the order of the adjective ``awful'' and the noun ``gold'' to match the expected output ``awful gold''. Similarly, for the other inputs, the friend has also rearranged the order of the adjective and the noun. \newline (\textbf{Acc: 73.0}) \\
\bottomrule
\end{tabular}
\caption{Prompts of Manual Prompt, GEPA, and \textsc{GFlowPO} on Sentence Similarity (II) and Hyperbaton (BBII). Test accuracy is reported in parentheses. Long GEPA prompts are truncated (\ldots).}
\label{tab:appen_prompt_comparison}
\end{table*}

\section{Meta prompt Example}
\label{appen:meta_prompt_design}
\begin{figure*}[htbp]
    \centering
    \includegraphics[width=0.9\textwidth]{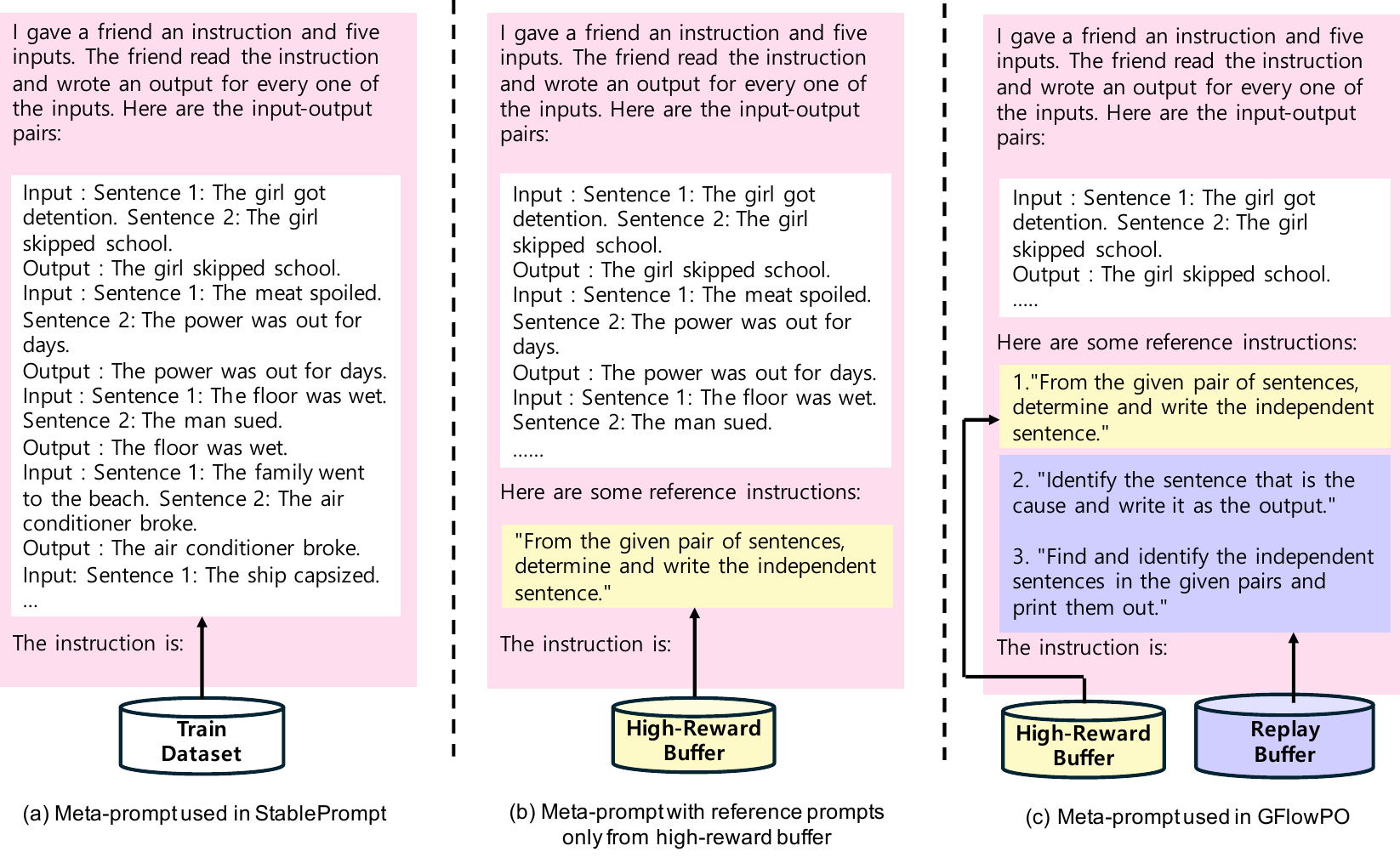}
    \caption{Meta prompt for prompt generation. \textbf{(a)} Meta-prompt used in StablePrompt, consisting of an instruction and few-shot examples. \textbf{(b)} Meta prompt with a reference prompt sampled only from high reward buffer $\mathcal{Q}$. \textbf{(c)} Meta prompt used in our proposed \textsc{GFlowPO}. Reference prompts are sampled from both high reward buffer $\mathcal{Q}$ and replay buffer $\mathcal{B}$, which balances exploitation and exploration during training.}
    \label{fig:appen_metaprompt}
\end{figure*}

We present the meta-prompt examples in \Cref{fig:appen_metaprompt}, where the template is used for experiment results in \Cref{sec:few_shot_tc} and \Cref{sec:induction_qa}. For reasoning-centric tasks in \Cref{sec:reasoning}, we use the modified meta-prompt from GEPA~\cite{agrawal2026gepareflectivepromptevolution}.

\section{Dataset Details and Verbalizer Settings on Few-Shot Text Classification Tasks}
\label{appen:few_shot_data_stat}
The dataset details and verbalizer settings on few-shot text classification tasks are presented in \Cref{tab:dataset_stats_verb}.

\begin{table*}[!t]
\centering
\begin{sc}
\begin{tabular}{@{}lccccc@{}}
\toprule
Dataset & Task type & \# classes & \# training & \# test & Verbalizer \\
\midrule
SST2 & Sentiment & 2 & 32 & 872   & [``yes'', ``no''] \\
MRPC & NLI       & 2 & 32 & 408   & [``yes'', ``no''] \\
RTE  & NLI       & 2 & 32 & 277   & [``yes'', ``no''] \\
QNLI & NLI       & 2 & 32 & 5,460 & [``yes'', ``no''] \\
MNLI & NLI       & 3 & 48 & 9,820 & [``yes'', ``maybe'', ``no''] \\
SNLI & NLI       & 3 & 48 & 9,842 & [``yes'', ``maybe'', ``no''] \\
\bottomrule
\end{tabular}
\end{sc}
\caption{Details of the dataset statistics and verbalizer on few-shot text classification tasks.}
\label{tab:dataset_stats_verb}
\end{table*}
\section{Dataset Details on BigBench-Hard Instruction Induction~(BBII) and Instruction Induction~(II)}
\label{appen:dataset_details_bbii_ii}
The dataset details on BigBench-Hard Instruction Induction~(BBII) and Instruction Induction~(II) are presented in \Cref{tab:dataset_details_bigbench} and \Cref{tab:dataset_details_ii}, respectively.

\begin{table*}[!htb]
\vspace{\baselineskip}

\vspace{\baselineskip}
\centering
\setlength{\tabcolsep}{1.5pt} 
\begin{sc}
\resizebox{0.75\textwidth}{!}{
\begin{tabular}{@{}lcccc@{}}
\toprule
Task name & Task type & Metric & \# training & \# test \\
\midrule
Causal judgment            & Multiple choice & Accuracy    & 30 & 160    \\
Disambiguation qa          & Multiple choice & Accuracy    & 30 & 228    \\
Epistemic reasoning        & Multiple choice & Accuracy    & 30 & 1,970  \\
Hyperbaton                 & Multiple choice & Accuracy    & 30 & 49,970 \\
Implicatures               & Multiple choice & Accuracy    & 30 & 462    \\
Logical fallacy detection  & Multiple choice & Accuracy    & 30 & 2,770  \\
Movie recommendation       & Multiple choice & Accuracy    & 30 & 470    \\
Navigate                   & Multiple choice & Accuracy    & 30 & 970    \\
Presuppositions as nli     & Multiple choice & Accuracy    & 30 & 705    \\
Ruin names                 & Multiple choice & Accuracy    & 30 & 418    \\
Snarks                     & Multiple choice & Accuracy    & 30 & 151    \\
Sports understanding        & Multiple choice & Accuracy    & 30 & 970    \\
Dyck languages             & Generation      & Exact match & 30 & 970    \\
Gender inclusive sentences & Generation      & Exact match & 30 & 170    \\
Object counting            & Generation      & Exact match & 30 & 970    \\
Operators                  & Generation      & Exact match & 30 & 181    \\
Tense                      & Generation      & Exact match & 30 & 256    \\
Word sorting               & Generation      & Exact match & 30 & 1,870  \\
\bottomrule
\end{tabular}
}
\end{sc}
\caption{Details of BigBench-Hard instruction induction datasets.}
\label{tab:dataset_details_bigbench}
\vspace{\baselineskip}
\end{table*}

\begin{table*}[!htb]
\vspace{\baselineskip}

\vspace{\baselineskip}
\centering
\setlength{\tabcolsep}{1.5pt} 
\begin{sc}
\resizebox{0.6\textwidth}{!}{
\begin{tabular}{@{}lccc@{}}
\toprule
Task name & Metric & \# training & \# test \\
\midrule
Antonyms                & Exact match & 32 & 100 \\
Word in context         & Exact match & 32 & 100 \\
Rhymes                  & Exact match & 32 & 100 \\
Num to verbal           & Exact match & 32 & 100 \\
Cause and effect        & Exact match & 26 & 25  \\
Larger animal           & Exact match & 32 & 100 \\
Second word letter      & Exact match & 32 & 100 \\
Taxonomy animal         & Exact set   & 32 & 100 \\
Negation                & Exact match & 32 & 100 \\
Common concept          & F1 score    & 17 & 16  \\
Diff                    & Exact match & 32 & 100 \\
Translation en-es       & Exact match & 32 & 100 \\
Orthography starts with & Exact set   & 32 & 100 \\
Sentiment               & Exact match & 32 & 100 \\
Informal to formal      & F1 score    & 15 & 15  \\
Sum                     & Exact match & 32 & 100 \\
Singular to plural      & Exact match & 32 & 100 \\
Active to passive       & Exact match & 32 & 100 \\
Translation en-de       & Exact match & 32 & 100 \\
Sentence similarity     & Exact match & 32 & 100 \\
Translation en-fr       & Exact match & 32 & 100 \\
Letters list            & Exact match & 32 & 100 \\
First word letter       & Exact match & 32 & 100 \\
Synonyms                & Contains    & 32 & 100 \\
\bottomrule
\end{tabular}
}
\end{sc}
\caption{Details of instruction induction datasets.}
\label{tab:dataset_details_ii}
\end{table*}

\section{Dataset Details on MMLU}
\label{appen:dataset_details_mmlu}

The dataset details on MMLU are presented in \Cref{tab:mmlu_stats_summary}.
\begin{table*}[!htb]
\vspace{\baselineskip}

\vspace{\baselineskip}
\centering
\setlength{\tabcolsep}{1.5pt} 
\begin{sc}
\renewcommand{\arraystretch}{0.8}
\begin{tabular}{@{}clcc@{}}
\toprule
Type & Subject & \# training & \# test \\
\midrule
STEM
 & Abstract algebra             & 11  & 100  \\
 & Anatomy                      & 14  & 135  \\
 & Astronomy                    & 16  & 152  \\
 & College biology              & 16  & 144  \\
 & College chemistry            & 8   & 100  \\
 & College computer science     & 11  & 100  \\
 & College mathematics          & 11  & 100  \\
 & College physics              & 11  & 102  \\
 & Computer security            & 11  & 100  \\
 & Conceptual physics           & 26  & 235  \\
 & Electrical engineering       & 16  & 145  \\
 & Elementary mathematics       & 41  & 378  \\
 & High school biology          & 32  & 310  \\
 & High school chemistry        & 22  & 203  \\
 & High school computer science & 9   & 100  \\
 & High school mathematics      & 29  & 270  \\
 & High school physics          & 17  & 151  \\
 & High school statistics       & 23  & 216  \\
 & Machine learning             & 11  & 112  \\
\midrule
Social science
 & Econometrics                        & 12  & 114  \\
 & High school geography               & 22  & 198  \\
 & High school government and politics & 21  & 193  \\
 & High school macroeconomics          & 43  & 390  \\
 & High school microeconomics          & 26  & 238  \\
 & High school psychology              & 60  & 545  \\
 & Human sexuality                     & 12  & 131  \\
 & Professional psychology             & 69  & 612  \\
 & Public relations                    & 12  & 110  \\
 & Security studies                    & 27  & 245  \\
 & Sociology                           & 22  & 201  \\
 & US foreign policy                   & 11  & 100  \\
\midrule
Humanities
 & Formal logic                  & 14  & 126  \\
 & High school European history  & 18  & 165  \\
 & High school US history        & 22  & 204  \\
 & High school world history     & 26  & 237  \\
 & International law             & 13  & 121  \\
 & Jurisprudence                 & 11  & 108  \\
 & Logical fallacies             & 18  & 163  \\
 & Moral disputes                & 38  & 346  \\
 & Moral scenarios               & 100 & 895  \\
 & Philosophy                    & 34  & 311  \\
 & Prehistory                    & 35  & 324  \\
 & Professional law              & 170 & 1534 \\
 & World religions               & 19  & 171  \\
\midrule
Others
 & Business ethics         & 11 & 100 \\
 & Clinical knowledge      & 29 & 265 \\
 & College medicine        & 22 & 173 \\
 & Global facts            & 10 & 100 \\
 & Human aging             & 23 & 223 \\
 & Management              & 11 & 103 \\
 & Marketing               & 25 & 234 \\
 & Medical genetics        & 11 & 100 \\
 & Miscellaneous           & 86 & 783 \\
 & Nutrition               & 33 & 306 \\
 & Professional accounting & 31 & 282 \\
 & Professional medicine   & 31 & 272 \\
 & Virology                & 18 & 166 \\
\bottomrule
\end{tabular}
\end{sc}
\caption{MMLU dataset statistics: training and test sizes by category.}
\label{tab:mmlu_stats_summary}
\vspace{\baselineskip}
\end{table*}

\section{Full Experimental results}
\label{appen:full_results}

In this section, we provide full experimental results in
\Crefrange{tab:appen_results_bbii}{tab:appen_bbh_performance}.

\begin{table*}[!htb]
\vspace{\baselineskip}

\vspace{\baselineskip}
\centering
\begin{sc}
\setlength{\tabcolsep}{1.5pt} 
\resizebox{\textwidth}{!}{
\begin{tabular}{@{}lcccccccccccccc@{}}
\toprule
Task name & Type
& Fewshot & Manual & APE & ProTeGi
& EvoPrompt & PromptWizard & GRACE & PRL & GEPA
& StablePrompt & \textsc{GFlowPO} \\
\midrule
Causal judgment            & Multiple choice
& \underline{58.8} & 52.5 & 58.1 & 56.7
& $50.6{\scriptstyle\pm0.0}$ & $47.5{\scriptstyle\pm0.6}$ & $55.4{\scriptstyle\pm1.4}$ & $\textbf{59.8}{\scriptstyle\pm1.0}$ & $58.1{\scriptstyle\pm1.7}$
& $57.1{\scriptstyle\pm1.8}$ & $\underline{58.8}{\scriptstyle\pm1.1}$ \\
Disambiguation qa          & Multiple choice
& 64.3 & 52.2 & 64.0 & 61.4
& $61.0{\scriptstyle\pm5.3}$ & $52.9{\scriptstyle\pm5.8}$ & $51.3{\scriptstyle\pm5.8}$ & $57.3{\scriptstyle\pm6.4}$ & $\textbf{66.2}{\scriptstyle\pm5.8}$
& $64.6{\scriptstyle\pm2.3}$ & $\underline{64.8}{\scriptstyle\pm1.1}$ \\
Epistemic reasoning        & Multiple choice
& 43.7 & 57.2 & 58.4 & \textbf{63.8}
& $54.0{\scriptstyle\pm15.0}$ & $50.3{\scriptstyle\pm7.5}$ & $50.9{\scriptstyle\pm14.1}$ & $50.9{\scriptstyle\pm14.5}$ & $44.0{\scriptstyle\pm1.3}$
& $\underline{61.0}{\scriptstyle\pm1.4}$ & $59.0{\scriptstyle\pm5.2}$ \\
Hyperbaton                 & Multiple choice
& 47.9 & 56.5 & 75.6 & \underline{76.1}
& $\textbf{79.2}{\scriptstyle\pm1.4}$ & $75.7{\scriptstyle\pm3.5}$ & $65.8{\scriptstyle\pm6.6}$ & $70.2{\scriptstyle\pm7.2}$ & $64.5{\scriptstyle\pm7.2}$
& $67.6{\scriptstyle\pm7.1}$ & $74.8{\scriptstyle\pm3.7}$ \\
Implicatures               & Multiple choice
& \underline{83.3} & 83.1 & 81.0 & 73.6
& $82.3{\scriptstyle\pm1.0}$ & $74.9{\scriptstyle\pm8.4}$ & $79.9{\scriptstyle\pm0.0}$ & $79.0{\scriptstyle\pm3.6}$ & $82.8{\scriptstyle\pm0.3}$
& $82.3{\scriptstyle\pm0.6}$ & $\textbf{84.5}{\scriptstyle\pm1.0}$ \\
Logical fallacy detection  & Multiple choice
& 58.2 & \textbf{63.5} & 56.5 & 58.2
& $\textbf{63.6}{\scriptstyle\pm6.3}$ & $53.2{\scriptstyle\pm0.3}$ & $56.8{\scriptstyle\pm0.5}$ & $60.3{\scriptstyle\pm4.5}$ & $\underline{60.9}{\scriptstyle\pm0.1}$
& $52.9{\scriptstyle\pm0.1}$ & $53.2{\scriptstyle\pm0.2}$ \\
Movie recommendation       & Multiple choice
& 49.4 & 37.7 & 55.3 & \underline{67.2}
& $62.6{\scriptstyle\pm4.6}$ & $58.9{\scriptstyle\pm7.5}$ & $58.4{\scriptstyle\pm9.2}$ & $61.1{\scriptstyle\pm2.0}$ & $\underline{66.1}{\scriptstyle\pm7.5}$
& $63.7{\scriptstyle\pm1.6}$ & $\textbf{73.6}{\scriptstyle\pm3.8}$ \\
Navigate                   & Multiple choice
& \textbf{69.2} & 49.8 & 52.3 & \underline{54.0}
& $\underline{52.6}{\scriptstyle\pm2.5}$ & $50.6{\scriptstyle\pm0.5}$ & $49.3{\scriptstyle\pm0.6}$ & $50.2{\scriptstyle\pm0.4}$ & $51.7{\scriptstyle\pm2.0}$
& $\textbf{51.6}{\scriptstyle\pm1.1}$ & $52.5{\scriptstyle\pm0.3}$ \\
Presuppositions as nli     & Multiple choice
& 42.6 & 40.8 & 41.6 & 41.4
& $40.9{\scriptstyle\pm0.1}$ & $41.0{\scriptstyle\pm0.9}$ & $41.6{\scriptstyle\pm0.6}$ & $43.4{\scriptstyle\pm1.7}$ & $43.2{\scriptstyle\pm0.8}$
& $\underline{43.6}{\scriptstyle\pm3.0}$ & $\textbf{46.0}{\scriptstyle\pm2.3}$ \\
Ruin names                 & Multiple choice
& 12.4 & 30.1 & \underline{32.5} & 28.0
& $\underline{36.4}{\scriptstyle\pm7.6}$ & $26.5{\scriptstyle\pm3.2}$ & $25.9{\scriptstyle\pm1.8}$ & $34.7{\scriptstyle\pm6.6}$ & $24.9{\scriptstyle\pm0.4}$
& $\textbf{27.7}{\scriptstyle\pm0.1}$ & $27.8{\scriptstyle\pm1.8}$ \\
Snarks                     & Multiple choice
& 35.8 & 42.4 & 51.0 & 51.0
& $48.6{\scriptstyle\pm1.4}$ & $49.0{\scriptstyle\pm1.8}$ & $46.1{\scriptstyle\pm2.0}$ & $48.1{\scriptstyle\pm2.7}$ & $50.3{\scriptstyle\pm1.1}$
& $\underline{52.8}{\scriptstyle\pm2.0}$ & $\textbf{60.0}{\scriptstyle\pm5.3}$ \\
Sportsunderstanding        & Multiple choice
& 52.4 & 59.4 & 56.5 & 56.0
& $\underline{62.2}{\scriptstyle\pm3.4}$ & $60.0{\scriptstyle\pm5.5}$ & $54.2{\scriptstyle\pm3.8}$ & $59.2{\scriptstyle\pm2.6}$ & $60.1{\scriptstyle\pm2.7}$
& $60.7{\scriptstyle\pm2.1}$ & $\textbf{65.9}{\scriptstyle\pm2.0}$ \\
Dyck languages             & Generation
& 0.0 & 0.0 & 0.0 & 0.0
& $0.0{\scriptstyle\pm0.0}$ & $0.0{\scriptstyle\pm0.0}$ & $0.0{\scriptstyle\pm0.0}$ & $0.0{\scriptstyle\pm0.0}$ & $0.0{\scriptstyle\pm0.0}$
& $0.0{\scriptstyle\pm0.0}$ & $0.0{\scriptstyle\pm0.0}$ \\
Gender inclusive sentences & Generation
& 9.3 & 86.0 & 67.1 & \textbf{93.8}
& $86.0{\scriptstyle\pm1.4}$ & $86.1{\scriptstyle\pm0.9}$ & $85.7{\scriptstyle\pm6.3}$ & $\underline{90.3}{\scriptstyle\pm0.3}$ & $83.4{\scriptstyle\pm5.8}$
& $86.4{\scriptstyle\pm3.2}$ & $\textbf{90.4}{\scriptstyle\pm0.7}$ \\
Object counting            & Generation
& 7.1 & 0.0 & 14.3 & \underline{33.3}
& $37.3{\scriptstyle\pm1.9}$ & $\underline{37.7}{\scriptstyle\pm2.1}$ & $31.4{\scriptstyle\pm3.9}$ & $37.5{\scriptstyle\pm1.2}$ & $33.2{\scriptstyle\pm0.6}$
& $30.9{\scriptstyle\pm1.8}$ & $\textbf{38.2}{\scriptstyle\pm2.8}$ \\
Operators                  & Generation
& 5.5 & 49.5 & \textbf{57.1} & 50.0
& $55.0{\scriptstyle\pm0.0}$ & $55.0{\scriptstyle\pm0.7}$ & $50.3{\scriptstyle\pm3.2}$ & $\textbf{56.1}{\scriptstyle\pm0.1}$ & $\underline{55.9}{\scriptstyle\pm1.2}$
& $49.5{\scriptstyle\pm2.6}$ & $53.5{\scriptstyle\pm4.0}$ \\
Tense                      & Generation
& 15.3 & 93.9 & \underline{96.8} & \textbf{100.0}
& $95.6{\scriptstyle\pm0.1}$ & $95.4{\scriptstyle\pm0.5}$ & $93.2{\scriptstyle\pm5.4}$ & $\textbf{96.7}{\scriptstyle\pm0.9}$ & $95.8{\scriptstyle\pm1.3}$
& $95.1{\scriptstyle\pm0.4}$ & $\underline{96.3}{\scriptstyle\pm0.7}$ \\
Word sorting               & Generation
& 0.0 & 20.1 & \textbf{96.4} & 75.0
& $46.3{\scriptstyle\pm34.9}$ & $64.9{\scriptstyle\pm19.8}$ & $66.0{\scriptstyle\pm35.5}$ & $68.3{\scriptstyle\pm21.1}$ & $\underline{92.7}{\scriptstyle\pm2.1}$
& $85.0{\scriptstyle\pm2.7}$ & $\textbf{95.6}{\scriptstyle\pm3.6}$ \\
\bottomrule
\end{tabular}
}
\end{sc}
\caption{Full experimental results on the Big-Bench Instruction Induction (BBII) datasets.}
\label{tab:appen_results_bbii}
\vspace{\baselineskip}
\end{table*}

\begin{table*}[!htb]
\vspace{\baselineskip}

\vspace{\baselineskip}
\centering
\begin{sc}
\setlength{\tabcolsep}{1.5pt}
\resizebox{\textwidth}{!}{
\begin{tabular}{@{}lcccccccccccc@{}}
\toprule
Task name & Metric
& Fewshot & Manual & APE & ProTeGi
& EvoPrompt & PromptWizard & GRACE & PRL & GEPA
& StablePrompt & \textsc{GFlowPO} \\
\midrule
Antonyms
& Exact match & 0.0 & 43.0 & 62.5 & 25.0
& $47.3{\scriptstyle\pm15.0}$ & $56.0{\scriptstyle\pm19.0}$ & $64.3{\scriptstyle\pm4.9}$ & $37.7{\scriptstyle\pm32.9}$ & $45.7{\scriptstyle\pm18.2}$
& $\underline{67.7}{\scriptstyle\pm3.2}$ & $\textbf{74.3}{\scriptstyle\pm2.3}$ \\
Word in context
& Exact match & 55.0 & 46.0 & 37.5 & 50.0
& $56.3{\scriptstyle\pm3.5}$ & $60.0{\scriptstyle\pm3.6}$ & $44.3{\scriptstyle\pm11.7}$ & $\textbf{61.0}{\scriptstyle\pm4.6}$ & $58.3{\scriptstyle\pm1.5}$
& $\underline{60.3}{\scriptstyle\pm5.0}$ & $\textbf{61.0}{\scriptstyle\pm4.4}$ \\
Rhymes
& Exact match & 0.0 & 3.0 & 6.3 & \textbf{25.0}
& $6.7{\scriptstyle\pm3.2}$ & $6.7{\scriptstyle\pm6.0}$ & $9.3{\scriptstyle\pm3.1}$ & $\textbf{12.0}{\scriptstyle\pm3.5}$ & $4.7{\scriptstyle\pm1.5}$
& $3.7{\scriptstyle\pm3.8}$ & $\underline{8.7}{\scriptstyle\pm2.1}$ \\
Num to verbal
& Exact match & 0.0 & 61.0 & 93.8 & \textbf{100.0}
& $73.7{\scriptstyle\pm6.4}$ & $83.7{\scriptstyle\pm4.0}$ & $82.7{\scriptstyle\pm18.6}$ & $81.3{\scriptstyle\pm12.7}$ & $82.3{\scriptstyle\pm14.2}$
& $\underline{89.0}{\scriptstyle\pm4.4}$ & $\textbf{97.7}{\scriptstyle\pm1.5}$ \\
Cause and effect
& Exact match & 0.0 & 24.0 & \underline{60.0} & 0.0
& $24.0{\scriptstyle\pm0.0}$ & $48.0{\scriptstyle\pm10.6}$ & $25.3{\scriptstyle\pm22.0}$ & $49.3{\scriptstyle\pm30.0}$ & $52.0{\scriptstyle\pm17.4}$
& $54.7{\scriptstyle\pm10.1}$ & $\textbf{81.3}{\scriptstyle\pm11.5}$ \\
Larger animal
& Exact match & 0.0 & 3.0 & 56.3 & 25.0
& $6.0{\scriptstyle\pm0.0}$ & $47.3{\scriptstyle\pm29.0}$ & $51.7{\scriptstyle\pm5.8}$ & $72.7{\scriptstyle\pm25.8}$ & $48.0{\scriptstyle\pm10.2}$
& $\underline{86.7}{\scriptstyle\pm8.5}$ & $\textbf{92.7}{\scriptstyle\pm0.6}$ \\
Second word letter
& Exact match & 12.0 & 8.0 & 6.3 & \underline{25.0}
& $17.7{\scriptstyle\pm4.0}$ & $18.7{\scriptstyle\pm10.1}$ & $\underline{21.0}{\scriptstyle\pm15.1}$ & $8.0{\scriptstyle\pm0.0}$ & $7.0{\scriptstyle\pm5.6}$
& $9.7{\scriptstyle\pm4.2}$ & $\textbf{27.3}{\scriptstyle\pm3.5}$ \\
Taxonomy animal
& Exact set & 0.0 & 0.0 & 37.5 & 37.5
& $25.3{\scriptstyle\pm25.6}$ & $4.0{\scriptstyle\pm4.6}$ & $26.7{\scriptstyle\pm44.5}$ & $\underline{63.7}{\scriptstyle\pm38.9}$ & $22.7{\scriptstyle\pm3.5}$
& $53.7{\scriptstyle\pm19.6}$ & $\textbf{93.7}{\scriptstyle\pm8.4}$ \\
Negation
& Exact match & 0.0 & 16.0 & 68.8 & 50.0
& $6.7{\scriptstyle\pm5.0}$ & $35.3{\scriptstyle\pm8.7}$ & $59.7{\scriptstyle\pm7.1}$ & $70.3{\scriptstyle\pm6.7}$ & $74.7{\scriptstyle\pm3.2}$
& $\underline{83.7}{\scriptstyle\pm2.3}$ & $\textbf{87.0}{\scriptstyle\pm0.0}$ \\
Common concept
& F1 & 3.0 & 4.0 & \textbf{50.0} & \textbf{50.0}
& $\underline{12.8}{\scriptstyle\pm0.0}$ & $5.0{\scriptstyle\pm2.1}$ & $2.6{\scriptstyle\pm3.2}$ & $6.7{\scriptstyle\pm0.5}$ & $12.1{\scriptstyle\pm3.9}$
& $3.3{\scriptstyle\pm1.2}$ & $\textbf{13.7}{\scriptstyle\pm3.2}$ \\
Diff
& Exact match & 2.0 & \underline{99.0} & \textbf{100.0} & \textbf{100.0}
& $\textbf{100.0}{\scriptstyle\pm0.0}$ & $\underline{97.3}{\scriptstyle\pm1.5}$ & $\textbf{100.0}{\scriptstyle\pm0.0}$ & $\textbf{100.0}{\scriptstyle\pm0.0}$ & $82.0{\scriptstyle\pm23.1}$
& $\textbf{100.0}{\scriptstyle\pm0.0}$ & $\textbf{100.0}{\scriptstyle\pm0.0}$ \\
Translation en-es
& Exact match & 0.0 & 15.0 & 25.0 & 25.0
& $33.0{\scriptstyle\pm4.6}$ & $27.3{\scriptstyle\pm15.3}$ & $36.3{\scriptstyle\pm14.2}$ & $35.3{\scriptstyle\pm1.2}$ & $40.0{\scriptstyle\pm4.6}$
& $\textbf{43.0}{\scriptstyle\pm0.0}$ & $\underline{40.3}{\scriptstyle\pm3.8}$ \\
Orthography starts with
& Exact set & 0.0 & \textbf{37.5} & 12.5 & 0.0
& $\textbf{60.7}{\scriptstyle\pm2.5}$ & $\underline{58.3}{\scriptstyle\pm4.0}$ & $41.3{\scriptstyle\pm13.9}$ & $35.3{\scriptstyle\pm1.2}$ & $8.0{\scriptstyle\pm1.0}$
& $13.3{\scriptstyle\pm3.2}$ & $15.0{\scriptstyle\pm1.0}$ \\
Sentiment
& Exact match & 50.0 & 83.0 & 68.8 & \textbf{100.0}
& $84.7{\scriptstyle\pm1.2}$ & $84.7{\scriptstyle\pm1.2}$ & $70.3{\scriptstyle\pm6.1}$ & $85.0{\scriptstyle\pm1.0}$ & $84.3{\scriptstyle\pm2.3}$
& $\underline{86.0}{\scriptstyle\pm3.0}$ & $\textbf{91.7}{\scriptstyle\pm2.1}$ \\
Informal to formal
& F1 & 0.0 & 27.4 & 42.5 & 24.2
& $24.8{\scriptstyle\pm9.1}$ & $28.3{\scriptstyle\pm3.9}$ & $18.6{\scriptstyle\pm2.9}$ & $34.8{\scriptstyle\pm3.4}$ & $30.0{\scriptstyle\pm9.7}$
& $\underline{43.5}{\scriptstyle\pm4.0}$ & $\textbf{47.3}{\scriptstyle\pm2.1}$ \\
Sum
& Exact match & 0.0 & \underline{99.0} & \textbf{100.0} & \textbf{100.0}
& $\textbf{100.0}{\scriptstyle\pm0.0}$ & $\underline{99.0}{\scriptstyle\pm1.0}$ & $\textbf{100.0}{\scriptstyle\pm0.0}$ & $\textbf{100.0}{\scriptstyle\pm0.0}$ & $\textbf{100.0}{\scriptstyle\pm0.0}$
& $\textbf{100.0}{\scriptstyle\pm0.0}$ & $\textbf{100.0}{\scriptstyle\pm0.0}$ \\
Singular to plural
& Exact match & 0.0 & 75.0 & 93.8 & \textbf{100.0}
& $93.0{\scriptstyle\pm0.0}$ & $95.0{\scriptstyle\pm0.0}$ & $95.0{\scriptstyle\pm3.5}$ & $79.3{\scriptstyle\pm27.2}$ & $\textbf{99.3}{\scriptstyle\pm1.2}$
& $96.0{\scriptstyle\pm1.0}$ & $\underline{98.3}{\scriptstyle\pm1.5}$ \\
Active to passive
& Exact match & 0.0 & 53.0 & \textbf{100.0} & \textbf{100.0}
& $89.3{\scriptstyle\pm9.5}$ & $95.3{\scriptstyle\pm2.5}$ & $63.3{\scriptstyle\pm45.9}$ & $98.3{\scriptstyle\pm2.1}$ & $98.3{\scriptstyle\pm2.1}$
& $\textbf{100.0}{\scriptstyle\pm0.0}$ & $\underline{99.0}{\scriptstyle\pm0.0}$ \\
Translation en-de
& Exact match & 0.0 & 10.0 & 18.8 & \textbf{50.0}
& $23.7{\scriptstyle\pm3.5}$ & $15.0{\scriptstyle\pm7.0}$ & $17.7{\scriptstyle\pm2.5}$ & $25.0{\scriptstyle\pm2.0}$ & $\underline{30.3}{\scriptstyle\pm3.5}$
& $29.7{\scriptstyle\pm5.0}$ & $\textbf{32.0}{\scriptstyle\pm2.7}$ \\
Sentence similarity
& Exact match & 0.0 & 20.0 & 31.5 & 25.0
& $0.0{\scriptstyle\pm0.0}$ & $0.0{\scriptstyle\pm0.0}$ & $0.0{\scriptstyle\pm0.0}$ & $0.0{\scriptstyle\pm0.0}$ & $17.7{\scriptstyle\pm14.5}$
& $\underline{32.7}{\scriptstyle\pm4.2}$ & $\textbf{38.7}{\scriptstyle\pm2.5}$ \\
Translation en-fr
& Exact match & 0.0 & 7.0 & 6.0 & \textbf{50.0}
& $22.3{\scriptstyle\pm4.2}$ & $11.3{\scriptstyle\pm5.8}$ & $21.3{\scriptstyle\pm6.5}$ & $23.7{\scriptstyle\pm4.2}$ & $\underline{35.0}{\scriptstyle\pm6.2}$
& $26.3{\scriptstyle\pm5.0}$ & $\textbf{40.3}{\scriptstyle\pm1.5}$ \\
Letters list
& Exact match & 0.0 & 0.0 & 68.8 & 50.0
& $1.7{\scriptstyle\pm1.5}$ & $12.7{\scriptstyle\pm21.9}$ & $49.3{\scriptstyle\pm47.2}$ & $0.0{\scriptstyle\pm0.0}$ & $77.0{\scriptstyle\pm17.8}$
& $\underline{82.3}{\scriptstyle\pm10.7}$ & $\textbf{95.3}{\scriptstyle\pm2.1}$ \\
First word letter
& Exact match & 3.0 & 73.0 & 87.8 & \textbf{100.0}
& $92.7{\scriptstyle\pm6.4}$ & $96.7{\scriptstyle\pm2.9}$ & $97.0{\scriptstyle\pm2.0}$ & $\underline{98.0}{\scriptstyle\pm0.0}$ & $80.7{\scriptstyle\pm15.0}$
& $97.0{\scriptstyle\pm1.0}$ & $\textbf{99.0}{\scriptstyle\pm1.0}$ \\
Synonyms
& Contains & 0.0 & 2.0 & 12.5 & \textbf{25.0}
& $11.0{\scriptstyle\pm4.6}$ & $12.7{\scriptstyle\pm3.1}$ & $18.3{\scriptstyle\pm3.1}$ & $22.7{\scriptstyle\pm3.5}$ & $5.7{\scriptstyle\pm5.0}$
& $\underline{23.0}{\scriptstyle\pm10.5}$ & $\textbf{24.7}{\scriptstyle\pm5.1}$ \\
\midrule
Average accuracy
& & 5.2 & 33.7 & 51.9 & 51.6
& $42.2{\scriptstyle\pm0.7}$ & $45.8{\scriptstyle\pm2.0}$ & $46.5{\scriptstyle\pm4.4}$ & $53.2{\scriptstyle\pm2.4}$ & $49.8{\scriptstyle\pm2.0}$
& $\underline{57.7}{\scriptstyle\pm0.4}$ & $\textbf{65.0}{\scriptstyle\pm0.5}$ \\
\bottomrule
\end{tabular}
}
\end{sc}
\caption{Full experimental results on the Instruction Induction (II).}
\label{tab:results_instruction_induction}
\vspace{\baselineskip}
\end{table*}

\begin{table*}[!htb]
\vspace{\baselineskip}

\vspace{\baselineskip}
\centering
\begin{sc}
\setlength{\tabcolsep}{1.5pt} 
\resizebox{\textwidth}{!}{
\begin{tabular}{@{}clcccccccccccc@{}}
\toprule
Task group & Task name & Metric & Fewshot & Manual & APE & ProTeGi & \shortstack{Evo\\Prompt} & \shortstack{Prompt\\Wizard} & GRACE & PRL & GEPA & \shortstack{Stable\\Prompt} & GFlowPO \\
\midrule
\multirow{5}{*}{Spelling}
  & First word letter      & Exact match & 3.0  & 73.0 & 87.8          & \textbf{100.0} & $92.7{\scriptstyle\pm6.4}$  & $96.7{\scriptstyle\pm2.9}$  & $97.0{\scriptstyle\pm2.0}$  & $\underline{98.0}{\scriptstyle\pm0.0}$  & $80.7{\scriptstyle\pm15.0}$ & 97.0          & \underline{99.0} \\
  & Second word letter     & Exact match & 12.0 & 8.0  & 6.3           & \underline{25.0} & $17.7{\scriptstyle\pm4.0}$ & $18.7{\scriptstyle\pm10.1}$ & $\underline{21.0}{\scriptstyle\pm15.1}$ & $8.0{\scriptstyle\pm0.0}$  & $7.0{\scriptstyle\pm5.6}$   & 9.7           & \textbf{27.3} \\
  & Letters list           & Exact match & 0.0  & 0.0  & 68.8          & 50.0           & $1.7{\scriptstyle\pm1.5}$   & $12.7{\scriptstyle\pm21.9}$ & $49.3{\scriptstyle\pm47.2}$ & $0.0{\scriptstyle\pm0.0}$  & $77.0{\scriptstyle\pm17.8}$ & \underline{82.3} & \textbf{95.3} \\
  & Orthography starts with & Exact set  & 0.0  & \textbf{37.5} & 12.5 & 0.0   & $\textbf{60.7}{\scriptstyle\pm2.5}$ & $\underline{58.3}{\scriptstyle\pm4.0}$ & $41.3{\scriptstyle\pm13.9}$ & $35.3{\scriptstyle\pm1.2}$ & $8.0{\scriptstyle\pm1.0}$ & 13.3 & 15.0 \\
\cmidrule(lr){2-14}
  & Average                &             & 3.8  & 29.6 & 43.8          & 43.8           & 42.2      & 46.6      & 52.2      & 35.3     & 43.2      & \underline{50.6} & \textbf{59.2} \\
\midrule
\multirow{5}{*}{Syntax}
  & Singular to plural     & Exact match & 0.0  & 75.0 & 93.8          & \textbf{100.0} & $93.0{\scriptstyle\pm0.0}$  & $95.0{\scriptstyle\pm0.0}$  & $95.0{\scriptstyle\pm3.5}$  & $79.3{\scriptstyle\pm27.2}$ & $\textbf{99.3}{\scriptstyle\pm1.2}$ & 96.0 & \underline{98.3} \\
  & Active to passive      & Exact match & 0.0  & 53.0 & \textbf{100.0} & \textbf{100.0} & $89.3{\scriptstyle\pm9.5}$ & $95.3{\scriptstyle\pm2.5}$  & $63.3{\scriptstyle\pm45.9}$ & $98.3{\scriptstyle\pm2.1}$ & $98.3{\scriptstyle\pm2.1}$  & \textbf{100.0} & \underline{99.0} \\
  & Negation               & Exact match & 0.0  & 16.0 & 68.8          & 50.0           & $6.7{\scriptstyle\pm5.0}$   & $35.3{\scriptstyle\pm8.7}$  & $59.7{\scriptstyle\pm7.1}$  & $70.3{\scriptstyle\pm6.7}$ & $74.7{\scriptstyle\pm3.2}$  & \underline{83.7} & \textbf{87.0} \\
  & Rhymes                 & Exact match & 0.0  & 3.0  & 6.3           & \textbf{25.0}  & $6.7{\scriptstyle\pm3.2}$   & $6.7{\scriptstyle\pm6.0}$   & $9.3{\scriptstyle\pm3.1}$   & $\underline{12.0}{\scriptstyle\pm3.5}$ & $4.7{\scriptstyle\pm1.5}$ & 3.7 & 8.7 \\
\cmidrule(lr){2-14}
  & Average                &             & 0.0  & 36.8 & 67.2          & 68.8           & 48.9      & 58.1      & 56.8      & 65.0     & 69.3      & \underline{70.8} & \textbf{73.3} \\
\midrule
\multirow{6}{*}{Semantics}
  & Antonyms               & Exact match & 0.0  & 43.0 & 62.5          & 25.0           & $47.3{\scriptstyle\pm15.0}$ & $56.0{\scriptstyle\pm19.0}$ & $64.3{\scriptstyle\pm4.9}$  & $37.7{\scriptstyle\pm32.9}$ & $45.7{\scriptstyle\pm18.2}$ & \underline{67.7} & \textbf{74.3} \\
  & Synonyms               & Contains    & 0.0  & 2.0  & 12.5          & \textbf{25.0}  & $11.0{\scriptstyle\pm4.6}$  & $12.7{\scriptstyle\pm3.1}$  & $18.3{\scriptstyle\pm3.1}$  & $22.7{\scriptstyle\pm3.5}$ & $5.7{\scriptstyle\pm5.0}$   & 23.0          & \underline{24.7} \\
  & Taxonomy animal        & Exact set   & 0.0  & 0.0  & 37.5          & 37.5           & $25.3{\scriptstyle\pm25.6}$ & $4.0{\scriptstyle\pm4.6}$   & $26.7{\scriptstyle\pm44.5}$ & $\underline{63.7}{\scriptstyle\pm38.9}$ & $22.7{\scriptstyle\pm3.5}$ & 53.7 & \textbf{93.7} \\
  & Cause and effect       & Exact match & 0.0  & 24.0 & \underline{60.0} & 0.0          & $24.0{\scriptstyle\pm0.0}$  & $48.0{\scriptstyle\pm10.6}$ & $25.3{\scriptstyle\pm22.0}$ & $49.3{\scriptstyle\pm30.0}$ & $52.0{\scriptstyle\pm17.4}$ & 54.7 & \textbf{81.3} \\
  & Common concept         & F1          & 3.0  & 4.0  & \textbf{50.0} & \textbf{50.0}  & $12.8{\scriptstyle\pm0.0}$ & $5.0{\scriptstyle\pm2.1}$ & $2.6{\scriptstyle\pm3.2}$ & $6.7{\scriptstyle\pm0.5}$ & $12.1{\scriptstyle\pm3.9}$ & 3.3 & \underline{13.7} \\
\cmidrule(lr){2-14}
  & Average                &             & 0.6  & 14.6 & \underline{44.5} & 27.5         & 24.1      & 25.1      & 27.5      & 36.0     & 27.6      & 40.5          & \textbf{57.5} \\
\midrule
\multirow{5}{*}{Translation}
  & Num to verbal          & Exact match & 0.0  & 61.0 & 93.8          & \textbf{100.0} & $73.7{\scriptstyle\pm6.4}$  & $83.7{\scriptstyle\pm4.0}$  & $82.7{\scriptstyle\pm18.6}$ & $81.3{\scriptstyle\pm12.7}$ & $82.3{\scriptstyle\pm14.2}$ & 89.0 & \underline{97.7} \\
  & Translation en-es      & Exact match & 0.0  & 15.0 & 25.0          & 25.0           & $33.0{\scriptstyle\pm4.6}$  & $27.3{\scriptstyle\pm15.3}$ & $36.3{\scriptstyle\pm14.2}$ & $35.3{\scriptstyle\pm1.2}$ & $40.0{\scriptstyle\pm4.6}$  & \textbf{43.0} & \underline{40.3} \\
  & Translation en-de      & Exact match & 0.0  & 10.0 & 18.8          & \textbf{50.0}  & $23.7{\scriptstyle\pm3.5}$  & $15.0{\scriptstyle\pm7.0}$  & $17.7{\scriptstyle\pm2.5}$  & $25.0{\scriptstyle\pm2.0}$ & $30.3{\scriptstyle\pm3.5}$ & 29.7 & \underline{32.0} \\
  & Translation en-fr      & Exact match & 0.0  & 7.0  & 6.0           & \textbf{50.0}  & $22.3{\scriptstyle\pm4.2}$  & $11.3{\scriptstyle\pm5.8}$  & $21.3{\scriptstyle\pm6.5}$  & $23.7{\scriptstyle\pm4.2}$ & $35.0{\scriptstyle\pm6.2}$ & 26.3 & \underline{40.3} \\
\cmidrule(lr){2-14}
  & Average                &             & 0.0  & 23.3 & 35.9          & \textbf{56.3}  & 38.2      & 34.3      & 39.5      & 41.3     & 46.9      & 47.0          & \underline{52.6} \\
\midrule
\multirow{4}{*}{GLUE}
  & Sentence similarity    & Exact match & 0.0  & 20.0 & 31.5          & 25.0           & $0.0{\scriptstyle\pm0.0}$   & $0.0{\scriptstyle\pm0.0}$   & $0.0{\scriptstyle\pm0.0}$   & $0.0{\scriptstyle\pm0.0}$  & $17.7{\scriptstyle\pm14.5}$ & \underline{32.7} & \textbf{38.7} \\
  & Word in context        & Exact match & 55.0 & 46.0 & 37.5          & 50.0           & $56.3{\scriptstyle\pm3.5}$  & $60.0{\scriptstyle\pm3.6}$  & $44.3{\scriptstyle\pm11.7}$ & $\textbf{61.0}{\scriptstyle\pm4.6}$ & $58.3{\scriptstyle\pm1.5}$ & \underline{60.3} & \textbf{61.0} \\
  & Sentiment              & Exact match & 50.0 & 83.0 & 68.8          & \textbf{100.0} & $84.7{\scriptstyle\pm1.2}$  & $84.7{\scriptstyle\pm1.2}$  & $70.3{\scriptstyle\pm6.1}$  & $85.0{\scriptstyle\pm1.0}$ & $84.3{\scriptstyle\pm2.3}$  & 86.0 & \underline{91.7} \\
\cmidrule(lr){2-14}
  & Average                &             & 35.0 & 49.7 & 45.9          & 58.3           & 47.0      & 48.2      & 38.2      & 48.7     & 53.4      & \underline{59.7} & \textbf{63.8} \\
\midrule
\multirow{5}{*}{Others}
  & Larger animal          & Exact match & 0.0  & 3.0  & 56.3          & 25.0           & $6.0{\scriptstyle\pm0.0}$   & $47.3{\scriptstyle\pm29.0}$ & $51.7{\scriptstyle\pm5.8}$  & $72.7{\scriptstyle\pm25.8}$ & $48.0{\scriptstyle\pm10.2}$ & \underline{86.7} & \textbf{92.7} \\
  & Informal to formal     & F1          & 0.0  & 27.4 & 42.5          & 24.2           & $24.8{\scriptstyle\pm9.1}$  & $28.3{\scriptstyle\pm3.9}$  & $18.6{\scriptstyle\pm2.9}$  & $34.8{\scriptstyle\pm3.4}$ & $30.0{\scriptstyle\pm9.7}$  & \underline{43.5} & \textbf{47.3} \\
  & Sum                    & Exact match & 0.0  & \underline{99.0} & \textbf{100.0} & \textbf{100.0} & $\textbf{100.0}{\scriptstyle\pm0.0}$ & $\underline{99.0}{\scriptstyle\pm1.0}$ & $\textbf{100.0}{\scriptstyle\pm0.0}$ & $\textbf{100.0}{\scriptstyle\pm0.0}$ & $\textbf{100.0}{\scriptstyle\pm0.0}$ & \textbf{100.0} & \textbf{100.0} \\
  & Diff                   & Exact match & 2.0  & \underline{99.0} & \textbf{100.0} & \textbf{100.0} & $\textbf{100.0}{\scriptstyle\pm0.0}$ & $\underline{97.3}{\scriptstyle\pm1.5}$ & $\textbf{100.0}{\scriptstyle\pm0.0}$ & $\textbf{100.0}{\scriptstyle\pm0.0}$ & $82.0{\scriptstyle\pm23.1}$ & \textbf{100.0} & \textbf{100.0} \\
\cmidrule(lr){2-14}
  & Average                &             & 0.5  & 57.1 & 74.7          & 62.3           & 57.7      & 68.0      & 67.6      & 76.9     & 65.0      & \underline{82.6} & \textbf{85.0} \\
\bottomrule
\end{tabular}
}
\end{sc}
\caption{Category-wise results on the Instruction Induction (II) tasks.}
\label{tab:ii_category_full}
\vspace{\baselineskip}
\end{table*}

\begin{table*}[!htb]
\vspace{\baselineskip}

\vspace{\baselineskip}
\centering
\begin{sc}
\setlength{\tabcolsep}{1.5pt}
\resizebox{\textwidth}{!}{
\begin{tabular}{@{}clcccccccccccc@{}}
\toprule
Type & Subject & Metric & \shortstack{Fewshot\\+Manual} & CoT & APE & ProTeGi & \shortstack{Evo\\Prompt} & \shortstack{Prompt\\Wizard} & GRACE & PRL & GEPA & \shortstack{Stable\\Prompt} & GFlowPO \\
\midrule
STEM
 & Abstract algebra             & Accuracy & 30.0 & 33.0 & 31.0 & \underline{35.0} & 31.0 & 30.0 & 32.0 & 29.0 & \textbf{38.0} & 33.0 & 31.0 \\
 & Anatomy                      & Accuracy & 50.4 & 51.9 & 49.6 & 53.0 & 54.1 & \underline{55.6} & 48.2 & \textbf{56.3} & 54.8 & 53.3 & 52.6 \\
 & Astronomy                    & Accuracy & 57.9 & \textbf{64.5} & 54.0 & 56.6 & 61.2 & 57.9 & 52.6 & 58.6 & 57.2 & \underline{63.2} & 57.9 \\
 & College biology              & Accuracy & 66.7 & \underline{67.4} & 57.0 & 65.8 & 55.0 & 46.0 & 45.0 & 60.4 & \textbf{68.1} & 65.3 & 56.9 \\
 & College chemistry            & Accuracy & 38.0 & 34.0 & 39.0 & 40.0 & \textbf{60.4} & \underline{46.4} & 45.3 & 32.0 & 35.0 & 39.0 & 36.0 \\
 & College computer science     & Accuracy & 41.0 & 48.0 & 32.8 & 37.0 & \textbf{60.4} & 56.3 & \underline{59.0} & 41.0 & 45.0 & 40.0 & 39.0 \\
 & College mathematics          & Accuracy & 32.0 & 34.0 & 33.0 & 33.0 & \textbf{43.0} & 33.0 & \underline{37.0} & 31.0 & 35.0 & 35.0 & \underline{37.0} \\
 & College physics              & Accuracy & 39.2 & 34.3 & 32.3 & 35.3 & 40.0 & \textbf{42.0} & 35.0 & 25.5 & 32.4 & \underline{40.2} & \underline{40.2} \\
 & Computer security            & Accuracy & \underline{70.0} & 67.0 & 62.2 & 67.0 & 28.0 & 28.0 & 32.0 & 57.0 & \textbf{74.0} & 64.0 & 61.0 \\
 & Conceptual physics           & Accuracy & 51.1 & \underline{55.3} & 51.1 & 49.8 & \textbf{55.5} & 39.3 & 49.1 & 48.1 & 50.2 & 52.3 & 48.1 \\
 & Electrical engineering       & Accuracy & 51.7 & \underline{55.2} & 46.2 & 40.0 & 39.2 & 33.3 & 25.5 & 42.8 & \textbf{58.6} & \textbf{58.6} & 53.8 \\
 & Elementary mathematics       & Accuracy & 38.9 & 60.1 & 38.1 & 37.3 & \textbf{66.0} & \underline{63.0} & 60.0 & 39.7 & 39.4 & 41.0 & 38.4 \\
 & High school biology          & Accuracy & 70.7 & 64.5 & 65.8 & 69.8 & 47.7 & 35.3 & 47.7 & 64.5 & \textbf{72.3} & \underline{71.9} & 70.3 \\
 & High school chemistry        & Accuracy & \textbf{52.7} & \textbf{52.7} & \underline{52.2} & 45.8 & 29.0 & 28.1 & 29.0 & 50.7 & 49.8 & 50.3 & 47.8 \\
 & High school computer science & Accuracy & 61.0 & 58.0 & 54.0 & 51.0 & 53.1 & 37.9 & 46.2 & \underline{61.0} & \textbf{63.0} & 56.0 & \underline{61.0} \\
 & High school mathematics      & Accuracy & 36.3 & 33.7 & \underline{38.5} & 33.0 & \textbf{40.0} & 34.4 & 37.6 & \underline{38.5} & 35.6 & 35.2 & 37.0 \\
 & High school physics          & Accuracy & 26.5 & 31.1 & 32.5 & 33.8 & \underline{45.2} & 31.8 & 31.0 & 30.5 & \textbf{50.2} & 31.8 & 31.1 \\
 & High school statistics       & Accuracy & 45.4 & 43.5 & \underline{46.8} & \textbf{50.5} & 36.0 & 22.0 & 33.0 & 41.7 & 43.5 & 39.8 & 45.4 \\
 & Machine learning             & Accuracy & 35.7 & 46.4 & 39.3 & 35.7 & \underline{60.7} & 59.7 & \textbf{70.3} & 33.9 & 34.8 & 38.4 & 40.2 \\
\midrule
Social science
 & Econometrics                      & Accuracy & 32.5 & 34.2 & 32.5 & 31.6 & \textbf{48.3} & 18.2 & 31.0 & 30.7 & 33.3 & \underline{37.7} & 36.0 \\
 & High school geography               & Accuracy & 66.7 & 61.1 & 56.6 & 59.7 & 60.0 & 58.0 & 62.0 & 57.6 & \textbf{69.2} & \underline{68.7} & \textbf{69.2} \\
 & High school government and politics & Accuracy & 74.1 & 76.2 & 67.9 & 70.9 & 63.6 & 52.1 & 57.0 & 67.4 & 76.2 & \textbf{80.3} & \underline{78.2} \\
 & High school macroeconomics          & Accuracy & 54.1 & 55.1 & 50.0 & 56.2 & \textbf{60.6} & 56.6 & \underline{58.1} & 39.2 & 56.9 & 54.1 & 52.8 \\
 & High school microeconomics          & Accuracy & 55.5 & 55.5 & 53.4 & 56.2 & \textbf{68.9} & 51.8 & 42.5 & 55.0 & 30.5 & 58.4 & \underline{59.3} \\
 & High school psychology              & Accuracy & \textbf{76.3} & 73.6 & 71.2 & 72.7 & 52.3 & 52.3 & 43.3 & \underline{75.1} & 72.7 & 74.5 & \underline{75.1} \\
 & Human sexuality                     & Accuracy & 62.6 & 52.8 & 61.1 & 58.8 & 34.4 & 21.5 & 27.4 & 62.6 & \textbf{71.1} & \underline{69.5} & 68.7 \\
 & Professional psychology             & Accuracy & 51.8 & 53.4 & 49.5 & 48.1 & \textbf{54.6} & 48.3 & 44.5 & 50.5 & 53.8 & 52.3 & \textbf{54.6} \\
 & Public relations                    & Accuracy & 60.0 & 54.6 & 63.6 & 59.1 & 32.5 & 27.2 & 27.8 & 60.0 & 60.0 & \textbf{68.2} & \underline{66.4} \\
 & Security studies                    & Accuracy & 50.2 & 48.6 & 52.2 & 47.4 & \underline{73.2} & \textbf{73.8} & 67.9 & 55.5 & 57.1 & 58.8 & 53.5 \\
 & Sociology                           & Accuracy & 66.2 & 67.2 & 65.2 & 70.7 & 46.8 & 41.7 & 39.4 & 68.7 & \textbf{78.1} & 70.7 & \underline{75.6} \\
 & US foreign policy                   & Accuracy & 75.0 & 69.0 & \underline{76.0} & 73.0 & 59.8 & 52.0 & 63.2 & 70.0 & \textbf{78.0} & 73.0 & 72.0 \\
\midrule
Humanities
 & Formal logic                  & Accuracy & 37.3 & 38.1 & 36.5 & 33.3 & \underline{67.5} & \textbf{68.4} & 62.5 & 46.0 & 42.9 & 32.5 & 36.5 \\
 & High school European history  & Accuracy & 63.6 & 57.6 & 62.4 & 65.5 & 59.2 & 53.8 & 50.2 & 63.6 & 64.2 & \textbf{67.9} & \underline{65.5} \\
 & High school US history        & Accuracy & 62.8 & 56.9 & 65.2 & 55.4 & 64.9 & 45.8 & 59.5 & 61.3 & 64.2 & \textbf{70.1} & \underline{68.1} \\
 & High school world history     & Accuracy & 68.4 & 67.5 & 71.2 & 64.1 & 66.9 & 52.9 & 57.0 & 64.6 & 71.3 & \underline{73.4} & \textbf{73.8} \\
 & International law             & Accuracy & 62.0 & 65.3 & 64.5 & 66.1 & 54.6 & 35.2 & 63.9 & 65.3 & \textbf{71.1} & \underline{69.4} & \textbf{71.1} \\
 & Jurisprudence                 & Accuracy & 57.4 & 63.9 & 62.0 & 62.0 & 66.3 & 59.5 & 39.9 & 56.5 & 65.7 & \underline{66.7} & \textbf{71.3} \\
 & Logical fallacies             & Accuracy & 63.2 & 65.0 & \underline{68.1} & 66.9 & 36.6 & 34.8 & 37.5 & 62.6 & \textbf{69.3} & 66.3 & 65.0 \\
 & Moral disputes                & Accuracy & 49.7 & 51.2 & 59.0 & 55.5 & 57.3 & 54.4 & 53.4 & 43.9 & \textbf{61.3} & \underline{59.8} & 58.7 \\
 & Moral scenarios               & Accuracy & 24.4 & 27.9 & 27.3 & 29.3 & \textbf{77.4} & 69.2 & \underline{70.1} & 26.0 & 31.3 & 29.1 & 28.3 \\
 & Philosophy                    & Accuracy & 56.9 & 54.7 & 54.7 & 57.2 & \underline{58.0} & \underline{58.0} & 56.0 & 55.3 & \textbf{60.8} & 54.7 & 56.3 \\
 & Prehistory                    & Accuracy & 60.5 & 52.2 & 58.6 & 56.2 & \textbf{71.8} & \underline{70.8} & 65.9 & 57.7 & 57.7 & 55.3 & 57.1 \\
 & Professional law              & Accuracy & 40.6 & 38.5 & 32.0 & 42.0 & \underline{54.1} & \textbf{56.4} & 43.1 & 40.9 & 39.4 & 39.8 & 41.2 \\
 & World religions               & Accuracy & 73.7 & 69.6 & 71.9 & \underline{74.3} & 25.6 & 26.6 & 24.7 & 73.7 & \textbf{76.0} & 73.7 & 73.7 \\
\midrule
Others
 & Business ethics         & Accuracy & 47.0 & \textbf{63.0} & 55.0 & 51.0 & \underline{60.5} & 57.8 & 55.2 & 53.0 & 57.0 & 58.0 & 56.0 \\
 & Clinical knowledge      & Accuracy & 54.3 & 56.6 & 51.2 & 51.7 & 54.0 & 42.8 & 51.5 & 47.6 & 61.1 & \underline{62.3} & \textbf{62.6} \\
 & College medicine        & Accuracy & 54.3 & 53.2 & 46.9 & 49.7 & \textbf{59.6} & 57.4 & 47.8 & 47.4 & \underline{57.8} & \underline{57.8} & 53.2 \\
 & Global facts            & Accuracy & 32.0 & \underline{39.0} & 32.0 & 35.0 & \textbf{41.1} & 24.1 & 33.7 & 37.0 & 37.0 & 37.0 & 34.0 \\
 & Human aging             & Accuracy & 56.5 & 55.6 & 58.7 & 58.3 & 40.7 & 40.0 & 34.0 & \underline{60.5} & \textbf{61.0} & 54.7 & 58.7 \\
 & Management              & Accuracy & 61.2 & 64.1 & 63.1 & 60.2 & 49.6 & 46.3 & 27.2 & 63.1 & \textbf{70.9} & \textbf{70.9} & \underline{69.9} \\
 & Marketing               & Accuracy & 75.6 & 80.3 & 76.9 & 77.4 & 51.6 & 48.0 & 45.3 & 76.1 & 79.9 & \textbf{82.9} & \underline{81.2} \\
 & Medical genetics        & Accuracy & 54.0 & 55.0 & 55.0 & 57.0 & \textbf{60.9} & 53.6 & 48.2 & 46.0 & 57.0 & \underline{59.0} & \underline{59.0} \\
 & Miscellaneous           & Accuracy & 73.3 & \underline{74.2} & 72.4 & 72.8 & 56.3 & 52.7 & 53.1 & 68.7 & \textbf{75.1} & 70.8 & 72.2 \\
 & Nutrition               & Accuracy & 59.2 & 53.6 & 56.9 & 62.1 & \textbf{67.7} & \underline{67.2} & 59.7 & 61.8 & 62.8 & 64.4 & 66.3 \\
 & Professional accounting & Accuracy & 40.1 & 41.5 & 46.5 & 42.9 & \textbf{75.0} & \underline{72.0} & \underline{72.0} & 39.0 & 44.3 & 40.8 & 39.4 \\
 & Professional medicine   & Accuracy & \textbf{55.2} & 45.2 & 0.5 & 50.7 & 48.2 & 43.4 & 28.9 & 50.0 & 50.4 & 44.1 & \underline{52.2} \\
 & Virology                & Accuracy & 46.4 & 47.2 & 48.8 & 50.0 & \textbf{76.0} & 71.4 & \underline{72.5} & 48.2 & 48.2 & 49.4 & 51.2 \\
\bottomrule
\end{tabular}
}
\end{sc}
\caption{Full experimental results on the MMLU.}
\label{tab:appen_mmlu_full_results}
\vspace{\baselineskip}
\end{table*}

\begin{table}[!t]

\makebox[\columnwidth][l]{%
\begin{minipage}{\textwidth}

\centering

\captionsetup{
    justification=centering,
    singlelinecheck=false
}

\vspace{0.5em}

\small
\scshape
\setlength{\tabcolsep}{3pt}
\renewcommand{\arraystretch}{1.15}

\resizebox{\textwidth}{!}{%
\begin{tabular}{@{}lccccccc@{}}

\toprule

Task
& \shortstack{Evo\\Prompt}
& \shortstack{Prompt\\Wizard}
& GRACE
& PRL
& GEPA
& \shortstack{Stable\\Prompt}
& GFlowPO \\

\midrule

Average accuracy
& \underline{80.9}
& 78.0
& 79.4
& 76.4
& 80.1
& 76.7
& \textbf{82.8} \\

\midrule

Tracking shuffled objects five objects
& $81.7{\scriptstyle\pm1.5}$
& $79.3{\scriptstyle\pm1.5}$
& $83.3{\scriptstyle\pm12.1}$
& $70.3{\scriptstyle\pm3.1}$
& $\textbf{90.0}{\scriptstyle\pm1.0}$
& $71.3{\scriptstyle\pm9.8}$
& $\underline{83.7}{\scriptstyle\pm2.1}$ \\

Object counting
& $\underline{97.3}{\scriptstyle\pm0.6}$
& $95.7{\scriptstyle\pm0.6}$
& $96.0{\scriptstyle\pm0.0}$
& $95.7{\scriptstyle\pm3.5}$
& $97.0{\scriptstyle\pm0.0}$
& $92.3{\scriptstyle\pm5.5}$
& $\textbf{97.7}{\scriptstyle\pm2.1}$ \\

Date understanding
& $82.3{\scriptstyle\pm1.5}$
& $\underline{83.7}{\scriptstyle\pm2.5}$
& $\underline{83.7}{\scriptstyle\pm3.8}$
& $80.7{\scriptstyle\pm1.5}$
& $\textbf{89.7}{\scriptstyle\pm0.6}$
& $80.7{\scriptstyle\pm4.0}$
& $82.0{\scriptstyle\pm0.0}$ \\

Disambiguation qa
& $67.7{\scriptstyle\pm3.1}$
& $69.7{\scriptstyle\pm5.5}$
& $64.3{\scriptstyle\pm9.6}$
& $\textbf{74.0}{\scriptstyle\pm3.6}$
& $69.3{\scriptstyle\pm2.9}$
& $66.0{\scriptstyle\pm7.6}$
& $\underline{72.7}{\scriptstyle\pm0.6}$ \\

Formal fallacies
& $\textbf{87.3}{\scriptstyle\pm2.5}$
& $76.0{\scriptstyle\pm5.6}$
& $\underline{82.0}{\scriptstyle\pm2.7}$
& $75.7{\scriptstyle\pm2.9}$
& $81.3{\scriptstyle\pm0.6}$
& $74.0{\scriptstyle\pm3.6}$
& $79.0{\scriptstyle\pm2.7}$ \\

Geometric shapes
& $\textbf{56.0}{\scriptstyle\pm2.7}$
& $45.7{\scriptstyle\pm2.3}$
& $47.0{\scriptstyle\pm1.7}$
& $39.0{\scriptstyle\pm1.0}$
& $42.0{\scriptstyle\pm5.2}$
& $42.3{\scriptstyle\pm5.5}$
& $\underline{54.0}{\scriptstyle\pm2.7}$ \\

Salient translation error detection
& $\textbf{70.0}{\scriptstyle\pm0.0}$
& $63.3{\scriptstyle\pm3.8}$
& $\underline{68.3}{\scriptstyle\pm1.2}$
& $66.7{\scriptstyle\pm1.2}$
& $67.3{\scriptstyle\pm4.7}$
& $66.3{\scriptstyle\pm3.2}$
& $66.7{\scriptstyle\pm1.2}$ \\

Penguins in a table
& $94.8{\scriptstyle\pm4.5}$
& $96.2{\scriptstyle\pm2.6}$
& $93.1{\scriptstyle\pm3.9}$
& $88.9{\scriptstyle\pm2.2}$
& $\underline{97.6}{\scriptstyle\pm2.4}$
& $93.4{\scriptstyle\pm3.4}$
& $\textbf{98.3}{\scriptstyle\pm1.6}$ \\

Causal judgement
& $\underline{59.7}{\scriptstyle\pm0.6}$
& $58.3{\scriptstyle\pm2.3}$
& $56.0{\scriptstyle\pm2.7}$
& $58.7{\scriptstyle\pm4.2}$
& $57.3{\scriptstyle\pm0.6}$
& $59.0{\scriptstyle\pm1.7}$
& $\textbf{61.3}{\scriptstyle\pm0.6}$ \\

Logical deduction five objects
& $\underline{72.3}{\scriptstyle\pm0.6}$
& $69.3{\scriptstyle\pm3.5}$
& $70.3{\scriptstyle\pm3.8}$
& $67.7{\scriptstyle\pm1.5}$
& $65.3{\scriptstyle\pm0.6}$
& $70.7{\scriptstyle\pm0.6}$
& $\textbf{76.7}{\scriptstyle\pm2.1}$ \\

Movie recommendation
& $68.7{\scriptstyle\pm0.6}$
& $67.7{\scriptstyle\pm0.6}$
& $66.7{\scriptstyle\pm1.5}$
& $67.3{\scriptstyle\pm0.6}$
& $\underline{70.3}{\scriptstyle\pm1.2}$
& $66.3{\scriptstyle\pm4.2}$
& $\textbf{79.0}{\scriptstyle\pm5.2}$ \\

Navigate
& $\textbf{93.0}{\scriptstyle\pm1.0}$
& $90.0{\scriptstyle\pm1.7}$
& $89.7{\scriptstyle\pm1.5}$
& $84.3{\scriptstyle\pm4.2}$
& $89.7{\scriptstyle\pm0.6}$
& $87.7{\scriptstyle\pm4.0}$
& $\underline{91.0}{\scriptstyle\pm1.0}$ \\

Sports understanding
& $73.3{\scriptstyle\pm0.6}$
& $78.0{\scriptstyle\pm3.5}$
& $\textbf{88.3}{\scriptstyle\pm2.5}$
& $72.0{\scriptstyle\pm5.3}$
& $83.7{\scriptstyle\pm0.6}$
& $77.7{\scriptstyle\pm5.7}$
& $\underline{88.0}{\scriptstyle\pm1.0}$ \\

Reasoning about colored objects
& $93.3{\scriptstyle\pm1.2}$
& $93.0{\scriptstyle\pm1.0}$
& $93.3{\scriptstyle\pm2.1}$
& $88.7{\scriptstyle\pm2.1}$
& $\underline{94.0}{\scriptstyle\pm1.7}$
& $\underline{94.0}{\scriptstyle\pm3.0}$
& $\textbf{95.3}{\scriptstyle\pm2.1}$ \\

Hyperbaton
& $\textbf{99.3}{\scriptstyle\pm0.6}$
& $97.3{\scriptstyle\pm1.2}$
& $97.3{\scriptstyle\pm1.5}$
& $96.3{\scriptstyle\pm2.5}$
& $98.3{\scriptstyle\pm0.6}$
& $98.3{\scriptstyle\pm1.2}$
& $\underline{99.0}{\scriptstyle\pm0.0}$ \\

Ruin names
& $\underline{71.7}{\scriptstyle\pm2.5}$
& $63.3{\scriptstyle\pm4.0}$
& $71.0{\scriptstyle\pm5.0}$
& $68.0{\scriptstyle\pm6.1}$
& $64.0{\scriptstyle\pm1.0}$
& $56.7{\scriptstyle\pm7.6}$
& $\textbf{73.0}{\scriptstyle\pm1.7}$ \\

Snarks
& $82.3{\scriptstyle\pm2.1}$
& $78.5{\scriptstyle\pm0.7}$
& $78.3{\scriptstyle\pm1.5}$
& $78.0{\scriptstyle\pm3.6}$
& $\underline{83.0}{\scriptstyle\pm3.0}$
& $79.0{\scriptstyle\pm5.3}$
& $\textbf{85.3}{\scriptstyle\pm2.5}$ \\

Temporal sequences
& $88.3{\scriptstyle\pm3.2}$
& $82.3{\scriptstyle\pm0.6}$
& $\underline{88.7}{\scriptstyle\pm4.0}$
& $86.7{\scriptstyle\pm5.9}$
& $88.3{\scriptstyle\pm1.2}$
& $87.7{\scriptstyle\pm5.9}$
& $\textbf{94.7}{\scriptstyle\pm3.5}$ \\

Boolean expressions
& $\textbf{97.0}{\scriptstyle\pm1.0}$
& $94.7{\scriptstyle\pm3.1}$
& $92.0{\scriptstyle\pm0.0}$
& $92.7{\scriptstyle\pm4.9}$
& $94.0{\scriptstyle\pm6.6}$
& $94.3{\scriptstyle\pm4.7}$
& $\underline{96.0}{\scriptstyle\pm1.7}$ \\

\bottomrule

\end{tabular}%
}
\caption{Full experimental results on Big-Bench Hard (BBH).}
\label{tab:appen_bbh_performance}

\end{minipage}%
}

\end{table}

\clearpage

\section{Dataset Details on Reasoning-Centric Tasks}
\label{appen:dataset_details_reasoning}

\textbf{BBH.} The benchmark originally consists of 27 tasks. Starting from the full 27-task set, we first followed prior works using the five-object versions of logical deduction and tracking shuffled objects as representative variants, reducing the evaluation set into 23 tasks. In addition, we adopted the evaluation procedure from \citet{das2025greatergradientsreasoningmakes}, where the LM first generates reasoning and then an extractor phrase (e.g., "The final answer is ...") is used to obtain the final answer. Under this setup, we found that for a certain subset of BBH tasks, Qwen3-8B frequently produced final answers inconsistent with reasoning or difficult-to-parse final answers, making automatic evaluation less reliable. For these reasons, we further restricted the evaluation to a 19-task subset. We evaluated on 19 tasks using 50 training samples and 100 test samples per task. 

\textbf{GSM8K.} We used 100 training samples and evaluated on the full test set of 1319 samples.

\textbf{AIME.} We used AIME 2022-2024 problems as training and validation data for prompt optimization. The learning process was executed among randomly split 45 train data and 45 validation data. The learned prompt was evaluated on AIME-2025 (30 problems), evaluating 5 runs per problem (150 generations total).  

\section{Learning Curves on BBH Tasks}
\label{appen:learning_curves_bbh}

\begin{figure*}[htbp]
    \centering
    \includegraphics[width=1.0\textwidth]{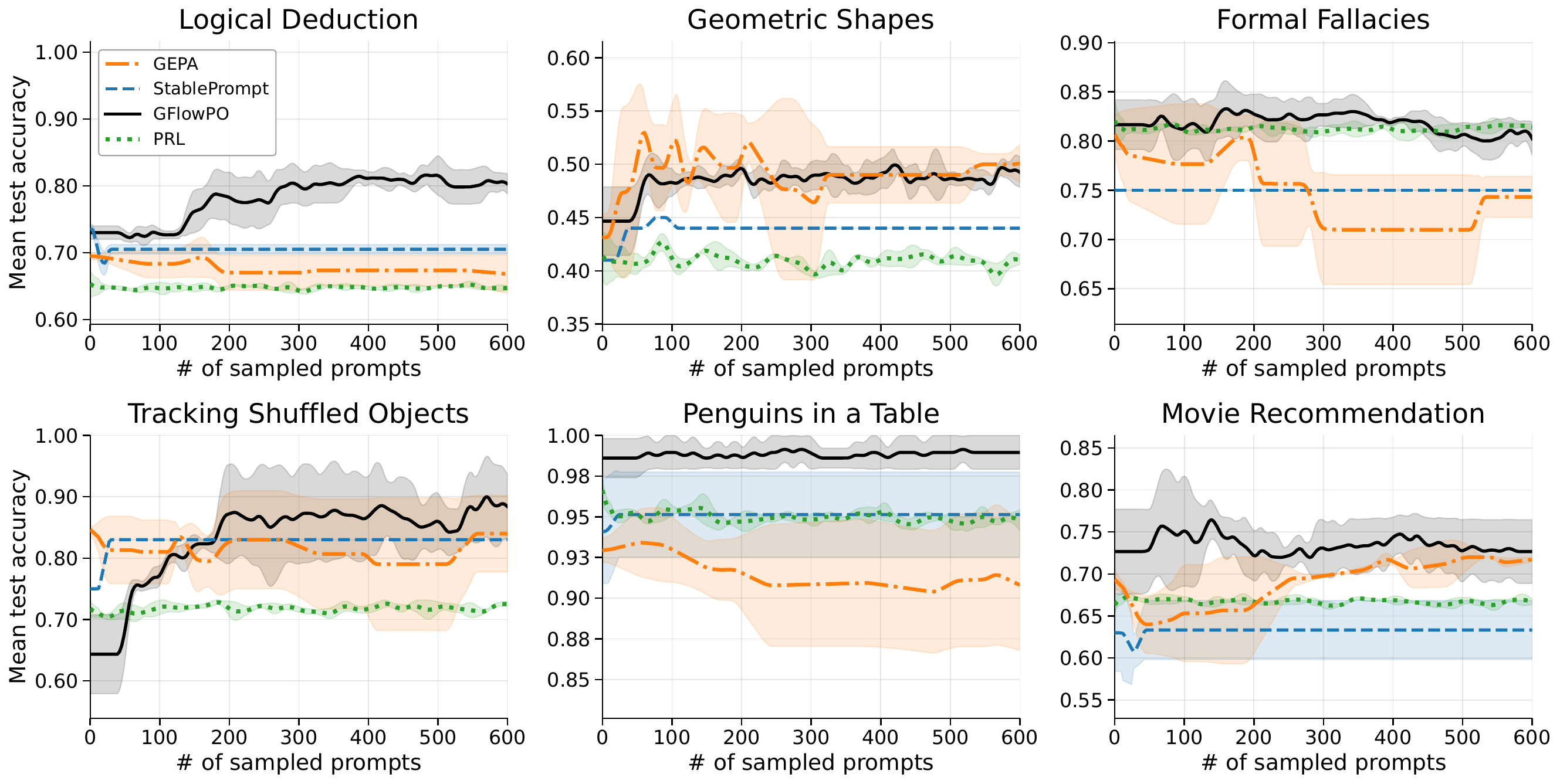}
    \caption{Learning curves of \textsc{GFlowPO}, StablePrompt, PRL, and GEPA. Given the same prompt sampling budgets, \textsc{GFlowPO} consistently shows higher sample efficiency in average test accuracy.}
    \label{fig:appen_lcbbh}
\end{figure*}

In this section, we report the learning curves of \textsc{GFlowPO}, RL-based prompt optimization methods including StablePrompt and PRL, as well as GEPA on 6 BBH tasks. As shown in \Cref{fig:appen_lcbbh}, under the same prompt-sampling budget, \textsc{GFlowPO} consistently achieves higher sample efficiency and average test accuracy than other baselines, except Geometric Shapes and Formal Fallacies. While GEPA attains higher performance than \textsc{GFlowPO} on Geometric Shapes, its performance remains high only in the early stage and becomes less stable as optimization proceeds. On Formal Fallacies, PRL shows slightly better performance than \textsc{GFlowPO}; however, its prompt performance does not substantially improve from the initial stage, suggesting limited exploration. In contrast, even when \textsc{GFlowPO} starts with relatively lower-performing prompts, it continues to improve throughout optimization. This indicates that off-policy learning enables \textsc{GFlowPO} to sustain exploration and discover prompts in high-reward regions over time.

\section{Early-stage performance and comparison with GEPA.}
\label{appen:early_stage_comparison_gepa}

The learning curves in \Cref{fig:gfnpo_exploration} should be interpreted in light of the Pre-Step protocol. The first 400 prompt evaluations of \textsc{GFlowPO} correspond to the Pre-Step, during which prompts are sampled only to initialize the replay buffer $\mathcal{B}$ and the prompt-LM is not updated. Consequently, no learning-driven improvement is expected in this interval, and methods that start optimizing from the first evaluation, such as GEPA, may appear competitive in the early stage of the curve. However, this early advantage does not persist: on Letters List, for instance, GEPA reaches a final test accuracy of 77.0, whereas \textsc{GFlowPO} achieves 95.3. Moreover, as shown in \Cref{tab:prestep_reference_prompt_sampling_ablation}, removing the Pre-Step lowers the average accuracy from 80.2 to 74.3, indicating that replay initialization is important for stable off-policy optimization. \textsc{GFlowPO} thus trades a small amount of early-stage sample efficiency for substantially better final performance. This trade-off also suggests a hybrid strategy in which GEPA is used to initialize the replay buffer and \textsc{GFlowPO} performs the subsequent optimization, potentially improving early-stage sample efficiency further; we leave this direction to future work.

\vspace{-0.5em}    
\section{Contribution of GFlowNet Learning}
\label{appen:gfn_contribution}
\vspace{-0.5em}    
The ablation in \Cref{sec:ablation_study_exploration} shows that DMU accounts for a substantial portion of the overall improvement. To isolate the contribution of GFlowNet-specific learning, we compare \textsc{GFlowPO} with two simpler alternatives, each evaluated with and without DMU.
\begin{itemize}[leftmargin=*, itemsep=2pt, topsep=2pt]
    \item \textbf{Frozen Prompt-LM.} The parameters of the prompt-LM are not updated. Without DMU, this variant serves as a generation-only control; with DMU, it becomes a retrieval-only baseline.
    \item \textbf{Reward-weighted off-policy MLE.} The off-policy GFlowNet objective is replaced with a reward-weighted maximum-likelihood objective over the same replay data, assigning larger weights to higher-reward prompts.
\end{itemize}
All other settings are identical to those in \Cref{sec:induction_qa}. The results are summarized in \Cref{tab:appen_gfn_contribution}. This extended ablation separates three possible sources of improvement: dynamic retrieval, generic reward-weighted replay learning, and GFlowNet-specific learning.

\begin{table}[!htb]
\centering
\begin{sc}
\footnotesize
\resizebox{\columnwidth}{!}{%
\begin{tabular}{@{}lcc@{}}
\toprule
Method & w/o DMU & w/ DMU \\
\midrule
Frozen Prompt-LM               & 69.9 & 74.4 \\
Reward-weighted off-policy MLE & 73.3 & 74.4 \\
On-policy GFlowNet             & 74.4 & \underline{78.0} \\
Off-policy GFlowNet            & \textbf{75.2} & \textbf{80.2} \\
\bottomrule
\end{tabular}%
}
\end{sc}
\caption{Average accuracy of different prompt-LM training strategies with and without DMU.}
\label{tab:appen_gfn_contribution}
\end{table}
\vspace{-0.5em}   
\paragraph{Dynamic retrieval is useful but insufficient on its own.}
Enabling DMU for the frozen prompt-LM improves accuracy from 69.9 to 74.4 (+4.5 points). However, this retrieval-only variant remains 5.8 points below the full \textsc{GFlowPO} (80.2), showing that retrieval alone cannot account for the full gain.

\paragraph{Generic reward-weighted replay learning does not recover the performance of \textsc{GFlowPO}.}
Reward-weighted off-policy MLE achieves 73.3 without DMU and 74.4 with DMU. Under otherwise matched conditions, off-policy GFlowNet training outperforms it by 1.9 points without DMU (75.2 vs.\ 73.3) and by 5.8 points with DMU (80.2 vs.\ 74.4). Off-policy GFlowNet training also consistently outperforms its on-policy counterpart (75.2 vs.\ 74.4 without DMU and 80.2 vs.\ 78.0 with DMU), highlighting the benefit of reusing past prompt evaluations.

\paragraph{DMU and GFlowNet learning are complementary.}
DMU improves off-policy GFlowNet training by 5.0 points (75.2 $\rightarrow$ 80.2), compared with only 1.1 points for reward-weighted MLE (73.3 $\rightarrow$ 74.4). Notably, the advantage of GFlowNet learning becomes larger when DMU is enabled.

Overall, these results indicate that DMU is an effective retrieval mechanism but does not substitute for GFlowNet-specific learning; the best performance is achieved by combining off-policy GFlowNet training with DMU.

\section{Effect of using large LM as target LM}
\label{appen:large_llm}

To investigate whether prompt optimization remains useful for very large LLMs, we conduct additional experiments with GPT-4.1 as the target LM on six BBH tasks: Disambiguation QA, Salient Translation Error Detection, Causal Judgement, Geometric Shapes, Movie Recommendation, and Sports Understanding. These tasks were selected solely based on the headroom of GPT-4.1 zero-shot performance, i.e., tasks where GPT-4.1 does not achieve near-perfect zero-shot accuracy, before running any prompt optimizer. We compare Zero-shot, Few-shot, Manual Prompt, StablePrompt, GEPA, and \textsc{GFlowPO} under the same experimental setting as in \Cref{sec:reasoning}.

As shown in \Cref{tab:appen_gpt41_results}, \textsc{GFlowPO} achieves the highest average accuracy (77.5), outperforming Manual Prompt (71.2), Few-shot (71.5), StablePrompt (72.3), and GEPA (76.8). \Cref{tab:appen_gpt41_prompts} compares the manual prompts with the prompts discovered by \textsc{GFlowPO} on three tasks. These results indicate that even for a very large proprietary model such as GPT-4.1, carefully handcrafted prompts remain suboptimal, while \textsc{GFlowPO} can discover more effective task-specific prompts.

\begin{table}[!htb]

\centering
\begin{sc}
\setlength{\tabcolsep}{3pt}
\resizebox{\columnwidth}{!}{
\begin{tabular}{@{}lcccccc@{}}
\toprule
Method & \shortstack{Zero\\shot} & \shortstack{Few\\shot} & Manual & \shortstack{Stable\\Prompt} & GEPA & GFlowPO \\
\midrule
Avg. (6 tasks) & 71.2 & 71.5 & 71.2 & $72.3{\scriptstyle\pm0.7}$ & $\underline{76.8}{\scriptstyle\pm1.8}$ & $\textbf{77.5}{\scriptstyle\pm1.4}$ \\
\bottomrule
\end{tabular}
}
\end{sc}
\caption{Average accuracy on six BBH tasks with GPT-4.1 as the target LM.}
\label{tab:appen_gpt41_results}
\end{table}

\begin{table*}[!htb]
\centering
\footnotesize
\renewcommand{\arraystretch}{1.2}
\begin{tabular}{@{}l>{\raggedright\arraybackslash}p{0.27\textwidth}>{\raggedright\arraybackslash}p{0.27\textwidth}>{\raggedright\arraybackslash}p{0.27\textwidth}@{}}
\toprule
& \textbf{Disambiguation QA} & \textbf{Geometric Shapes} & \textbf{Movie Recommendation} \\
\midrule
\textbf{Manual}
& Identify what the pronoun refers to. Weigh each possible antecedent against the sentence; if the reference is genuinely ambiguous, say so rather than forcing a choice. \newline (\textbf{Acc: 64.0})
& Treat the SVG path as a figure defined by vertices and line segments. Trace the path, count the distinct vertices and edges, and identify the shape. \newline (\textbf{Acc: 65.0})
& Recall the genre, tone, era, and reception of the listed movies. Compare each option against these shared traits and pick the most similar one. \newline (\textbf{Acc: 80.0}) \\
\midrule
\textbf{\textsc{GFlowPO}}
& For each sentence, identify the antecedent of the pronoun (which noun phrase it refers to) and select the correct option. The pronoun refers to the subject of the sentence unless context clearly indicates otherwise. If the antecedent is ambiguous or unclear, choose (C) Ambiguous. Provide only the letter (A, B, or C) as the output. \newline (\textbf{Acc: 67.0})
& Identify the geometric shape described by the SVG path element. Count the number of distinct vertices connected by line segments to determine the shape. If the shape has 6 vertices, output `C' for hexagon. If it has 8 vertices, output `F' for octagon. If it contains an arc and is not a polygon, output `K' for ellipse. If it has 5 vertices, output `G' for pentagon. If it contains an arc and forms a sector, output `I'. If it contains an arc and forms a kite, output `D'. If it contains an arc and forms a circle, output `A'. If it has 7 vertices, output `B' for heptagon. If it \ldots \newline (\textbf{Acc: 88.0})
& Select the movie from the options that is most similar in theme, genre, or narrative to the listed films, with a focus on critical acclaim and shared storytelling elements. Output the letter corresponding to the most appropriate choice. \newline (\textbf{Acc: 88.0}) \\
\bottomrule
\end{tabular}
\caption{Comparison of manual prompts and prompts discovered by \textsc{GFlowPO} with GPT-4.1 as the target LM. Test accuracy is reported in parentheses.}
\label{tab:appen_gpt41_prompts}
\end{table*}

\vspace{-1em}   
\section{Examples of Manual Prompts}
\label{appen:manual_prompt_examples}
\vspace{-0.5em}   
\Cref{tab:appen_manual_vs_gflowpo} presents the full Manual and \textsc{GFlowPO}-optimized prompts for one II task and one BBII task from \Cref{sec:induction_qa}: Sentence Similarity and Hyperbaton. The Manual Prompts are concise fixed task descriptions, whereas \textsc{GFlowPO} generates its prompts through optimization over task-specific training examples.

As shown in \Cref{tab:appen_manual_vs_gflowpo}, the optimized prompts do more than simply paraphrase the Manual Prompts. Instead, \textsc{GFlowPO} infers task-specific rules that remain implicit in the original instructions, producing more explicit and executable prompts that include reasoning procedures, decision criteria, or transformation rules. These qualitative differences are consistent with the substantial performance improvements over the manually designed prompts.

\begin{table*}[!htb]
\centering
\footnotesize
\renewcommand{\arraystretch}{1.2}
\begin{tabular}{@{}l>{\raggedright\arraybackslash}p{0.41\textwidth}>{\raggedright\arraybackslash}p{0.41\textwidth}@{}}
\toprule
& \textbf{Sentence Similarity} & \textbf{Hyperbaton} \\
\midrule
\textbf{Manual}
& Rate from 0 (definitely not) to 5 (perfectly) the degree in which both sentences describe the same event. \newline (\textbf{Acc: 20.0})
& Which sentence has the correct adjective order? \newline (\textbf{Acc: 56.5}) \\
\midrule
\textbf{\textsc{GFlowPO}}
& I will provide you with two sentences. Based on the given sentences, determine the level of similarity between them. The output should be one of the following numbers: 0 - definitely not 1 - probably not 2 - possibly 3 - probably 4 - almost perfectly Based on the provided input-output pairs, it seems that your friend correctly identified the level of similarity between the two sentences for each pair. \newline (\textbf{Acc: 50.0})
& ``Rearrange the adjective and the noun in each input.'' based on this instruction, the friend's output for each input is correct. They have successfully rearranged the order of the adjective and the noun in each input to match the expected output. For example, with the first input: Input: a ``gold awful knife'' b ``awful gold knife'' Output: A ``awful gold knife'' Your friend has correctly rearranged the order of the adjective ``awful'' and the noun ``gold'' to match the expected output ``awful gold''. Similarly, for the other inputs, the friend has also rearranged the order of the adjective and the noun. \newline (\textbf{Acc: 73.0}) \\
\bottomrule
\end{tabular}
\caption{Comparison of Manual Prompts and prompts discovered by \textsc{GFlowPO} on Sentence Similarity (II) and Hyperbaton (BBII). Test accuracy is reported in parentheses.}
\label{tab:appen_manual_vs_gflowpo}
\end{table*}

\section{Prompt Evaluation}
\label{appen:prompt_eval}

Generated prompts are evaluated using the template \textit{``[prompt] Input: [input] Output:''}. We randomly sampled the training and test dataset, for each benchmark which are detailed in Appendix~\ref{appen:few_shot_data_stat}, \ref{appen:dataset_details_bbii_ii}, \ref{appen:dataset_details_mmlu}, and \ref{appen:dataset_details_reasoning}. We follow the same evaluation protocol in StablePrompt (Kwon et al., 2024).

\section{Sensitivity to DMU Hyperparameters}
\label{appen:dmu_sensitivity}

The main hyperparameter of DMU is the number of reference prompts $k = k_b + k_q$, where $k_b$ and $k_q$ denote the number of prompts sampled from the replay buffer $\mathcal{B}$ and the high-reward buffer $\mathcal{Q}$, respectively. As shown in \Cref{tab:prestep_reference_prompt_sampling_ablation}, with $k=3$ fixed, sampling reference prompts exclusively from either the replay buffer ($k_b=3, k_q=0$; 78.4) or the high-reward buffer ($k_b=0, k_q=3$; 74.5) degrades performance compared to using both sources (80.2). This indicates that combining both sources is important, as it balances exploration (replay buffer) and exploitation (high-reward buffer).

Given the importance of this balance, we further evaluate the sensitivity of \textsc{GFlowPO} to the total number of reference prompts $k$ while maintaining a similar replay/high-reward composition. All other settings are identical to those in \Cref{sec:induction_qa}. As shown in \Cref{tab:appen_dmu_sensitivity}, performance remains relatively stable for $k \geq 3$, with the configuration used in the main experiments ($k=3$) achieving the best performance. Performance degrades only when $k$ is too small ($k=2$), suggesting that \textsc{GFlowPO} is not particularly sensitive to the choice of $k$ within a reasonable range.

\begin{table}[!htb]

\centering
\setlength{\tabcolsep}{4pt}
\begin{tabular}{@{}lcccc@{}}
\toprule
$k$ & 2 & 3 & 4 & 5 \\
$(k_b, k_q)$ & (1, 1) & (2, 1) & (2, 2) & (3, 2) \\
\midrule
Avg. accuracy & 77.6 & \textbf{80.2} & 78.9 & \underline{79.2} \\
\bottomrule
\end{tabular}
\caption{Sensitivity of \textsc{GFlowPO} to the number of reference prompts $k=k_b+k_q$ in DMU. We report average accuracy over the same 10 II tasks as in \Cref{tab:prestep_reference_prompt_sampling_ablation}.}
\label{tab:appen_dmu_sensitivity}
\end{table}

\section{Prompt Evolution during \textsc{GFlowPO} Training}
\label{appen:prompt_evolution}
\begin{figure*}[!htbp]
    \centering
    \includegraphics[width=1.0\textwidth]{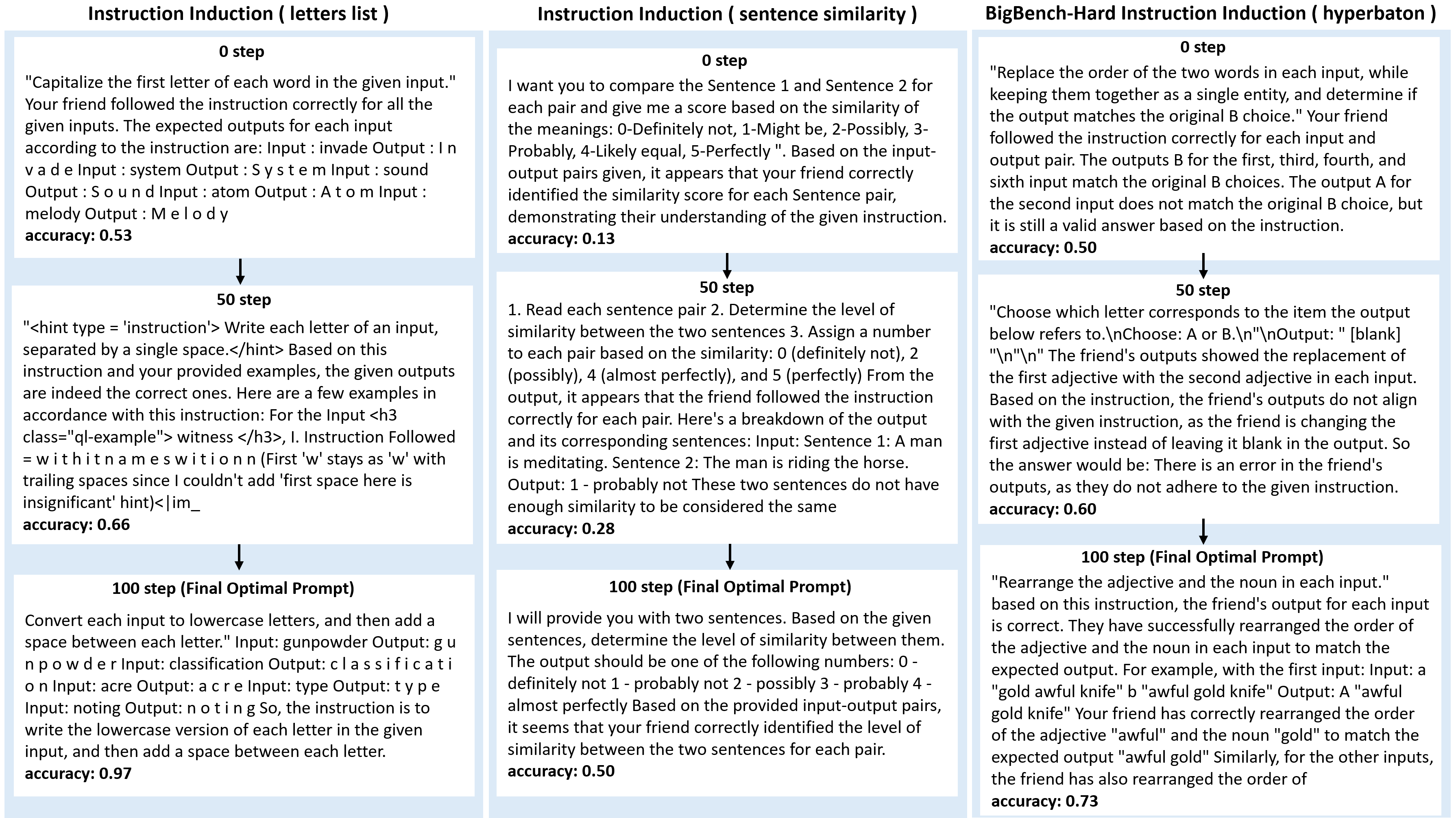}
    \caption{Prompt evolution during \textsc{GFlowPO} training. The prompts become increasingly precise and specific as training steps change. Initially (at step 0), prompts tend to include ambiguous or irrelevant context, but by step 50 they are refined to focus on the core task rules, and by step 100 they converge to a more explicit form with finer-grained criteria.}
    \label{fig:appen_promptevol}
\end{figure*}

In this section, we show how prompt evolves during \textsc{GFlowPO} training. We select two tasks from II, and one task from BBII. As in \Cref{fig:appen_promptevol}, the initial prompts tend to include ambiguous or irrelevant context, while the prompts are refined to extract the core task rules and converge to a more explicit form as step increases. This suggests that \textsc{GFlowPO} progressively explores and refines instructions that are practically effective for task performance throughout training.

\section{Information About Substantive AI Assistance}

We used AI assistants  during manuscript preparation to (i) detect errors in the references, (ii) verify bibliographic information of cited works (e.g., author lists, venues, and DOIs), (iii) check the manuscript against the ACL formatting guidelines, and (iv) suggest sentence-level edits for clarity and grammar. They were not used to generate research ideas, methods, experiments, analyses, or paper content.

\end{document}